\documentclass[twocolumn,journal]{IEEEtran}
\usepackage[T1]{fontenc}
\usepackage[latin9]{inputenc}
\usepackage{array}
\usepackage{booktabs}
\usepackage{mathtools}
\usepackage{multirow}
\usepackage{amsmath}
\usepackage{amssymb}
\usepackage{graphicx}
\usepackage[unicode=true,
 bookmarks=true,bookmarksnumbered=true,bookmarksopen=true,bookmarksopenlevel=1,
 breaklinks=false,pdfborder={0 0 0},pdfborderstyle={},backref=false,colorlinks=false]
 {hyperref}
\hypersetup{pdftitle={Your Title},
 pdfauthor={Your Name},
 pdfpagelayout=OneColumn, pdfnewwindow=true, pdfstartview=XYZ, plainpages=false}

\makeatletter

\providecommand{\tabularnewline}{\\}

\let\oldforeign@language\foreign@language
\DeclareRobustCommand{\foreign@language}[1]{%
  \lowercase{\oldforeign@language{#1}}}

\usepackage[caption=false,font=footnotesize]{subfig}
\usepackage[ruled,vlined]{algorithm2e}

\usepackage[table]{xcolor}
\usepackage{pgfplots}
\usepackage{tikz}
\usepackage{amsmath}
\usepackage{latexsym}
\usepackage{rotating}
\usepackage{lscape}
\usepackage{pgfplotstable}     
\usepgfplotslibrary{statistics}    
\usepackage{cite} 

\cellcolor[gray]{0.8}

\@ifundefined{showcaptionsetup}{}{%
 \PassOptionsToPackage{caption=false}{subfig}}
\usepackage{subfig}
\makeatother

\newcommand{\sectionname}{Section}

\pgfplotsset{compat=1.18} 

\begin{document}

\title{Generalized Population-Based Training for Hyperparameter Optimization in Reinforcement Learning}
\author{
Hui~Bai and Ran Cheng,~\IEEEmembership{Senior Member,~IEEE}\thanks{The authors are with the Department of Computer Science and Engineering,
Southern University of Science and Technology, Shenzhen 518055, China.
E-mail: \protect{huibaimonky@163.com},
\protect{ranchengcn@gmail.com}.
\emph{(Corresponding author: Ran Cheng)}}
}
\markboth{IEEE TRANSACTIONS ON Emerging Topics in Computational Intelligence, 2024}{}

\maketitle

\begin{abstract}
Hyperparameter optimization plays a key role in the machine learning domain. 
Its significance is especially pronounced in reinforcement learning (RL), where agents continuously interact with and adapt to their environments, requiring dynamic adjustments in their learning trajectories. 
To cater to this dynamicity, the Population-Based Training (PBT) was introduced, leveraging the collective intelligence of a population of agents learning simultaneously. 
However, PBT tends to favor high-performing agents, potentially neglecting the explorative potential of agents on the brink of significant advancements. 
To mitigate the limitations of PBT, we present the Generalized Population-Based Training (GPBT), a refined framework designed for enhanced granularity and flexibility in hyperparameter adaptation. 
Complementing GPBT, we further introduce Pairwise Learning (PL). 
Instead of merely focusing on elite agents, PL employs a comprehensive pairwise strategy to identify performance differentials and provide holistic guidance to underperforming agents. 
By integrating the capabilities of GPBT and PL, our approach significantly improves upon traditional PBT in terms of adaptability and computational efficiency. 
Rigorous empirical evaluations across a range of RL benchmarks confirm that our approach consistently outperforms not only the conventional PBT but also its Bayesian-optimized variant.
Source codes are available at \url{https://github.com/EMI-Group/gpbt-pl}.
\end{abstract}


\section{Introduction}

Deep neural networks (DNNs) have established themselves as the de facto function approximators in the realm of reinforcement learning (RL). 
Their powerful representational capacities have been instrumental in enabling RL to make significant inroads into a wide array of challenges. 
This includes solving deterministic systems like board games, as evidenced by AlphaGo \cite{Silver2016}, to mastering the stochastic dynamics of arcade games \cite{Mnih2015}. 
Furthermore, DNNs have been pivotal in addressing more complex, real-world challenges, such as robot control, where the interaction with the environment is multifaceted and nuanced \cite{Lillicrap2015}.

At the heart of these impressive RL accomplishments lies the intricate task of hyperparameter tuning. 
Both the RL algorithms, which dictate agent learning, and the DNN architectures, which define the model's complexity and capacity, are governed by a myriad of hyperparameters \cite{Chen2018a}. 
These hyperparameters, when optimally configured, have the potential to unlock the full capabilities of DNNs, as has been observed across various deep learning applications \cite{Elsken2018}. 
However, a unique challenge in RL is its non-stationary nature \cite{Parker-Holder2022}.
Unlike traditional supervised learning, where data distribution remains static, RL involves agents that learn from a dynamically changing environment. 
This constant evolution, influenced by the agent's interactions and its learning trajectory, means that a static set of hyperparameters may not remain optimal throughout the training process.

Recognizing the need for a dynamic and automated approach to hyperparameter tuning, the machine learning community has turned to hyperparameter optimization (HPO). 
Within the burgeoning field of automated machine learning (AutoML) \cite{Feurer2019}, HPO has become indispensable. The allure of HPO lies in its promise to reduce the manual, often tedious, trial-and-error based approach to model tuning. 
By automating this process, HPO not only enhances the efficiency of learning algorithms but also contributes to scientific rigor, ensuring experiments are reproducible and unbiased \cite{Feurer2019}. 
Predominantly, HPO techniques can be stratified into two camps: sequential optimization methods, where each evaluation informs the next (e.g., Bayesian optimization \cite{Wu2019}), and parallel search strategies, where multiple evaluations occur independently (e.g., random search and grid search \cite{Bergstra2012}).

Amidst the plethora of HPO techniques, the Population-Based Training (PBT) \cite{Jaderberg2017} has emerged as a front-runner, especially given its proven empirical successes across an array of computational domains \cite{Jaderberg2019,Liu2019a}.
What sets PBT apart is its ability to optimize DNN weights and hyperparameters in tandem, by extracting and aggregating insights from a population of agents during a single training run. 
PBT's asynchronous nature means that agents can periodically refine their hyperparameters by emulating better-performing counterparts, thereby charting an effective hyperparameter trajectory. 
Yet, PBT is not without its limitations. 
Its propensity to focus primarily on top-performing agents can sometimes stymie broader exploration. 
This bias towards \emph{winners} can lead to premature convergence, neglecting agents that might have exhibited superior performance given more time or slightly different conditions -- the so-called \emph{late bloomers}.

To address the inherent limitations of PBT, we introduce the Generalized Population-Based Training (GPBT), a flexible HPO framework that builds upon PBT's asynchronous parallelism. 
In GPBT's HPO phase, agents are randomly paired, offering the opportunity to adjust hyperparameters through user-defined strategies. 
Coupled with GPBT, we have tailored the Pairwise Learning (PL) method, which leverages a pseudo-gradient approach reminiscent of Stochastic Gradient Descent with Momentum (SGDM) \cite{Liu2020e}.
PL computes a pseudo-gradient for underperforming agents based on the performance difference with their paired counterparts. 
These agents subsequently refine their behavior using this derived gradient and their previous updates. 
By continually resampling throughout training and harnessing the aggregate knowledge of the population, agents iteratively refine their update trajectories towards optimal directions.
Crucially, our approach retains a broad spectrum of agents, both high-performing and those lagging, ensuring a diverse population. This diversity fosters exploration, enabling agents to navigate beyond local optima, striving for superior performance outcomes.
In summary, our main contributions are:
\begin{itemize}
\item  We present GPBT, a versatile HPO framework that builds upon the foundational principles of PBT. 
Distinctively, GPBT is architected to be inherently adaptable, accommodating a broad range of optimization strategies. 
This adaptability ensures that GPBT remains pertinent across diverse hyperparameter tuning contexts, offering researchers and practitioners a flexible tool that can be tailored to specific challenges and scenarios.

\item  We have conceptualized and developed PL as an optimization method tailored for HPO. 
At its core, PL leverages pseudo-gradients, which serve as heuristics to guide the update trajectories of agents, particularly in complex black-box HPO landscapes. 
This method ensures that agents can make informed adjustments to their behaviors, even when the optimization landscape is intricate and lacks explicit gradients.

\item  We have empirically assessed the efficacy of the integrated GPBT-PL approach (GPBT with PL) in the realm of HPO in RL. 
Through rigorous experiments benchmarked against a comprehensive suite of OpenAI Gym environments, we provide conclusive evidence of GPBT-PL's robust performance. 
Notably, our findings underscore its superior performance relative to the traditional PBT approach and its Bayesian-optimized variant. 
This superiority is maintained even when computational resources are stringent, attesting to the efficiency and effectiveness of our proposed methodology.
\end{itemize}

The paper is structured as follows: \sectionname~\ref{sec:Related-Work} reviews relevant literature, \sectionname~\ref{sec:Motivation} presents the research motivation, \sectionname~\ref{sec:Proposed-Approach} describes our approach, \sectionname~\ref{sec:Experiments} discusses experimental results, and \sectionname~\ref{sec:Conclusions} provides concluding remarks.

\section{Related Work \label{sec:Related-Work}}

HPO is indispensable for optimizing machine learning model performance. As models grow in complexity, the demand for sophisticated HPO techniques intensifies. 
This section provides an overview of the evolution of HPO, from general-purpose methods to those specifically crafted for RL, with a particular focus on population-based HPO methods.

\subsection{General HPO Methods}
At its core, HPO is designed to address black-box optimization problems where the objective function is typically non-differentiable. 
This necessitates iterative sampling and evaluation of multiple hyperparameter configurations to identify the most effective setup. 

Typically, any black-box optimization method can be utilized for HPO. 
The most fundamental HPO method is grid search, and however this method suffers from the curse of dimensionality. A better alternative method is random search, which provides a valuable baseline as it is anticipated to converge toward optimal performance if given enough resources.
Though straightforward, these two methods often lack computational efficiency, evaluating unpromising models extensively \cite{Bergstra2012}.
When introducing guidance, population-based methods such as evolutionary algorithms usually perform better than random search by applying local perturbations (mutations) and combinations of different members (crossover) to generate a new generation of improved configurations.
Additionally, Bayesian optimization emerged as a pivotal method, leveraging probabilistic models to predict promising hyperparameters \cite{Snoek2012}. Though this method can efficiently explore the search space by updating a surrogate probabilistic model of the objective function and using an acquisition function to guide the search toward promising regions, it is computationally intensive, especially for high-dimensional parameter spaces or expensive-to-evaluate objective functions.
Finally, multi-fidelity optimization entails a balance between optimization efficacy and runtime efficiency by evaluating a configuration on a small data subset or with limited resources.
Despite these advancements, HPO still grapples with challenges like extended optimization durations and striking the right exploration-exploitation balance.

\subsection{HPO in RL}
In RL, hyperparameters play a pivotal role in shaping agent-environment dynamics, thereby directly influencing learning trajectories and decision outcomes. 
A unique challenge in RL is the non-stationarity of the environment. 
As the agent iteratively updates its policy, its interactions with the environment evolve, making the hyperparameter tuning process intricate.
At times, minor hyperparameter adjustments can yield profound impacts on agent performance.

Several methods have been utilized, albeit without specific customization for HPO in RL, such as the multi-fidelity optimization methods (e.g., HyperBand \cite{Li2017d} and  ASHA \cite{Li2018b}), which incorporate dynamic resource allocation and early termination for unpromising configurations.
Nonetheless, these methods might prematurely discard promising configurations and encounter challenges in striking a balance between performance and computational budgets.

In contrast, several methods have been tailored for HPO in RL. 
In \cite{Eriksson2003}, the learning rate $\alpha$ and temperature $\tau$ in Sarsa($\lambda$) have been optimized by genetic algorithms to balance exploration and exploitation for food capture tasks, which integrates learning and evolution to effectively enhance the performance of RL algorithms and obtain sim-to-real robust policies.
Another method involves executing parallel RL instances with distinct initial hyperparameters, augmented with Gaussian noise \cite{Elfwing2018}. 
Notably, an off-policy HPO method for policy gradient RL algorithms has emerged  \cite{Paul2019}, which estimates the performance of candidate configurations by an off-policy method and updates the current policy greedily. 

Recently, the population-based HPO methods have risen to prominence, offering dynamic hyperparameter adjustments throughout training, free from usage restrictions \cite{Jaderberg2017,Jaderberg2019,Parker-Holder2020a}.

\subsection{Population-Based HPO Methods}
In contrast to traditional HPO strategies that sequentially optimize individual hyperparameter sets, population-based methods optimize multiple configurations in parallel~\cite{Bai2023}.
These methods maintain a diverse set of hyperparameters and employ evolutionary algorithms to navigate the hyperparameter space, ensuring a harmonious blend of exploration and exploitation.

Key population-based methods include Bayesian Optimization and HyperBand (BOHB) \cite{Falkner2018}, Genetic HPO \cite{Aszemi2019}, and PBT. 
For instance, BOHB marries Bayesian optimization's probabilistic function modeling with HyperBand's resource-efficient early termination. 
Genetic HPO employs evolutionary techniques like crossover and mutation to generate new hyperparameter sets.
However, these methods may encounter challenges in scenarios that demand extensive evaluations or have highly variable evaluation durations, such as in RL.

\subsection{Population-Based Training}
Population-based Training (PBT)  \cite{Jaderberg2017} stands out as a representative approach in the realm of population-based HPO methods.
Central to PBT is the notion of Lamarckian evolution, wherein agents not only inherit but also evolve their attributes. 
This ensures that hyperparameters are dynamically attuned to the ongoing learning phase, thereby optimizing learning outcomes.

In PBT, each agent, armed with specific weights and hyperparameters, periodically assesses its performance based on a predefined step count, signifying its \emph{ready} state. Once in this state, all agents are ranked by performance. Agents who do not measure up to their peers adopt attributes from superior agents through a two-pronged strategy:
\begin{itemize}
\item \emph{exploit}: The lagging agent inherits both weights and hyperparameters from a superior agent.
\item \emph{explore}: Hyperparameters are subjected to random perturbation, either amplified by a factor of 1.2, diminished by 0.8, or resampled according to their original distribution.
\end{itemize}
This strategy empowers each agent to traverse multiple hyperparameter landscapes during its training journey, an invaluable trait considering the fluidity inherent to deep RL agent training.

A key strength of PBT lies in its amalgamation of both sequential and parallel optimization techniques. 
Its asynchronous architecture ensures continuous training for some agents even as others update, resulting in superior performance across varied applications \cite{Paul2019,Espeholt2018,Wu2020}. 
This asynchronous paradigm augments training efficiency, replacing stagnating configurations with emerging ones and introducing slight random perturbations, all while other configurations proceed undisturbed. PBT's proficiency in sculpting hyperparameter schedules tailored for complex RL tasks has been well-documented and validated across numerous benchmarks \cite{Jaderberg2019, Schmitt2018, Liu2019a}.

Except for the PBT paradigm, several other population-guided methodologies have been proposed for RL \cite{Jung2020, Liu2021b, Khadka2018, Khadka2019, Majumdar2020}. However, they have distinctive functional differences in the population, which sample diverse experiences through the evolution of populations to address the exploration challenges of gradient-based RL algorithms.

\section{Motivation \label{sec:Motivation}}
PBT's asynchronous parallel paradigm has indeed proven efficient in a multitude of scenarios. 
Nonetheless, its predominant focus on elite agents inadvertently stymies broader exploration capabilities due to its inherent predilection for immediate gains. 
Specifically, the limitation manifests in two primary ways:
\begin{itemize}
\item 
PBT's direct replacement strategy might prematurely discard promising regions of the search space. 
These areas, albeit seemingly suboptimal in the short term, might harbor superior solutions in a longer timeframe.
\item 
The overemphasis on exploiting top-tier solutions raises the specter of converging to local optima.
While such a strategy may deliver satisfactory solutions in the short run, it contravenes the quintessential ethos of population-based HPO, which is to strike a judicious balance between exploration and exploitation.
\end{itemize}

Maintaining diversity within the population is of paramount importance, especially in the context of RL tasks. 
A diverse population fosters the emergence of varied behaviors, culminating in more resilient and holistic performance solutions, as underscored by the principles of novelty search techniques \cite{Conti2018, Lehman2011}.
PBT's structural design, which typically yields a single solution per generation, accentuates the importance of crafting diverse and efficient successors.

In light of these challenges, the avant-garde Population-Based Bandits (PB2) technique leverages Bayesian optimization to invigorate underperforming agents, outclassing PBT in terms of efficiency, albeit at an elevated computational expense \cite{Parker-Holder2020a}. 
However, Bayesian optimization, with its computational intricacies, often imposes overheads that sometimes eclipse the actual costs associated with hyperparameter evaluations \cite{Golovin2017}. 
As a result, neither PBT nor PB2 truly achieve their full efficiency potential.
More recently, while advanced PBT variants like FIRE PBT and BG-PBT \cite{Dalibard2021,Wan2022} have emerged, they often grapple with challenges tied to their inherent greedy behavior or the time-intensive process of generating new hyperparameters.

To redress these shortcomings, we introduce the Generalized Population-Based Training (GPBT) framework. 
While it builds on PBT's asynchronous underpinnings, GPBT distinguishes itself by replacing PBT's direct substitution method with a nuanced dual-agent learning mechanism. 
When viewed through the lens of evolutionary computation (EC), GPBT aligns with the steady-state EC paradigm, where each iteration introduces a single new agent. 
This characteristic makes GPBT especially adept at navigating dynamic environments, echoing proficient steady-state EC methodologies \cite{Vavak1996}.
Interestingly, both PBT and PB2 can be seamlessly encapsulated within the GPBT framework, with the key differentiation being their respective strategies for engendering new agents.
Essentially, PBT can be considered a variant within the GPBT framework. In this context, PBT differentiates itself through its specific method for hyperparameter update, which is conducted via random perturbation.

To further enhance GPBT and expedite hyperparameter refinement, we propose the Pairwise Learning (PL) paradigm. 
In this method, a lagging agent refines its parameters by leveraging insights from a superior counterpart, thus illuminating potentially overlooked regions of the search space. 
The diversity introduced by myriad pairings throughout training ensures a comprehensive exploration. 
Continual steering towards propitious directions allows the solution sets to progressively gravitate towards optimal configurations. 
Drawing inspiration from the Stochastic Gradient Descent with Momentum (SGDM) methodology~\cite{Liu2020e}, which is prevalent in deep learning, PL employs a pseudo-gradient as an approximation to the elusive genuine gradient of the hyperparameter objective function. 
To ensure stable and coherent updates, PL incorporates historical data of the lagging solution as a momentum component.

\section{Proposed Approach\label{sec:Proposed-Approach}}

We begin by delineating the problem statement for HPO and discussing the nuances and constraints of PBT in \sectionname~\ref{subsec:Problem-Statement}. 
We then introduce the GPBT framework and discuss its efficacy in \sectionname~\ref{subsec:Generalized-Population-Based}. 
Finally, we detail the implementation of PL for hyperparameter updates in \sectionname~\ref{subsec:Pairwise-Learning}.

\subsection{Problem Statement \label{subsec:Problem-Statement}}
HPO is tasked with identifying an optimal hyperparameter vector \(\boldsymbol{x}\) within the search domain \(\mathcal{D}\in\mathbb{R}^{d}\), where \(d\) denotes the number of hyperparameters. 
In RL, this is tantamount to maximizing the cumulative reward across domain \(\mathcal{D}\). 
The cumulative reward, represented as total discounted reward \(\sum_{t\geq0}\gamma^{t}r_{t}\) with \(\gamma\) as the discount factor, is derived from a trajectory \(\tau=(s_{0},a_{0},r_{0},s_{1},...)\) depicting the interplay between an RL agent and its environment. 
Here, the agent's policy governs its actions \(a\) based on the current state \(s\), and the environment provides a corresponding reward \(r\).
The policy, \(\pi_{\theta}\), maps the environment's state to the agent's actions, and is implemented using a neural network with weights \(\theta\). 
Thus, the HPO challenge can be articulated as a bi-level optimization problem:

\begin{equation}
\begin{array}{c}
\begin{array}{ccc}
\underset{\boldsymbol{x}}{max}f(\boldsymbol{x},\theta^{*}) & \mathrm{s.t.} & \theta^{*}\in\mathrm{arg}\underset{\theta}{max}J(\theta;\boldsymbol{x})\end{array}\\
\begin{array}{ccc}
\underset{\theta}{max}J(\theta;\boldsymbol{x}) & \mathrm{where} & J(\theta;\boldsymbol{x})=\mathbb{E}_{\tau\sim\pi_{\theta}}[\underset{t\geq0}{\sum}\gamma^{t}r_{t}]\end{array}
\end{array},
\end{equation}
where the outer loop optimization problem is \(max_{\boldsymbol{x}}f(\boldsymbol{x},\theta^{*})\), and the inner loop optimization problem is \(max_{\theta}J(\theta;\boldsymbol{x})\).

Specially, in population-based HPO, we consider \(n\) agents \((\boldsymbol{x}_{1}, \boldsymbol{x}_{2}, ..., \boldsymbol{x}_{n})\), with each agent evolving its hyperparameters over time \((\boldsymbol{x}_{1}^{t}, \boldsymbol{x}_{2}^{t}, ..., \boldsymbol{x}_{n}^{t})_{t=1,...,T}\), where \(t\) represents the elapsed time steps, epochs, or iterations of an agent's training.

\begin{figure*}[tbh]
\centering
\includegraphics[scale=0.7]{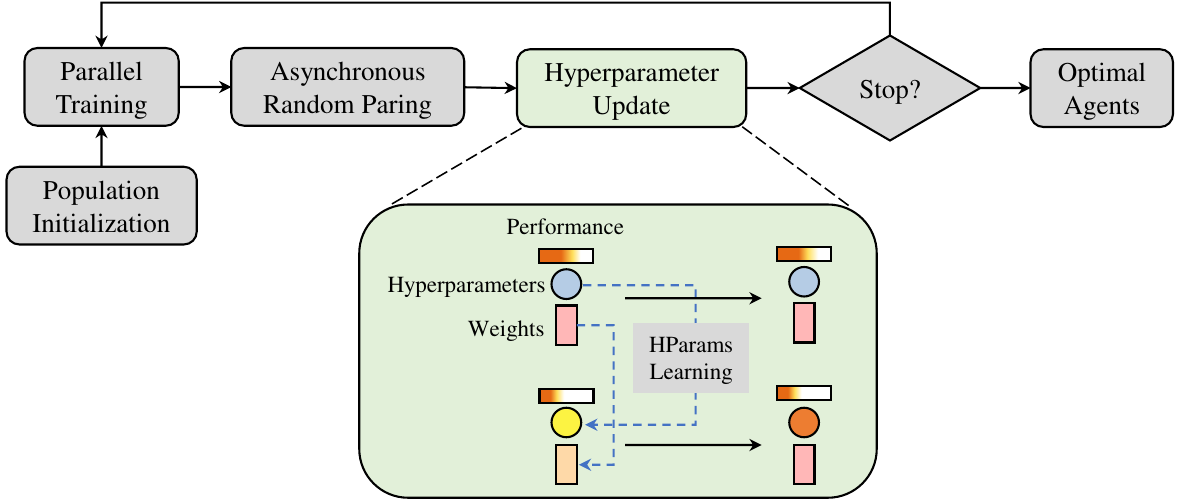}
\caption{\label{fig:GPBT framework} Framework of Generalized Population-Based Training (GPBT). 
A population of agents are initialized with random weights and hyperparameters and then trained in parallel. Upon reaching designated hyperparameter update intervals, ready agents undergo asynchronous random pairing for updates. If the ready agent underperforms, it adopts the weights of its superior counterpart and updates its hyperparameters using specialized learning techniques. After fulfilling the stopping criteria, top-performing agents are identified.}
\end{figure*}

\subsection{Generalized Population-Based Training (GPBT) \label{subsec:Generalized-Population-Based}}

\figurename~\ref{fig:GPBT framework} illustrates the GPBT framework. 
A population of agents are initialized with random weights and hyperparameters, and then trained and evaluated in parallel. Upon reaching designated hyperparameter update intervals (i.e., perturbation interval), ready agents undergo asynchronous random pairing to form parent pairs. 
If a ready agent underperforms, it adopts the weights of its superior counterpart and adjusts its hyperparameters using specialized learning techniques (e.g., random perturbation or pairwise learning).
After fulfilling the stopping criteria, top-performing agents are identified.

From an evolutionary computation (EC) standpoint, GPBT aligns with the steady-state EC methodology, which introduces one new agent per iteration. 
Steady-state EC is recognized for its aptitude in addressing non-stationary challenges characterized by gradual, low-frequency alterations \cite{Vavak1996,Jiang2017}. 
Such dynamics closely mirror HPO scenarios, where consecutive training sessions often involve subtle hyperparameter modifications. 
A salient feature of steady-state EC is its quick adaptability, made possible as the newly introduced agent immediately joins the mating pool. 
This swift integration facilitates an early progression towards the optimal solution during the optimization phase \cite{Vavak1996,Dyer2012}.
Considering system stability, the steady-state EC introduces minimal diversity to the population after an environmental modification, like a hyperparameter revision. 
Consequently, GPBT not only functions as an asynchronous HPO framework but also excels in stably navigating changes in non-static scenarios.

In population-based strategies, a pivotal challenge arises when deciding which agent should be replaced by new entrants \cite{Lozano2008}.
Traditional techniques in steady-state EC typically advocate for replacing either the oldest agent or the least performing one. 
However, these strategies are not directly applicable to the GPBT paradigm. 
A primary reason is the asynchronous design: if the system engages in continual evaluations to identify the least effective agent, the benefits of asynchrony are undermined. 
Intriguingly, within GPBT, replacements are exclusively between paired agents, often resulting in the substitution of the ready (and comparatively older) underperforming agent. 
This strategy presents a nuanced balance between age-driven and performance-driven replacements. 
Such a strategy not only maintains the asynchrony but also fosters diversity within the population.

This emphasis on diversity is further underscored by various studies that delve into the evolutionary trajectory of EC. 
They suggest an incremental improvement in population performance over iterations \cite{Cormen2001}. 
At a high level, when two high-performing agents are paired, the likelihood of producing an equally or more competent offspring is elevated compared to pairings between lower-performing agents. 
However, this does not insinuate that elite pairings always produce top-tier offspring, nor is there an assurance that all initial agents will exhibit high performance. 
Consequently, the design of offspring generation algorithms is critical, with an emphasis on consistently producing high-quality agents.
To address these challenges, especially in the context of HPO in RL, we introduce the Pairwise Learning (PL) method.

\subsection{Pairwise Learning (PL) \label{subsec:Pairwise-Learning}}

For the preservation of promising candidates and the formulation of optimal hyperparameters, we introduce the Pairwise Learning (PL) approach tailored for hyperparameter refinement. 
In PL, beyond the hyperparameters \( \boldsymbol{x} \) and weights \( \theta \), agents possess a \( d \)-dimensional velocity vector \( \boldsymbol{v} \) designated for hyperparameter adjustments, with \( d \) denoting hyperparameter count. 
This vector is initialized to zero. 
Each generation witnesses the random pairing of two agents, followed by performance comparisons. 
The superior performing agent, termed the 'fast learner', is directly incorporated into the population. Conversely, the lesser-performing `slow learner' adopts the weights of its counterpart and amends its hyperparameters and velocity through a learning mechanism derived from the fast learner.

Consider \( \boldsymbol{x}_{f}^{g} \), \( \boldsymbol{x}_{s}^{g} \), \( \boldsymbol{v}_{f}^{g} \), and \( \boldsymbol{v}_{s}^{g} \) as the hyperparameters and velocities of the fast and slow learners at generation \( g \) respectively. The slow learner's updates are guided by:
\begin{equation}
\boldsymbol{x}_{s}^{g+1}=\boldsymbol{x}_{s}^{g}+\boldsymbol{v}_{s}^{g+1},\label{eq:2}
\end{equation}
\begin{equation}
\boldsymbol{v}_{s}^{g+1}=\boldsymbol{r}_{1}\boldsymbol{v}_{s}^{g}+\boldsymbol{r}_{2}(\mathrm{G}_{f}^{g}(\boldsymbol{x}_{f};\boldsymbol{u}_{f})-\mathrm{G}_{s}^{g}(\boldsymbol{x}_{s};\boldsymbol{u}_{s})),\label{eq:3}
\end{equation}
where \( \boldsymbol{r}_{1} \) and \( \boldsymbol{r}_{2} \) are uniformly distributed random vectors within \( [0,1]^{d} \). 
The terms \( \mathrm{G}_{f}^{g}(\boldsymbol{x}_{f};\boldsymbol{u}_{f}) \) and \( \mathrm{G}_{s}^{g}(\boldsymbol{x}_{s};\boldsymbol{u}_{s}) \) symbolize distributions of the fast and slow learners respectively. 
The difference between these distributions reflects in \( \mathrm{G}_{f}^{g}(\boldsymbol{x}_{f};\boldsymbol{u}_{f})-\mathrm{G}_{s}^{g}(\boldsymbol{x}_{s};\boldsymbol{u}_{s}) \). 
Therefore, both learners can be conceptualized as samples from their respective distributions.

Distinguishing between learners with varying performances is achieved through their distributions. 
Sorting based on performance, top-tier agents epitomize fast learner distributions, while their lower-tier counterparts represent slow learners. 
This stratified approach enhances the learning efficiency of slow learners. 
Continuous sampling ensures gradient corrections between distributions, gravitating towards optimal gradients.

Inspired by Stochastic Gradient Descent with Momentum (SGDM), a prevalent optimization strategy in deep learning, PL operates as a pseudo-gradient-driven approach. 
SGDM, an enhancement of conventional SGD \cite{Livni2014}, integrates momentum to expedite optimization convergence. 
SGDM's update equations for a maximization problem involving parameter \( \theta \) are:
\begin{equation}
\theta^{t+1}=\theta^{t}+\boldsymbol{v}^{t+1},
\end{equation}
\begin{equation}
\boldsymbol{v}^{t+1}=\beta\boldsymbol{v}^{t}+\eta \times gradient,
\end{equation}
where \( \beta \) dictates momentum contribution and \( \eta \) modulates the gradient's learning step size. The term \( \beta\boldsymbol{v}^{t} \) acts as a velocity component, aiding SGD in oscillation mitigation and convergence acceleration by capturing parameter movement trends across iterations.

Contrary to SGDM, which computes true gradients of a loss function via a training data subset, PL estimates surrogate gradients concerning hyperparameter values using merely two agents. 
In essence, PL embodies an SGDM variant with a batch size of one, optimized for PBT's asynchronous nature. 
Given that PL's gradient estimates are based on dual samples, the learning direction occasionally deviates from the optimal, introducing noise. Introducing the momentum term \( \boldsymbol{r}_{1}\boldsymbol{v}_{s}^{g} \) in PL averages out this noise, offering a refined estimate closer to the original function's precise derivation. 
Consequently, PL's frequent hyperparameter updates yield rapid convergence. 
Apart from its ease of implementation, convergence guarantees, and scalability, PL also mirrors traits of the competitive swarm optimizer \cite{Cheng2014}, wherein losers are made to learn from the winners via randomly paired competitions. 
In essence, PL exemplifies hyperparameter update strategies, and further optimization techniques from gradient-agnostic methods can be woven into GPBT to enhance hyperparameter update methodologies.

\subsection{GPBT-PL}

\begin{algorithm}[tbh]
\footnotesize
\caption{Generalized Population-Based Training with Pairwise Learning (GPBT-PL)}
\LinesNumbered  
\KwIn{Population size $n$, Hyperparameter ranges $R$, Perturbation interval $\delta$, Training cost $T$.}
\KwOut{The optimal agent.} 
$t \leftarrow 0$\;
$P \leftarrow$ initialize population with random HParams and weights\; 
\For{agents $\in$ $P$}{
	\While{$t < T$}{ 
		$P \leftarrow $ parallel training of population $P$\;
		\If{$t$ mod $\delta==0$}{
			$a(\theta,\boldsymbol{x}) \leftarrow$ get the ready agent\;
			$P \leftarrow $ rank population $P$ according to performance\;
			$a'(\theta',\boldsymbol{x}') \leftarrow$ select a better-performing agent\;
			$a(\theta)=a'(\theta')$\;
			$a(\boldsymbol{x}) \leftarrow Pairwise Learning(a(\boldsymbol{x}), a'(\boldsymbol{x}'))$ using (2)\&(3)\;
		}
	} 
}
\label{alg:GPBT-PL}
\end{algorithm}

By marrying the flexibility and asynchronous features of GPBT with the adaptive learning mechanisms of PL, we provide the integrated approach -- GPBT-PL.
Algorithm 1 provides a detailed procedure for this integrated approach. 
Commencing with the initialization of agents with random hyperparameters and weights, the algorithm then delves into the parallel training of the entire population. 
Periodic checks, as determined by the perturbation interval \( \delta \), identify agents ready for updates. 
These agents then leverage the PL mechanism to adapt their hyperparameters, ensuring a continuous push toward enhanced performance.

Unlike PBT, which follows a predetermined update strategy, GPBT offers a more adaptable platform for hyperparameter refinement. 
This adaptability is further enhanced with the integration of PL, ensuring that hyperparameters are fine-tuned through a comparative process between two agents. 
This promotes knowledge sharing and quick adaptation, capitalizing on the strengths of both the \emph{fast learner} and the \emph{slow learner} agents.

Another crucial aspect of GPBT-PL lies in its asynchronous nature, where agents can undergo updates without waiting for the entire population to be ready. 
This ensures that the system remains dynamic and responsive to changes via continuously evolving and adapting.
Furthermore, the modularity of GPBT-PL allows for extensibility. 
While the current implementation is grounded on PL for hyperparameter updates, the framework can easily accommodate other learning or optimization methods, making it a versatile tool in the domain of HPO.

\section{Experiments \label{sec:Experiments}}

In our experimental design, we categorize our focus into two main areas: on-policy RL and off-policy RL. 
First, we conduct experiments to assess the general performance of the proposed GPBT-PL in several tasks in each area. 
Then, acknowledging the sensitivities of PBT-based HPO algorithms to perturbation intervals, hyperparameter boundaries, and scalability with larger populations, we design experiments emphasizing robustness and scalability.
Specifically, for on-policy RL, we assess the stability against perturbation interval variations and population size changes; for off-policy RL, we examine the HPO algorithms' robustness to alterations in the hyperparameter range.

\begin{table}[tbh]
\caption{\label{tab:Parameter-settings}Parameter settings}

\hfill{}%
\begin{tabular}{llc}
\toprule 
\multirow{2}{*}{RL Algorithms} & \multirow{2}{*}{Hyperparameter} & \multirow{2}{*}{Value}\tabularnewline
 &  & \tabularnewline
\midrule 
\multirow{10}{*}{PPO} & Batch size & {[1000, 60000]}\tabularnewline
 & GAE $\lambda$ & [0.9, 1.0)\tabularnewline
 & PPO Clip $\epsilon$ & 0.99, [0.95, 1.0)\tabularnewline
 & Learning Rate $\eta$ & [10$^{-5}$, 10$^{-3}$)\tabularnewline
 & Discount $\gamma$ & 0.99, [0.95, 1.0)\tabularnewline
 & SGD Minibatch Size & 128, [16, 256]\tabularnewline
 & SGD Iterations & 10, [5, 15]\tabularnewline
 & Policy Architecture & \{32, 32\}\tabularnewline
 & Filter & MeanStdFilter\tabularnewline
 & Population Size & \{4, 8, 16\}\tabularnewline
 & Perturbation Interval & \{1$\times$10$^{4}$, 5$\times$10$^{4}$\}\tabularnewline
\midrule 
\multirow{9}{*}{IMPALA} & Epsilon & [0.01, 0.5)\tabularnewline
 & Learning Rate $\eta$ & [10$^{-5}$, 10$^{-3}$), [10$^{-5}$, 10$^{-2}$)\tabularnewline
 & Entropy Coefficient & [0.001, 0.1)\tabularnewline
 & Batch Size & 500\tabularnewline
 & Discount $\gamma$ & 0.99\tabularnewline
 & SGD Minibatch Size & 500\tabularnewline
 & SGD Iterations & 1\tabularnewline
 & Policy Architecture & \{256, 256\}\tabularnewline
 & Population Size & 4\tabularnewline
 & Perturbation Interval & 5$\times$10$^{4}$\tabularnewline
\midrule 
\multicolumn{3}{c}{Common Hyperparameters}\tabularnewline
\midrule
\multicolumn{2}{l}{Number of Workers} & 5\tabularnewline
\multicolumn{2}{l}{Number of GPUs} & 0\tabularnewline
\multicolumn{2}{l}{Optimizer} & Adam\tabularnewline
\multicolumn{2}{l}{Nonlinearity} & Tanh\tabularnewline
\bottomrule
\end{tabular}\hfill{}
\end{table}

\begin{figure*}[tbh]
\hfill{}%
\begin{tabular}{cccc}
\subfloat[BipedalWalker]{\includegraphics[scale=0.245]{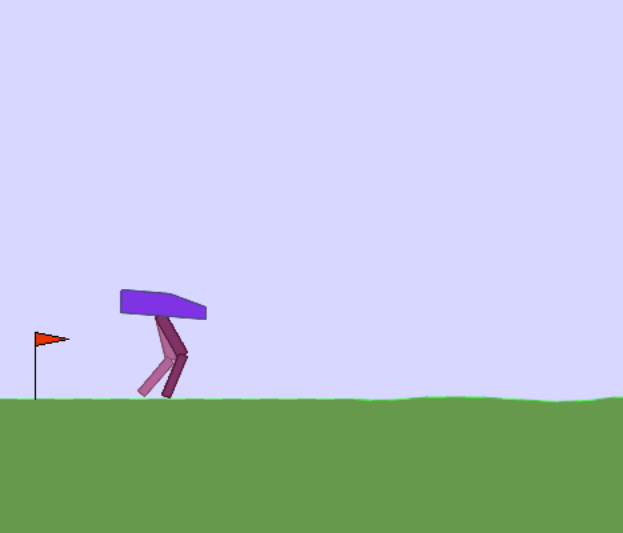}

} & \subfloat[Ant]{\includegraphics[scale=0.19]{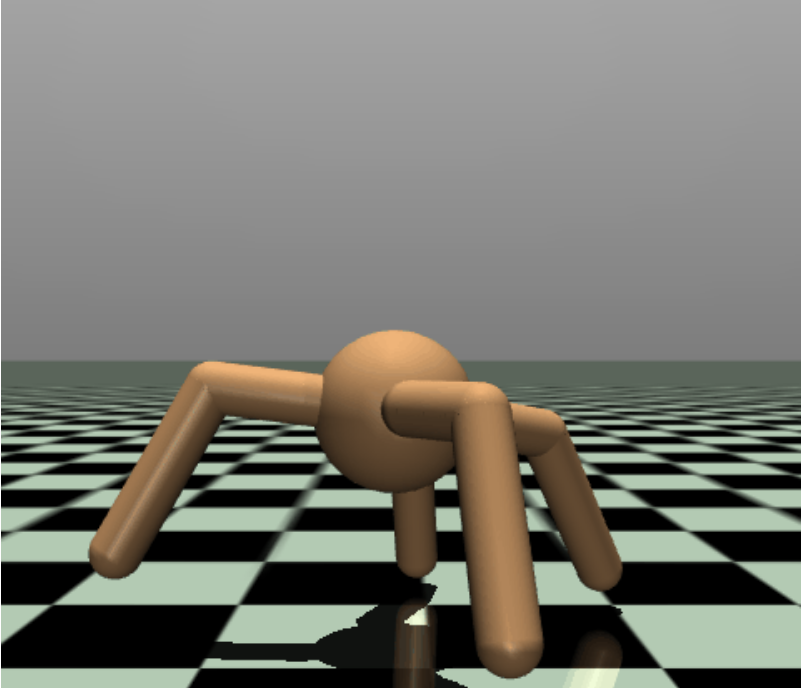}

} & \subfloat[HalfCheetah]{\includegraphics[scale=0.197]{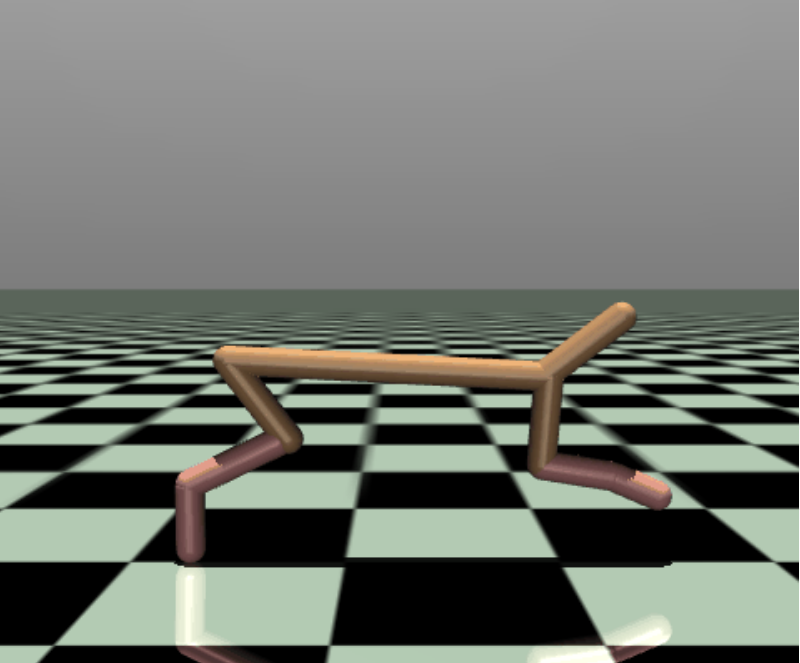}

} & \subfloat[InvertedDoublePendulum]{\includegraphics[scale=0.255]{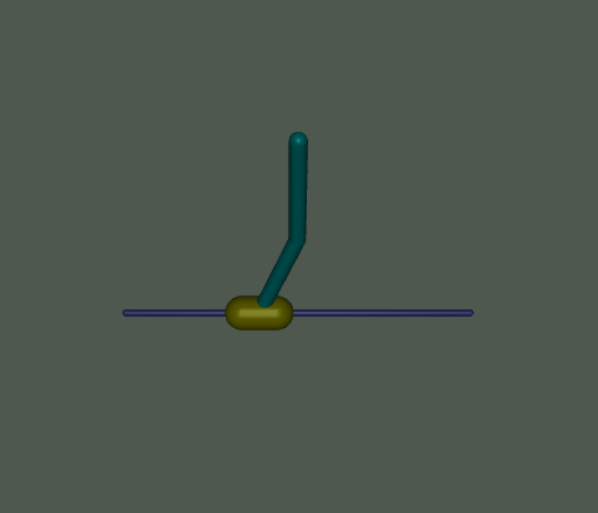}

}\tabularnewline
\subfloat[Swimmer]{\includegraphics[scale=0.255]{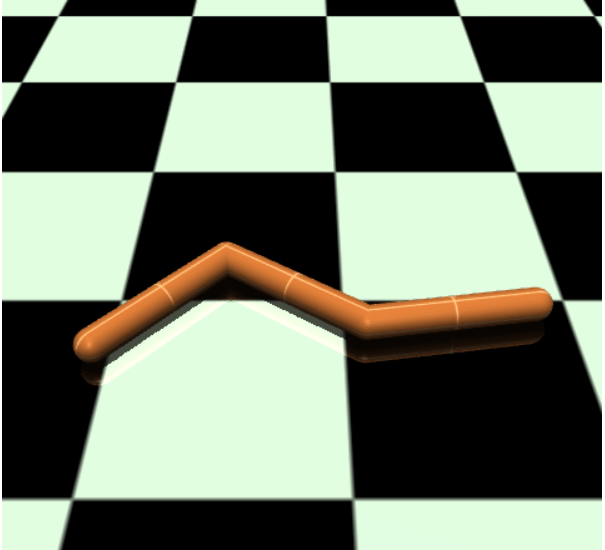}

} & \subfloat[Walker2D]{\includegraphics[scale=0.255]{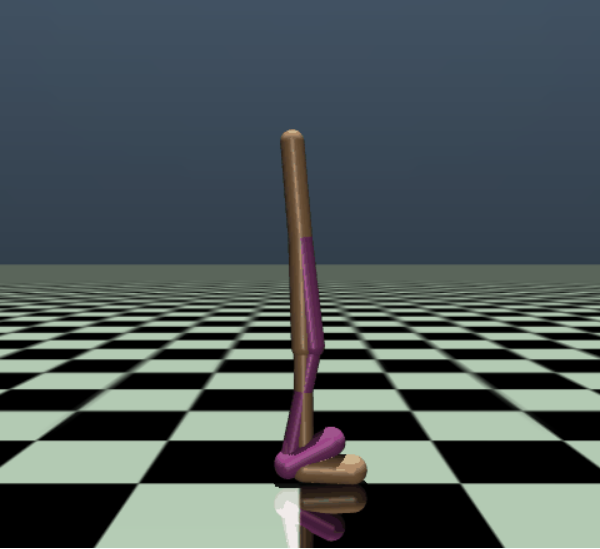}

} & \subfloat[Breakout]{\includegraphics[scale=0.225]{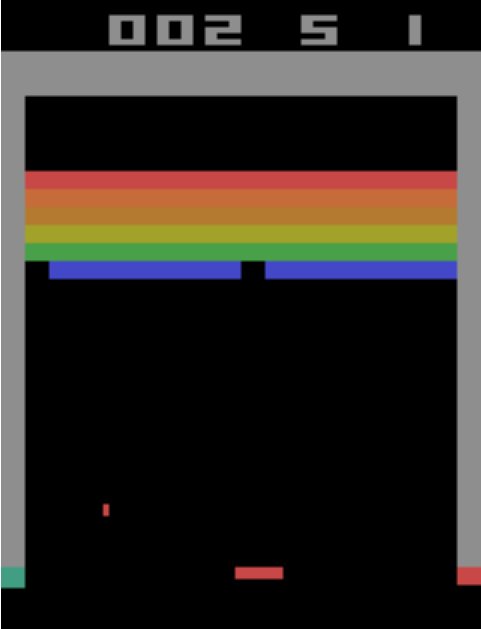}

} & \subfloat[SpaceInvaders]{\includegraphics[scale=0.225]{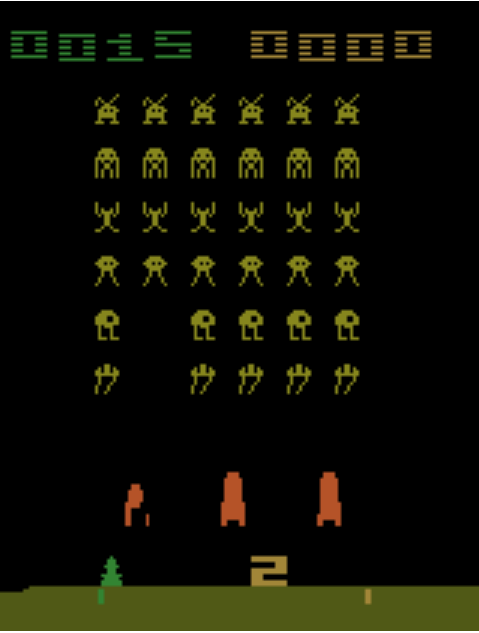}

}\tabularnewline
\end{tabular}\hfill{}

\caption{\label{fig:tasks}Eight RL tasks selected from OpenAI Gym.}
\end{figure*}

\subsection{Experimental Settings}

Our experiments predominantly target the RL domain, emphasizing its notorious susceptibility to hyperparameters \cite{Henderson2018}. 
We embarked on an extensive study across a plethora of tasks from OpenAI Gym, version five \cite{Brockman2016}. 
Fig.~\ref{fig:tasks} showcases our chosen eight task scenarios. 
The experimentation bifurcates into:
\begin{enumerate}
    \item Optimizing four hyperparameters for the on-policy RL algorithm, Proximal Policy Optimization (PPO) \cite{Schulman2017}, across the initial six continuous control challenges.
    \item Refining three hyperparameters for the off-policy RL algorithm, Importance Weighted Actor-Learner Architecture (IMPALA) \cite{Espeholt2018}, for the concluding two discrete control tasks.
\end{enumerate}
Each experimental run is iterated with seven distinct seeds. 
Tables present the apex mean rewards across all seeds, which are defined as the average of the final 10 episodic rewards during training. 
Correspondingly, figures illustrate both the mean and standard deviation of these pinnacle rewards. 
Our objective is to spotlight the zenith of mean rewards within tables, a crucial metric in practical applications. 
All experiments were facilitated by the Ray Tune library \cite{Moritz2018} and Ray RLlib \cite{Liang2018}.

\subsubsection{Hyperparameter Settings} 
Table~\ref{tab:Parameter-settings} catalogues the optimized hyperparameters, their respective boundaries, and certain constant hyperparameters. 
For both experimental categories, we scrutinized population sizes \( n \in \{4,8\} \) and designated five workers for every agent, ensuring the algorithm's local executability on contemporary computational platforms. 
For the on-policy RL experiments, we delved into robustness against perturbation interval adjustments (precisely, curtailing it from \( 5 \times 10^{4} \) to \( 1 \times 10^{4} \)), and scalability concerning augmented populations (\( n = 16 \)) and increased number of hyperparameters (from 4 to 7).
In the realm of off-policy RL, we assessed resilience against hyperparameter boundary modifications, notably expanding the learning rate domain from \([10^{-5}, 10^{-3}]\) to \([10^{-5}, 10^{-2}]\).

\subsubsection{HPO Baselines} 
Random search (RS) serves as our foundational benchmark, courtesy of its assumption-agnostic approach, often culminating in asymptotically near-optimal performance \cite{Bergstra2012}. 
To augment the challenge, RS's initialization process was refined by sampling hyperparameters from partitioned grid intervals. 
PBT remains our principal comparative standard, maintaining configurations congruent with GPBT-PL. 
Specifically, in both PBT and GPBT-PL, agents positioned in the bottom quartile (representing slower learners) are supplanted by their counterparts from the top quartile (exemplifying rapid learners), with a resample probability pegged at 0.25. 
Additionally, our outcomes were juxtaposed with PB2 \cite{Parker-Holder2020a}, an enhancement of PBT that substitutes the random heuristic with Bayesian optimization.

\begin{table}[tbh]
\caption{\label{tab:PPO-50000}Best mean rewards across 7 seeds. The best-performing
algorithms are bolded. The last column presents the percentage of
performance difference between GPBT and PBT, where differences less
than 1\% are represented with $\approx$.}

\hfill{}\subfloat{\hfill{}%
\begin{tabular}{ccccccc}
\toprule 
Benchmarks & $n$ & RS & PB2 & PBT & GPBT-PL & vs. PBT\tabularnewline
\midrule 
BipedalWalker & 4 & 292 & \textbf{303} & 282 & 293 & +4\%\tabularnewline
Ant & 4 & 4283 & 5000 & 5347 & \textbf{5497} & +3\%\tabularnewline
HalfCheetah & 4 & 4834 & 4938 & 4911 & \textbf{5262} & +7\%\tabularnewline
InvertedDP & 4 & 8531 & \textbf{9356} & 9354 & 9274 & $\approx$\tabularnewline
Swimmer & 4 & 133 & 153 & 134 & \textbf{166} & +24\%\tabularnewline
Walker2D & 4 & 2267 & 1851 & 1800 & \textbf{2372} & +32\%\tabularnewline
\midrule 
BipedalWalker & 8 & 284 & 297 & 298 & \textbf{303} & +2\%\tabularnewline
Ant & 8 & 4508 & 5150 & 5705 & \textbf{6116} & +7\%\tabularnewline
HalfCheetah & 8 & 4842 & 5152 & 5292 & \textbf{5463} & +3\%\tabularnewline
InvertedDP & 8 & 9185 & 9340 & 9340 & \textbf{9356} & $\approx$\tabularnewline
Swimmer & 8 & 155 & 135 & 157 & \textbf{168} & +7\%\tabularnewline
Walker2D & 8 & 2776 & 2528 & 2855 & \textbf{3047} & +7\%\tabularnewline
\bottomrule
\end{tabular}\hfill{}}\hfill{}
\end{table}

\begin{figure*}[tbh]
\hfill{}\subfloat[\label{fig:bipedal_4_50000}BipedalWalker (4)]{\includegraphics[scale=0.18]{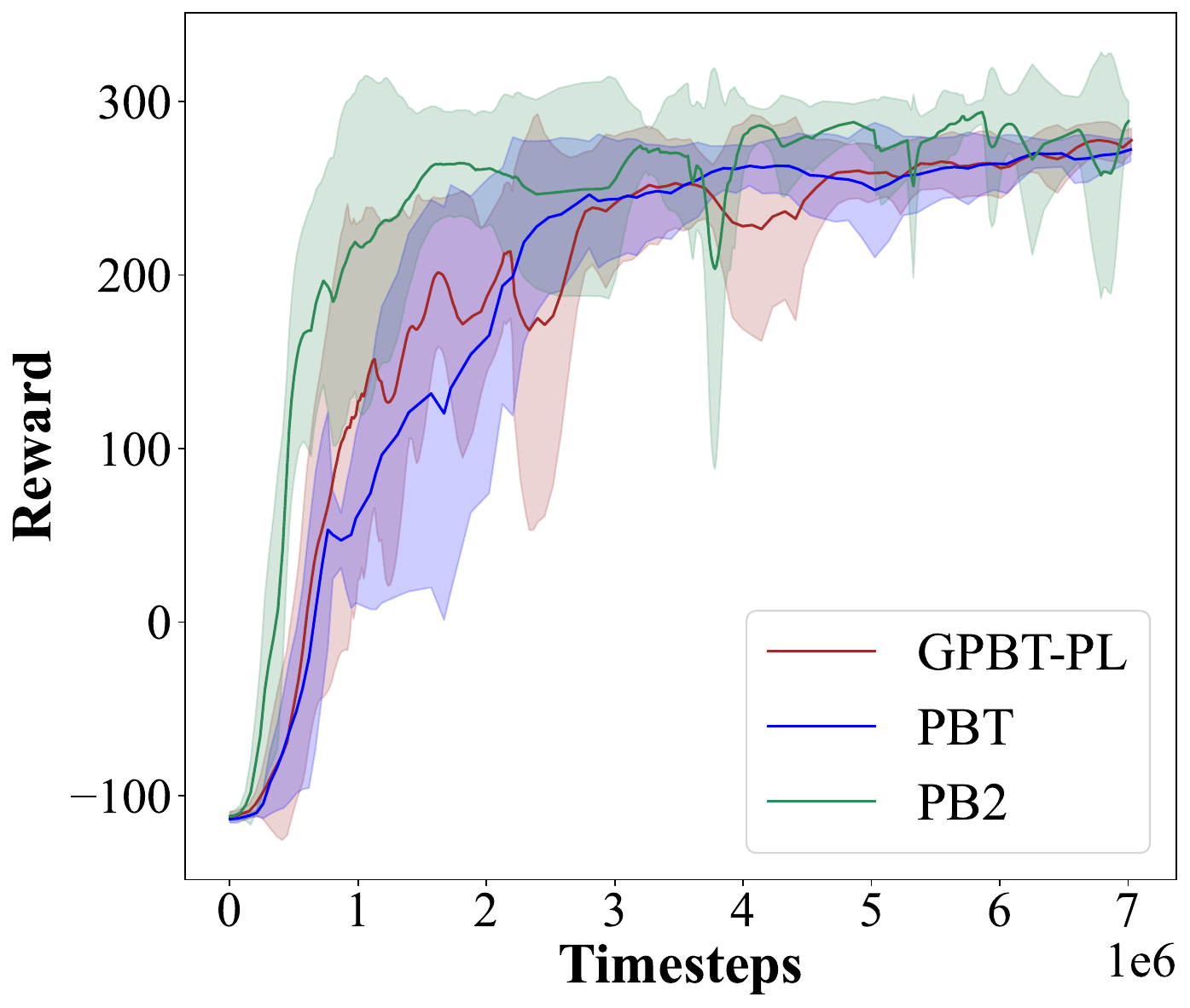}

}\hfill{}\subfloat[\label{fig:ant_4_50000}Ant (4)]{\includegraphics[scale=0.18]{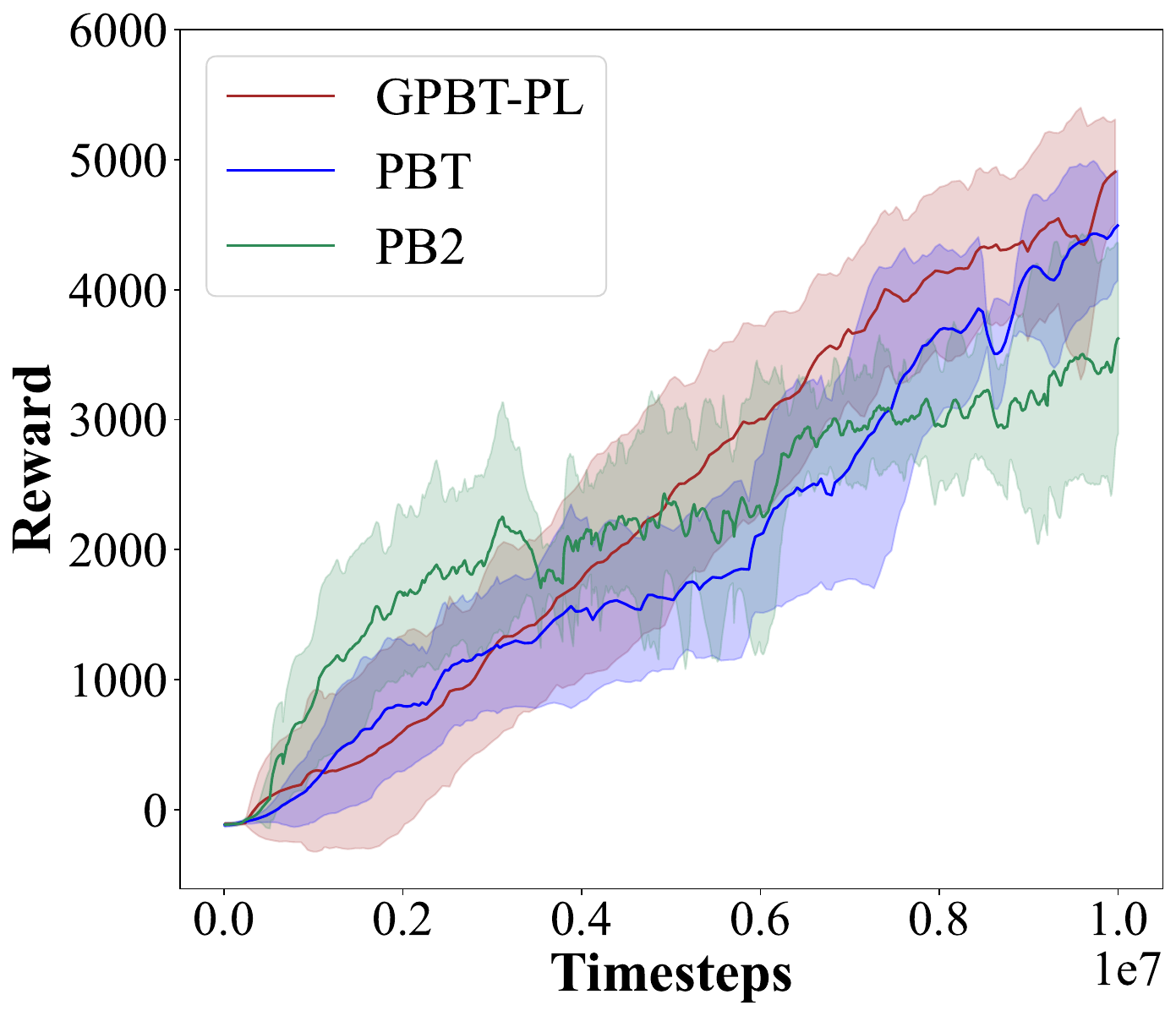}

}\hfill{}\subfloat[\label{fig:halfcheetah_4_50000}HalfCheetah (4)]{\includegraphics[scale=0.18]{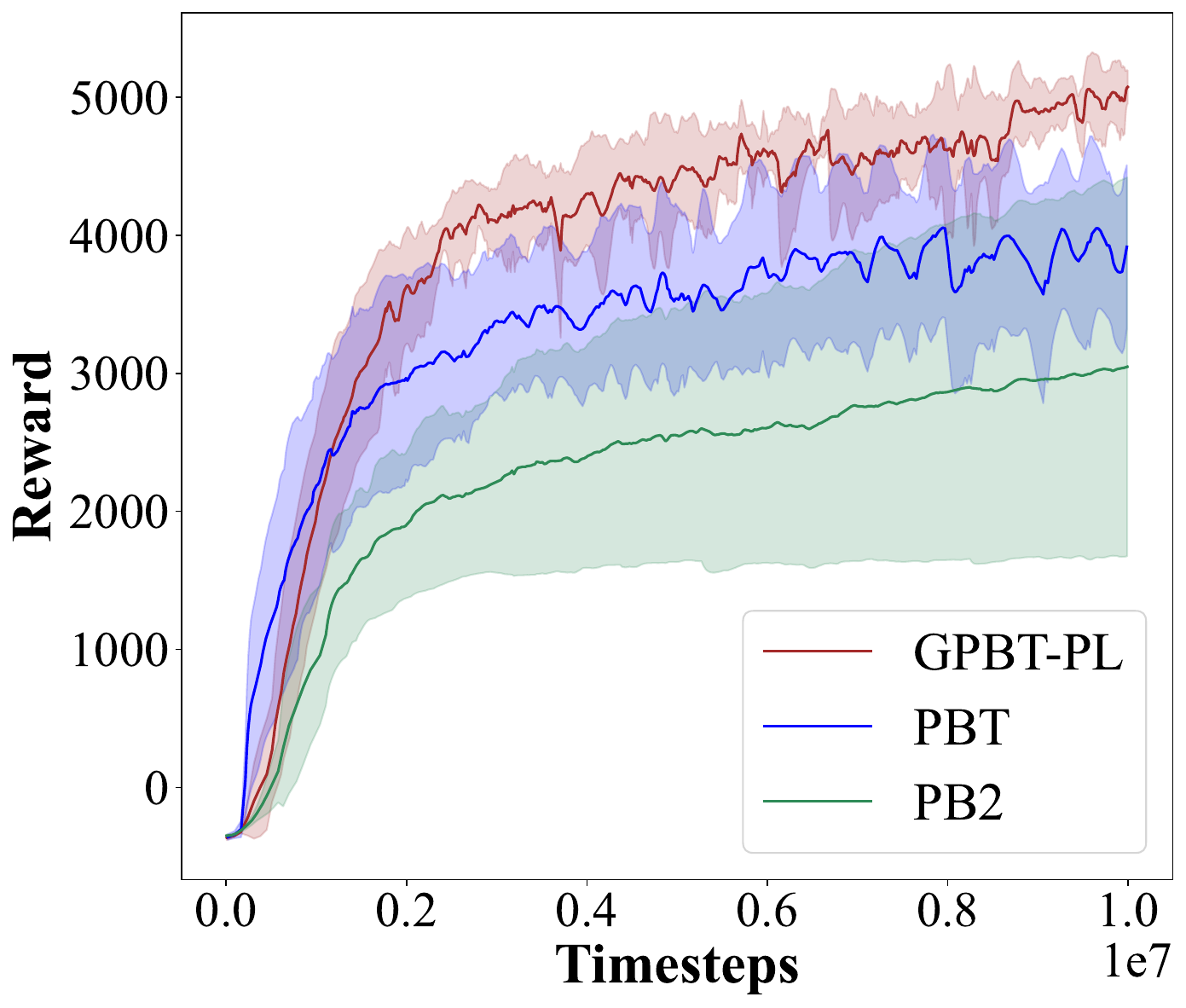}

}\hfill{}\subfloat[\label{fig:invertedDP_4_50000}InvertedDoublePendulum (4)]{\includegraphics[scale=0.18]{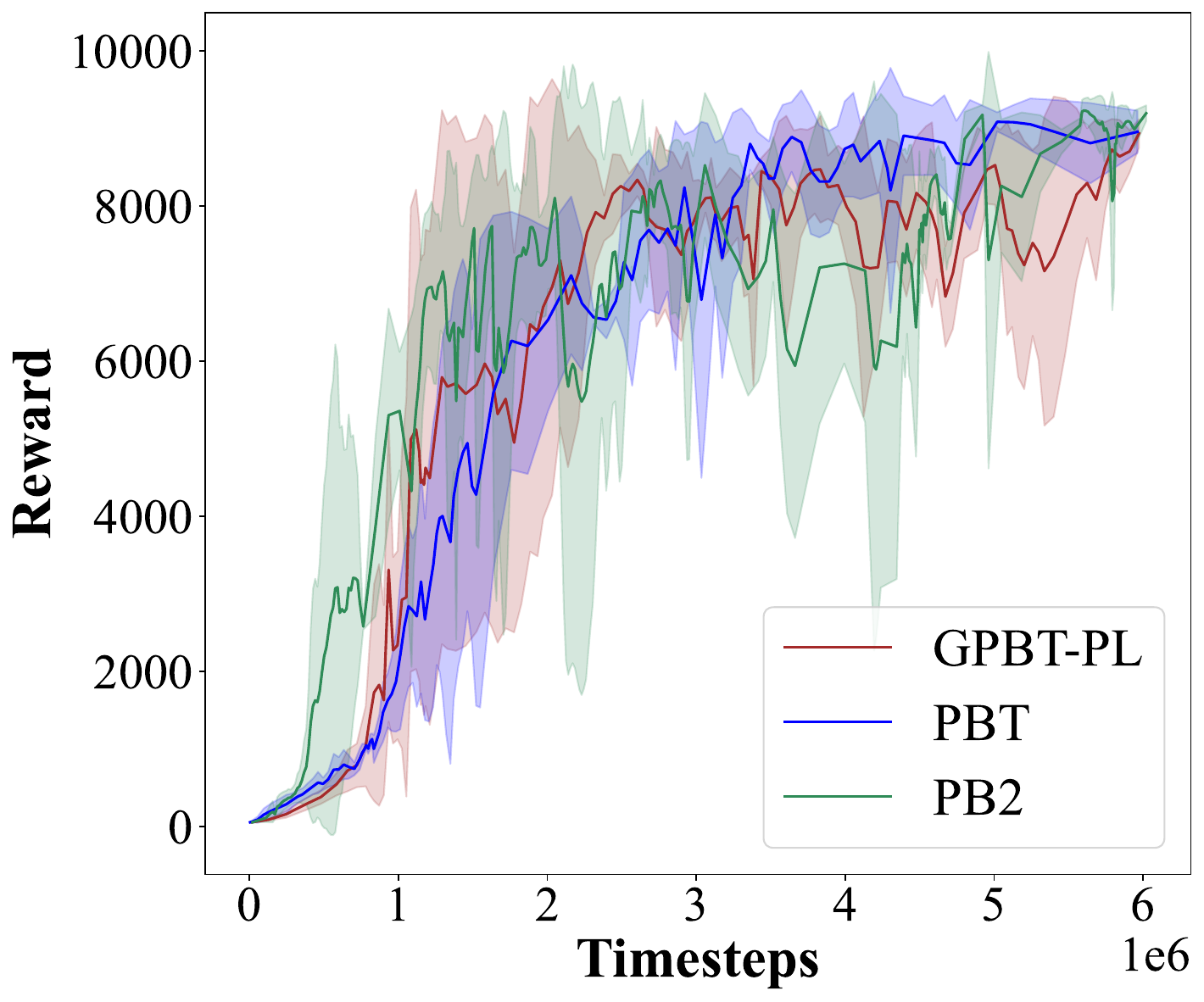}

}\hfill{}

\hfill{}\subfloat[\label{fig:swimmer_4_50000}Swimmer (4)]{\includegraphics[scale=0.18]{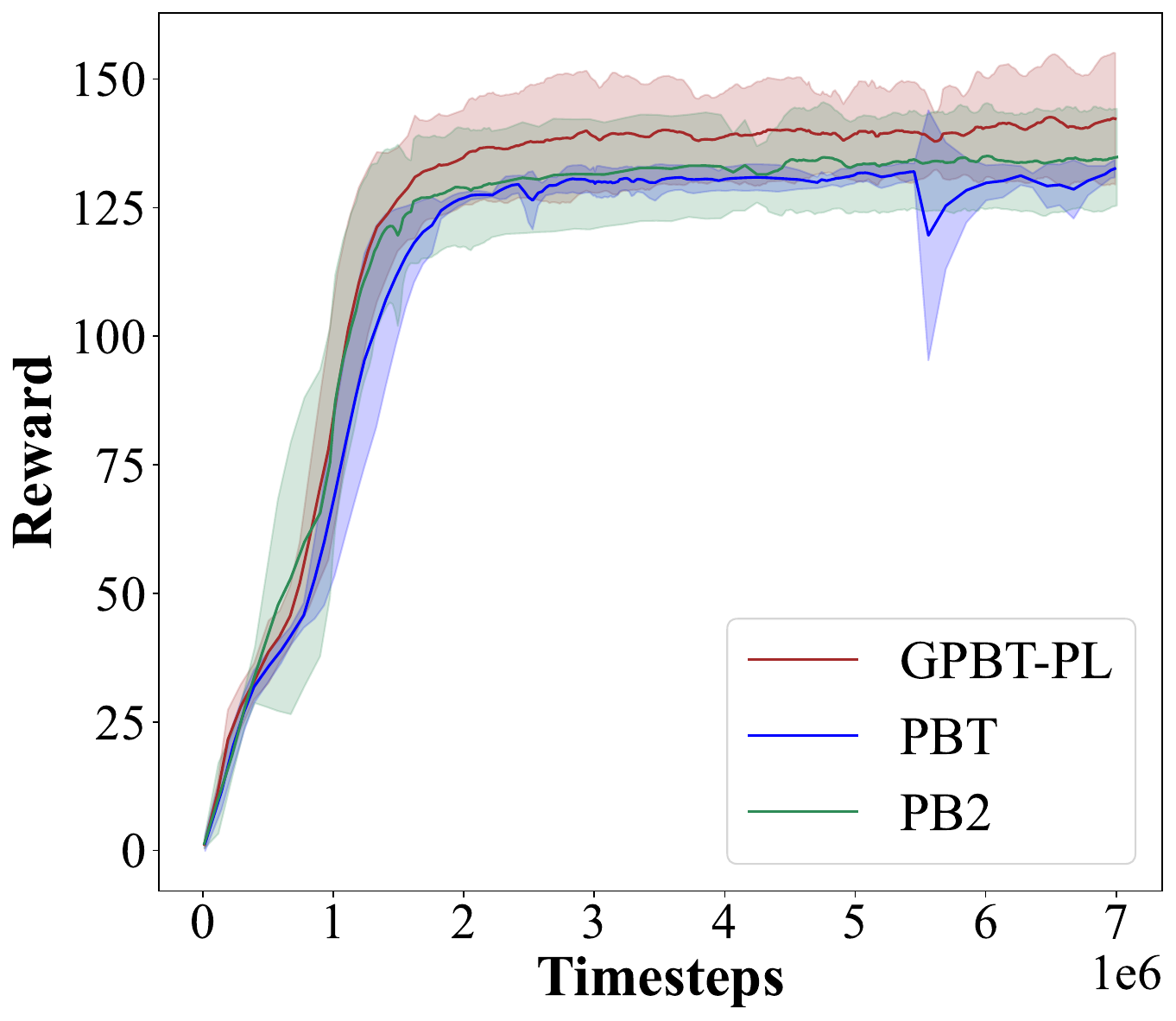}

}\hfill{}\subfloat[\label{fig:walker2d_4_50000}Walker2D (4)]{\includegraphics[scale=0.18]{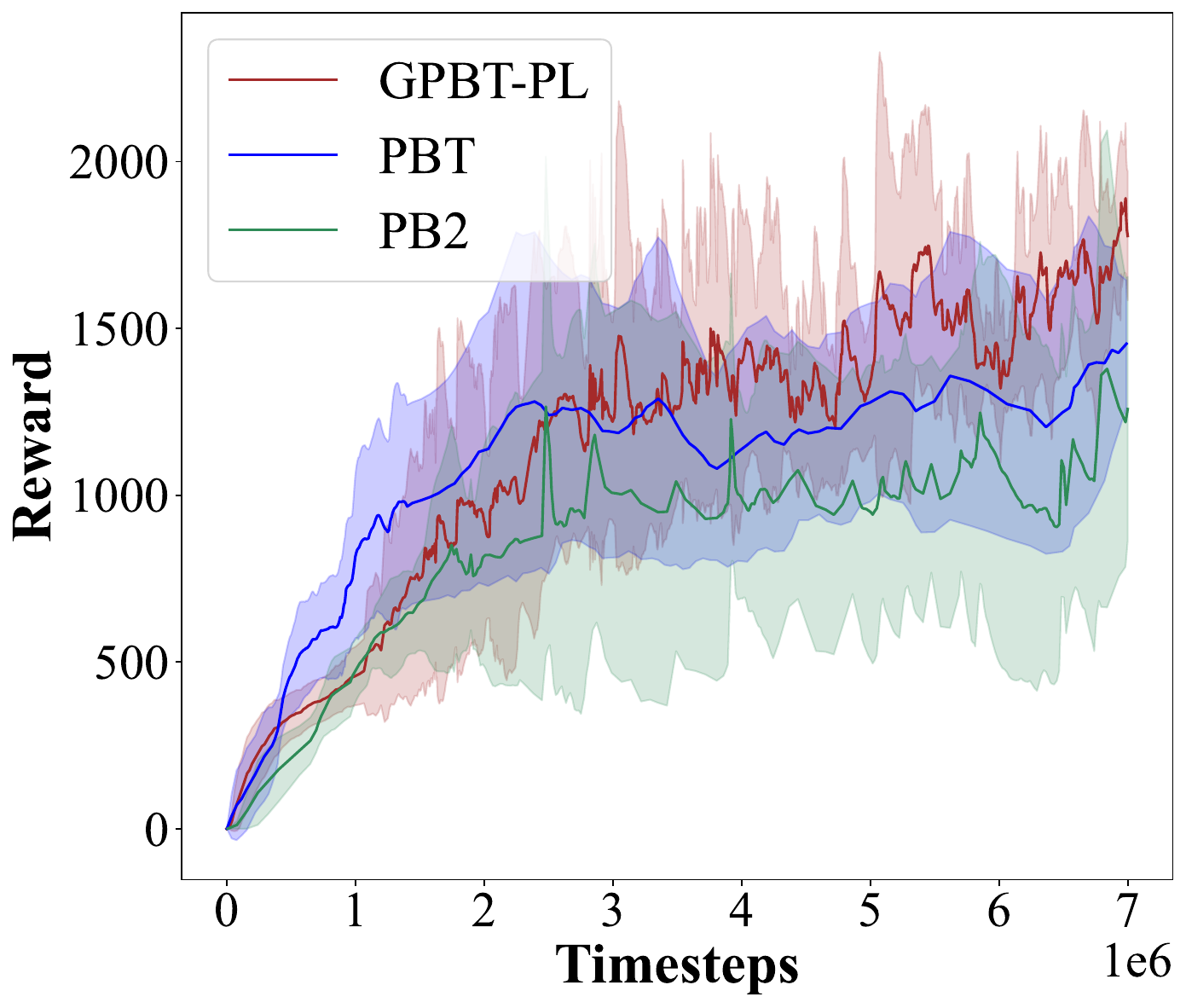}

}\hfill{}\subfloat[\label{fig:bipedal_8_50000}BipedalWalker (8)]{\includegraphics[scale=0.18]{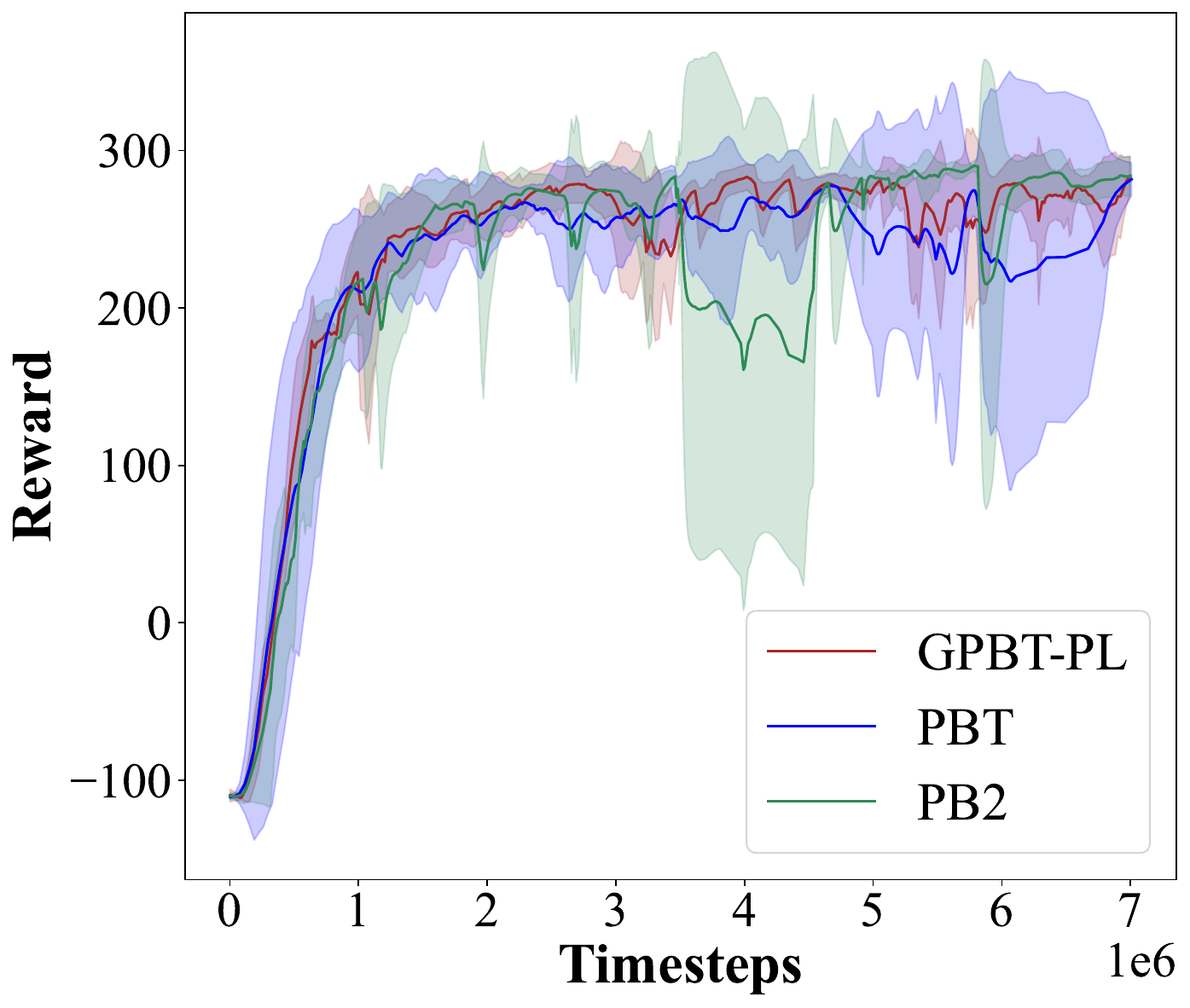}

}\hfill{}\subfloat[\label{fig:ant_8_50000}Ant (8)]{\includegraphics[scale=0.18]{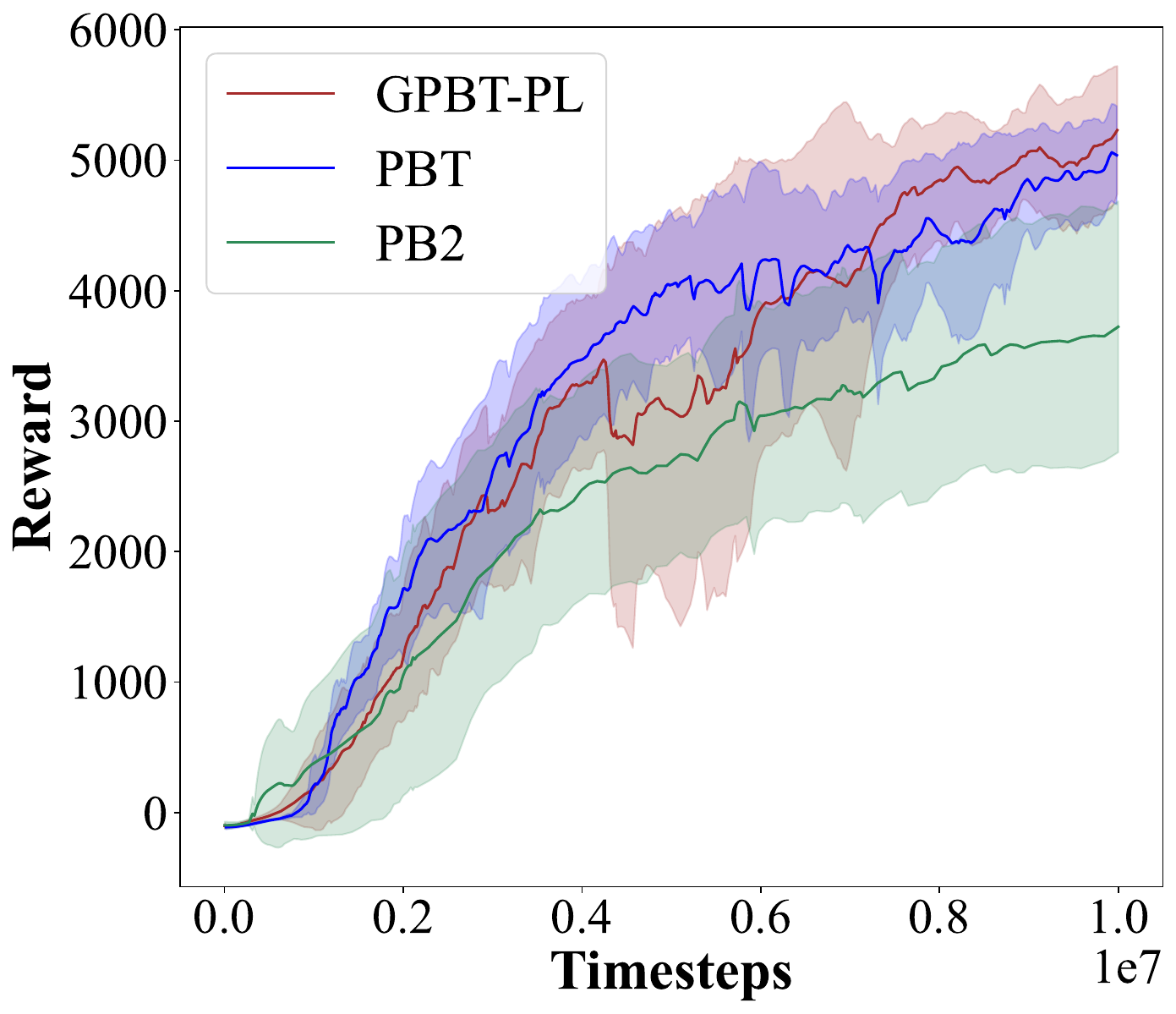}

}\hfill{}

\hfill{}\subfloat[\label{fig:halfcheetah_8_50000}HalfCheetah (8)]{\includegraphics[scale=0.18]{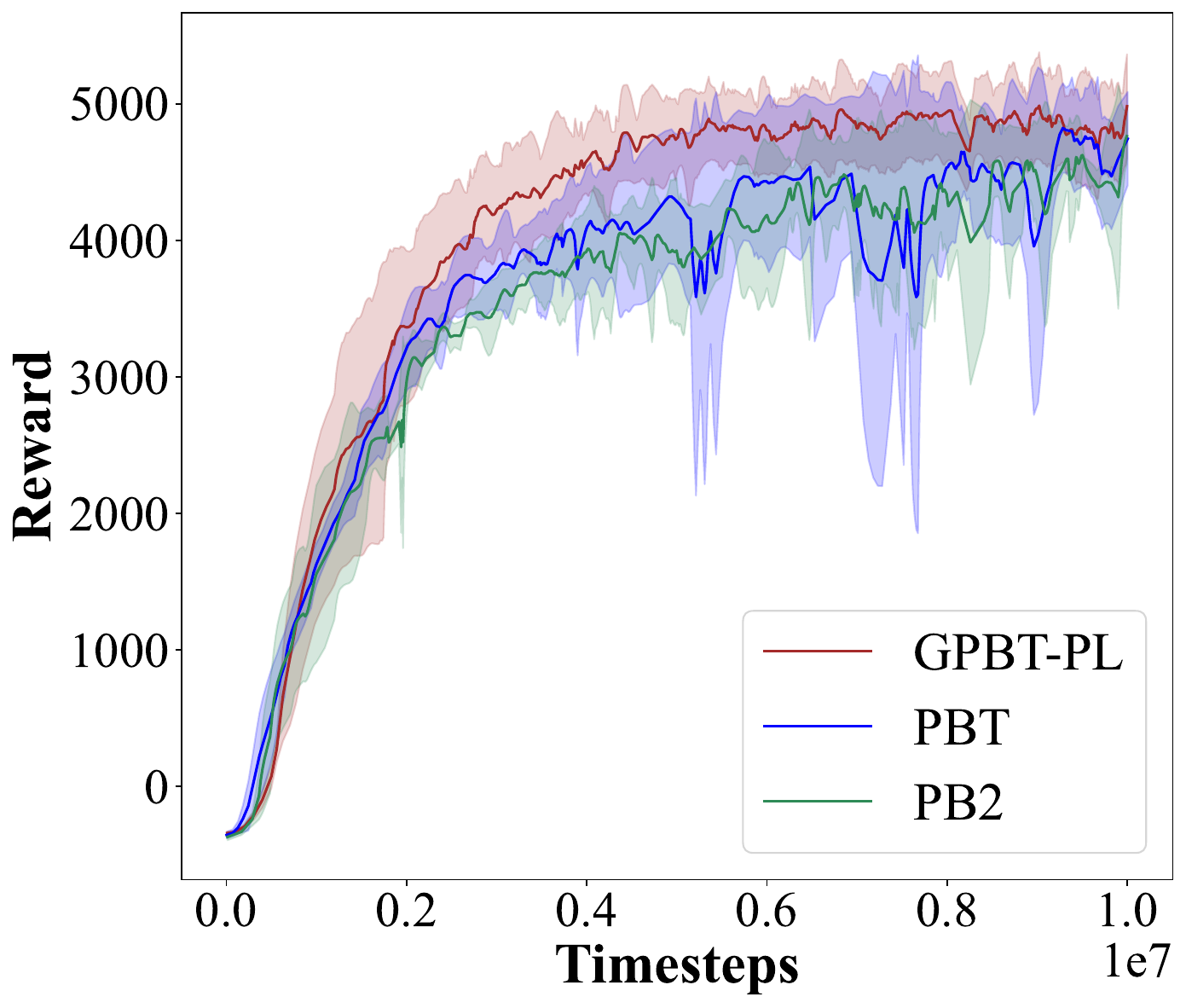}

}\hfill{}\subfloat[\label{fig:invertedDP_8_50000}InvertedDoublePendulum (8)]{\includegraphics[scale=0.18]{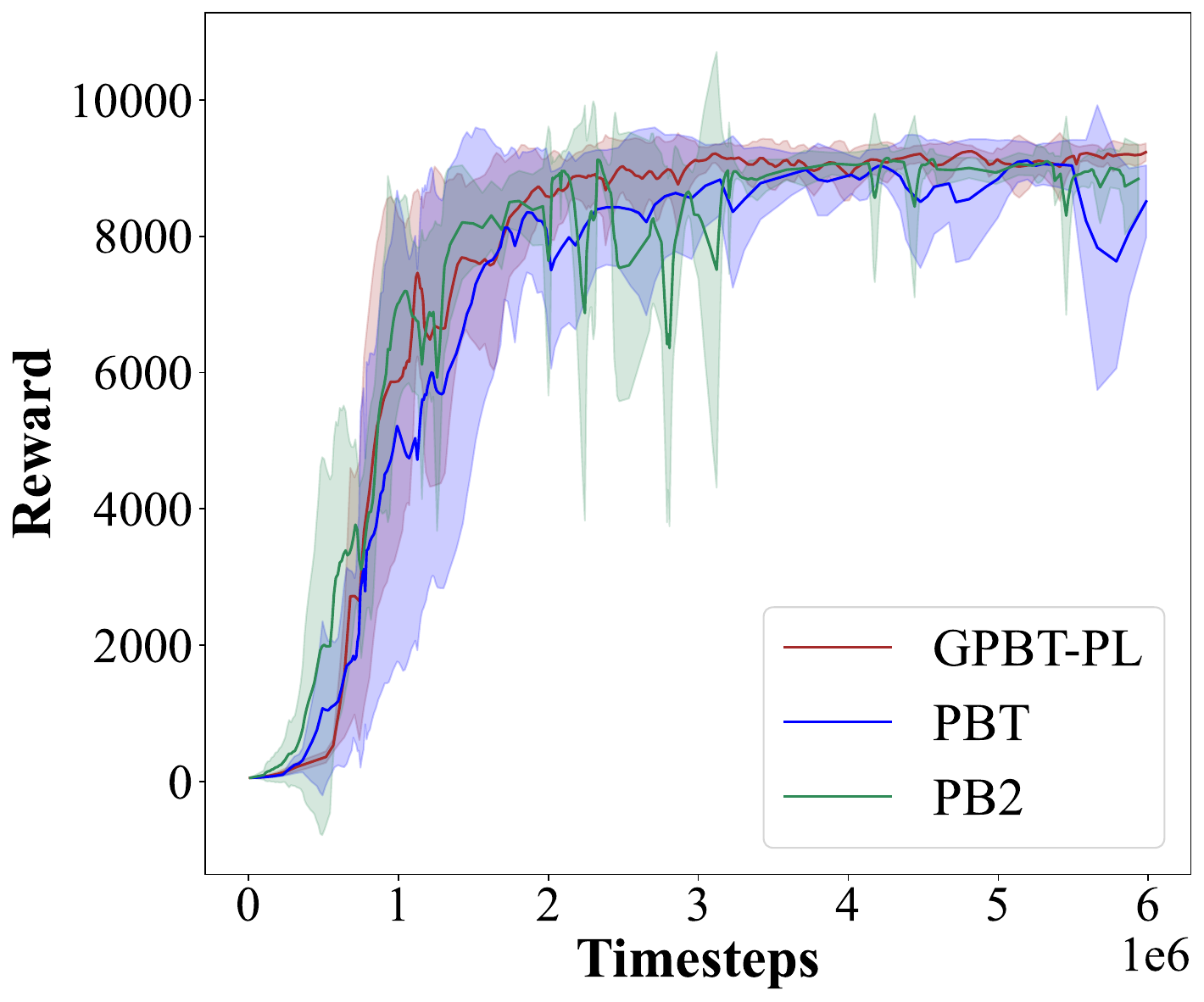}

}\hfill{}\subfloat[\label{fig:swimmer_8_50000}Swimmer (8)]{\includegraphics[scale=0.18]{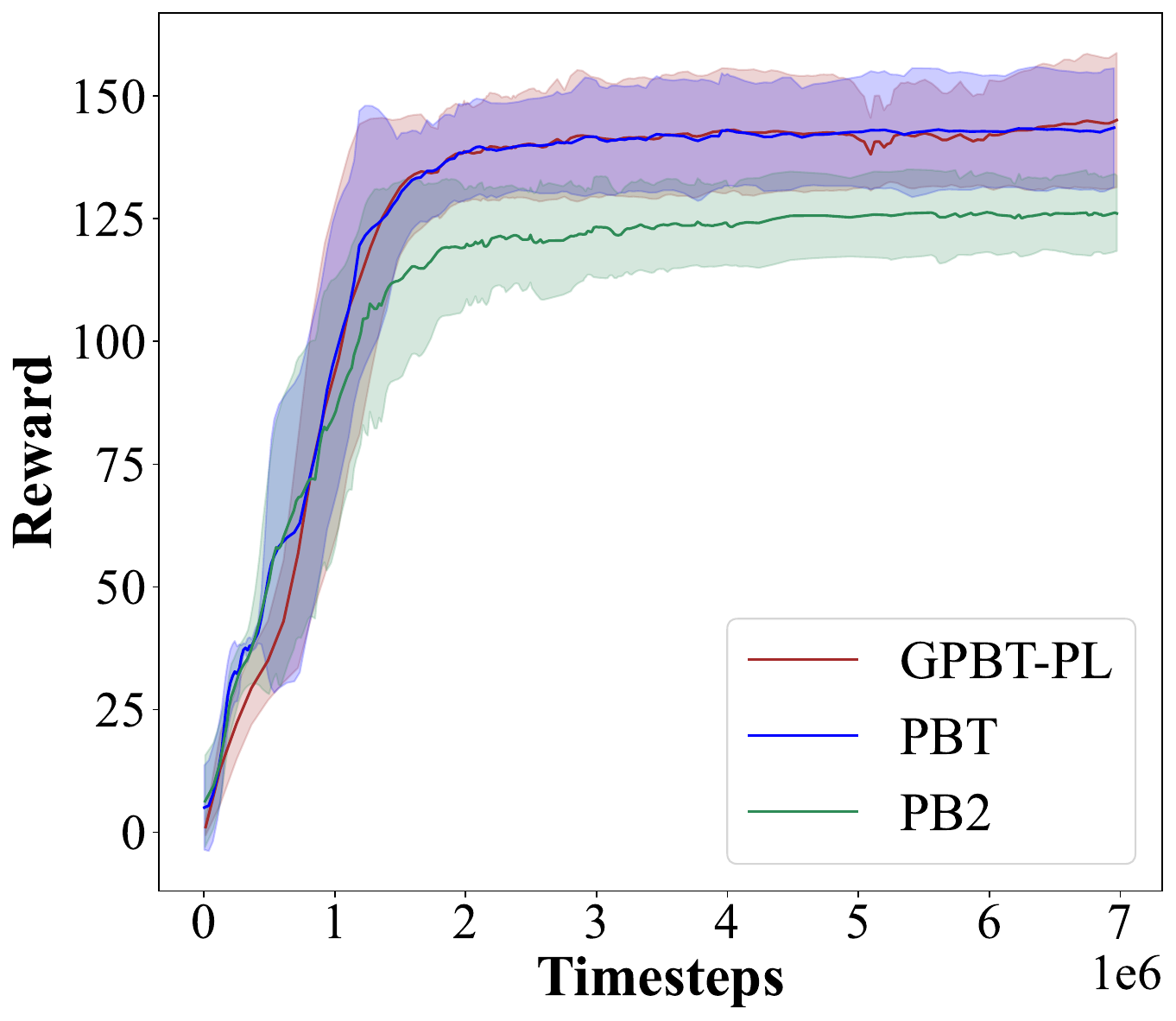}

}\hfill{}\subfloat[\label{fig:walker2d_8_50000}Walker2D (8)]{\includegraphics[scale=0.18]{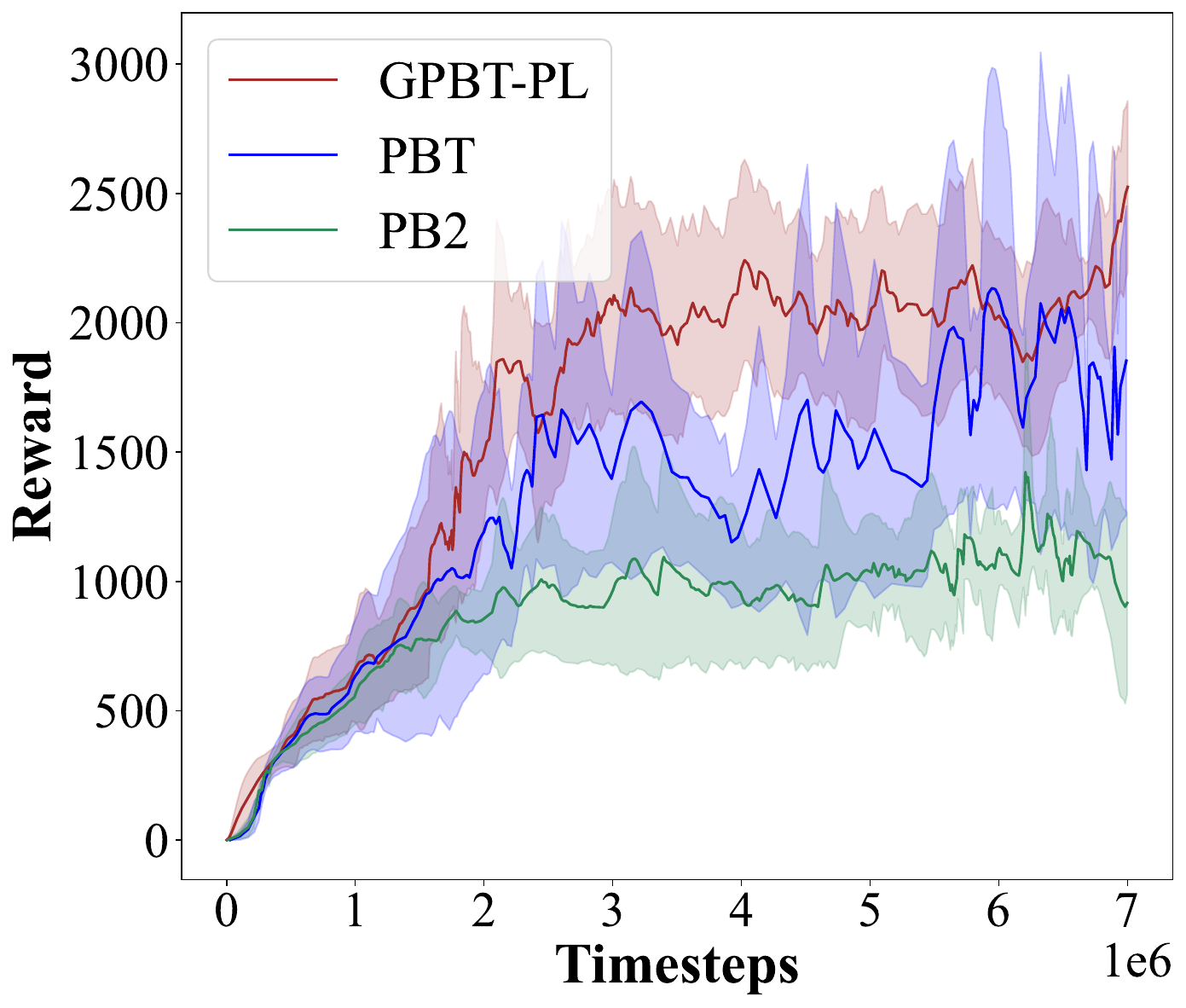}

}\hfill{}

\caption{\label{fig:PPO_50000}Training curves for six OpenAI Gym benchmarks using populations of 4 and 8 agents with GPBT-PL, PBT, and PB2. Thick lines represent the average of the best mean rewards over 7 seeds, with shaded regions denoting the standard deviation. Brackets specify the population size, and the perturbation interval is set to \(5 \times 10^{4}\).}
\end{figure*}

\begin{figure*}[tbh]
\hfill{}\subfloat[\label{fig:ant_4_5000_}Ant (4)]{\includegraphics[scale=0.18]{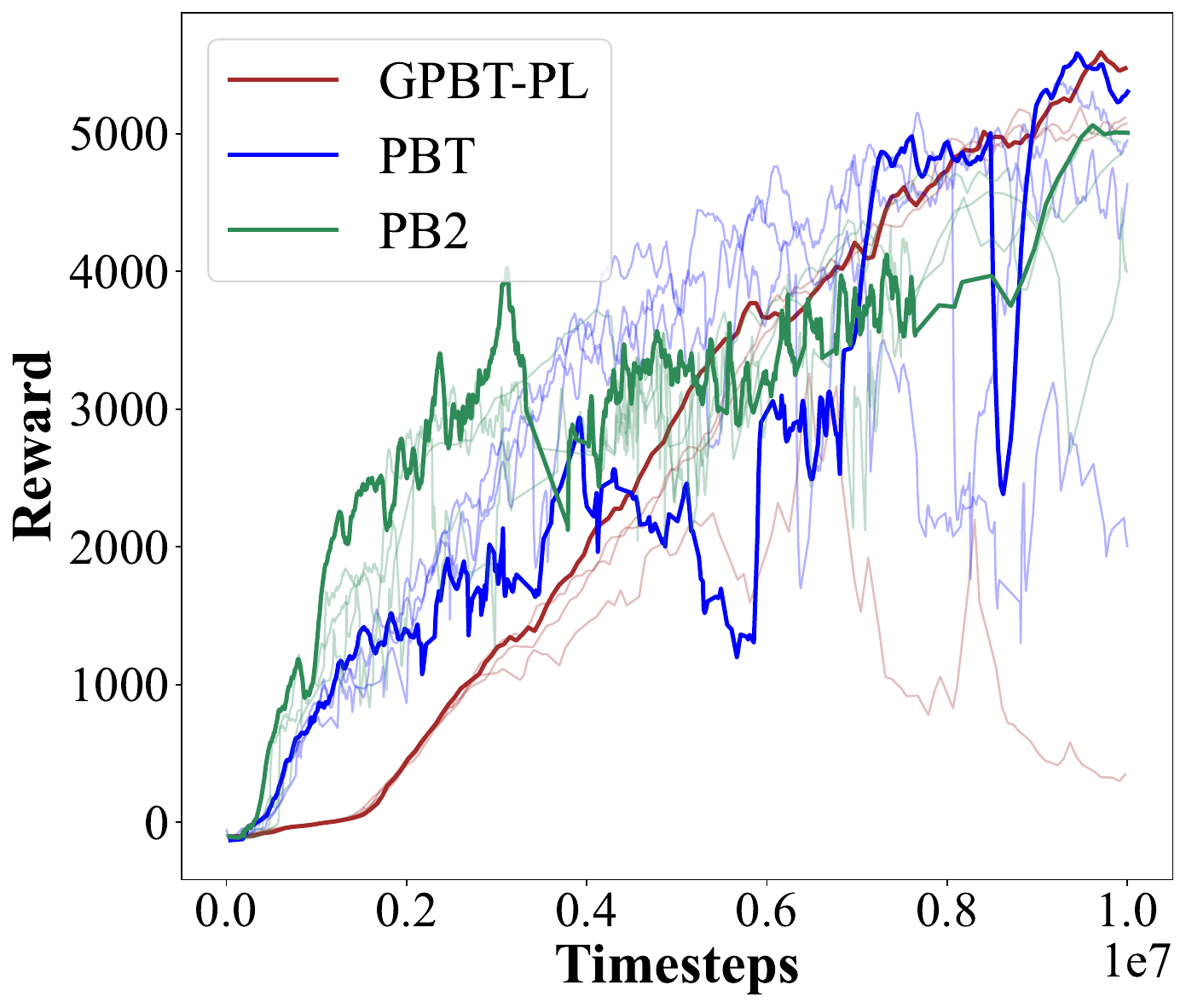}

}\hfill{}\subfloat[\label{fig:ant_8_5000_}Ant (8)]{\includegraphics[scale=0.18]{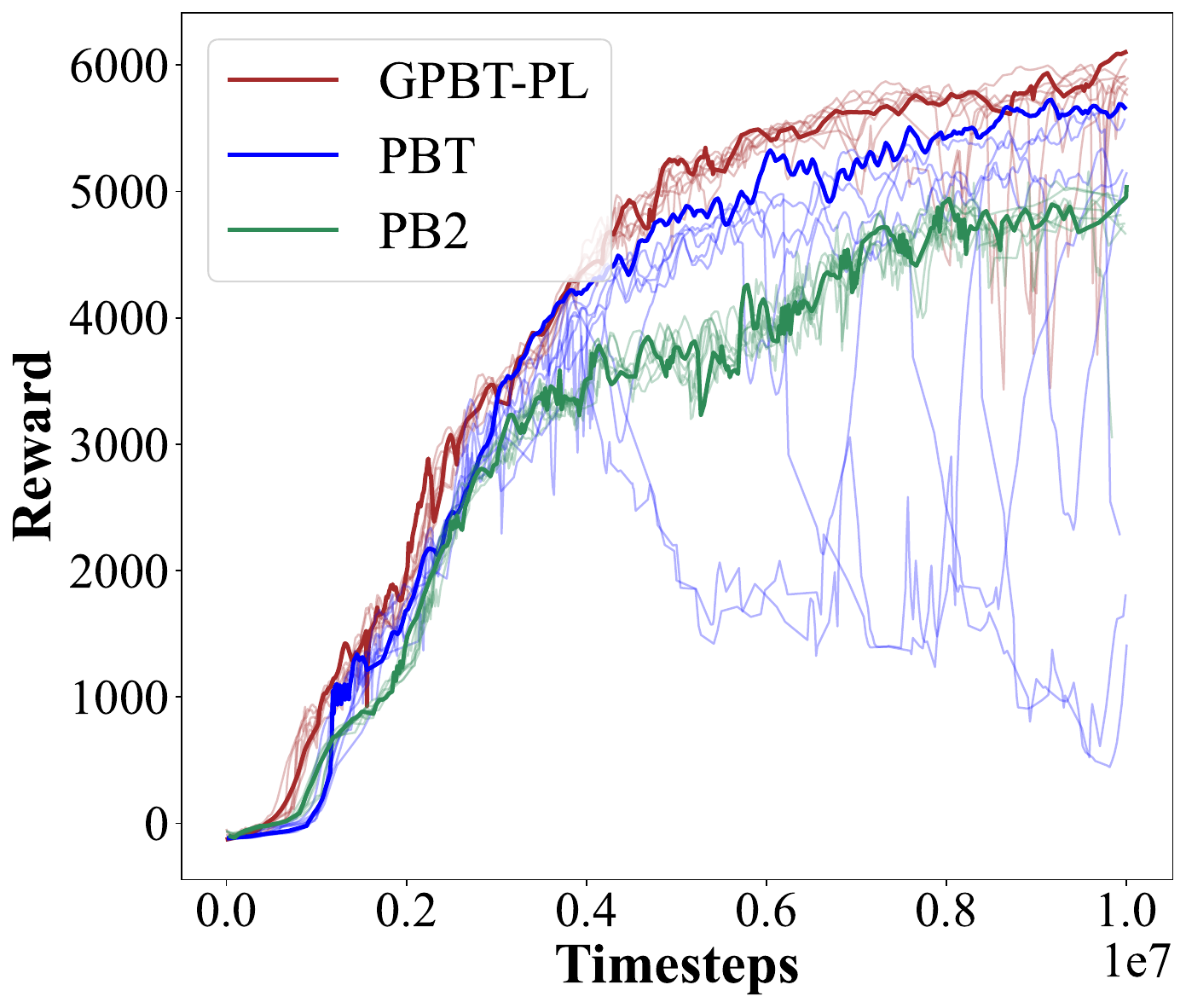}

}\hfill{}\subfloat[\label{fig:halfcheetah_4_50000_}HalfCheetah (4)]{\includegraphics[scale=0.18]{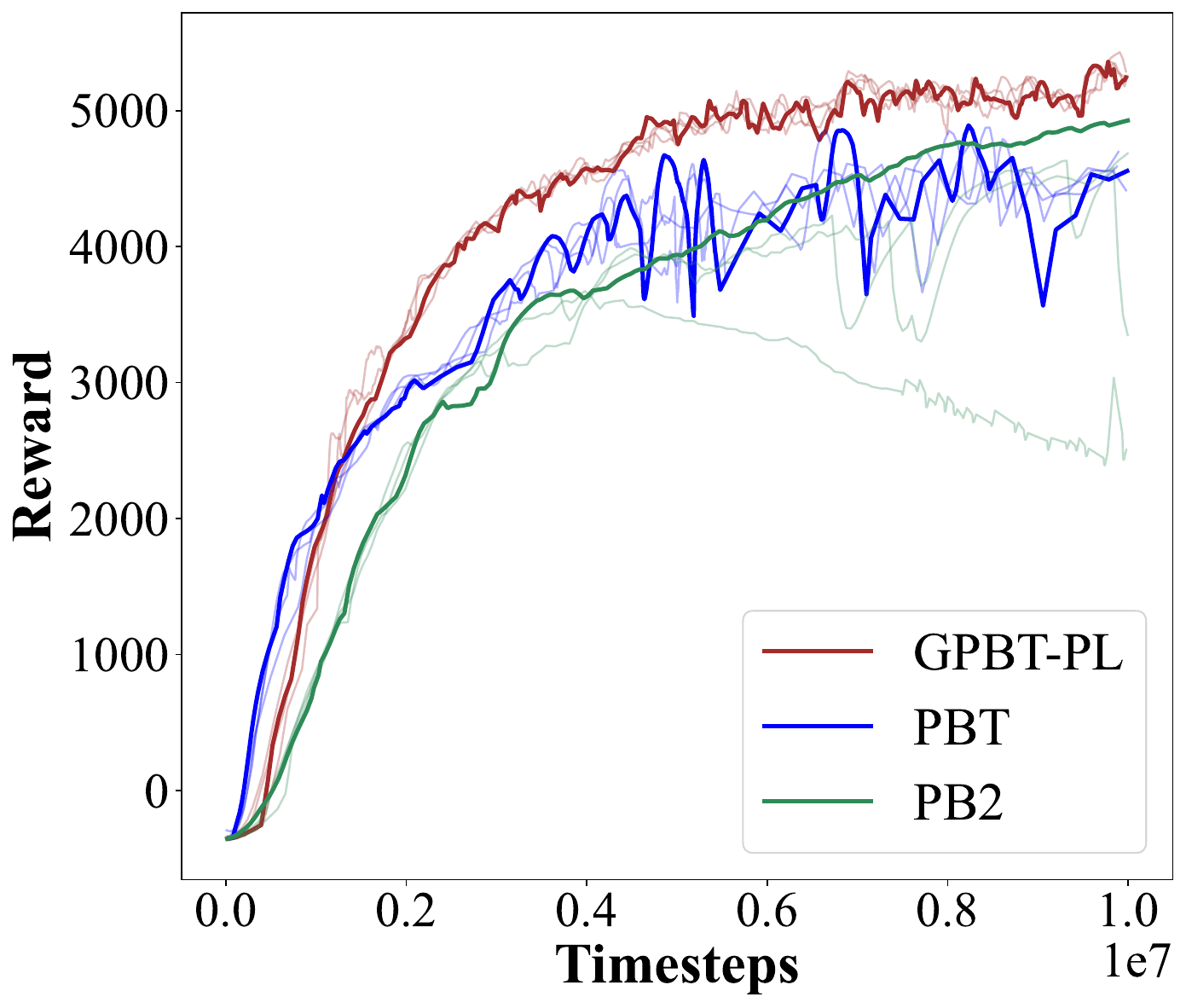}

}\hfill{}\subfloat[\label{fig:halfcheetah_8_50000_}HalfCheetah (8)]{\includegraphics[scale=0.18]{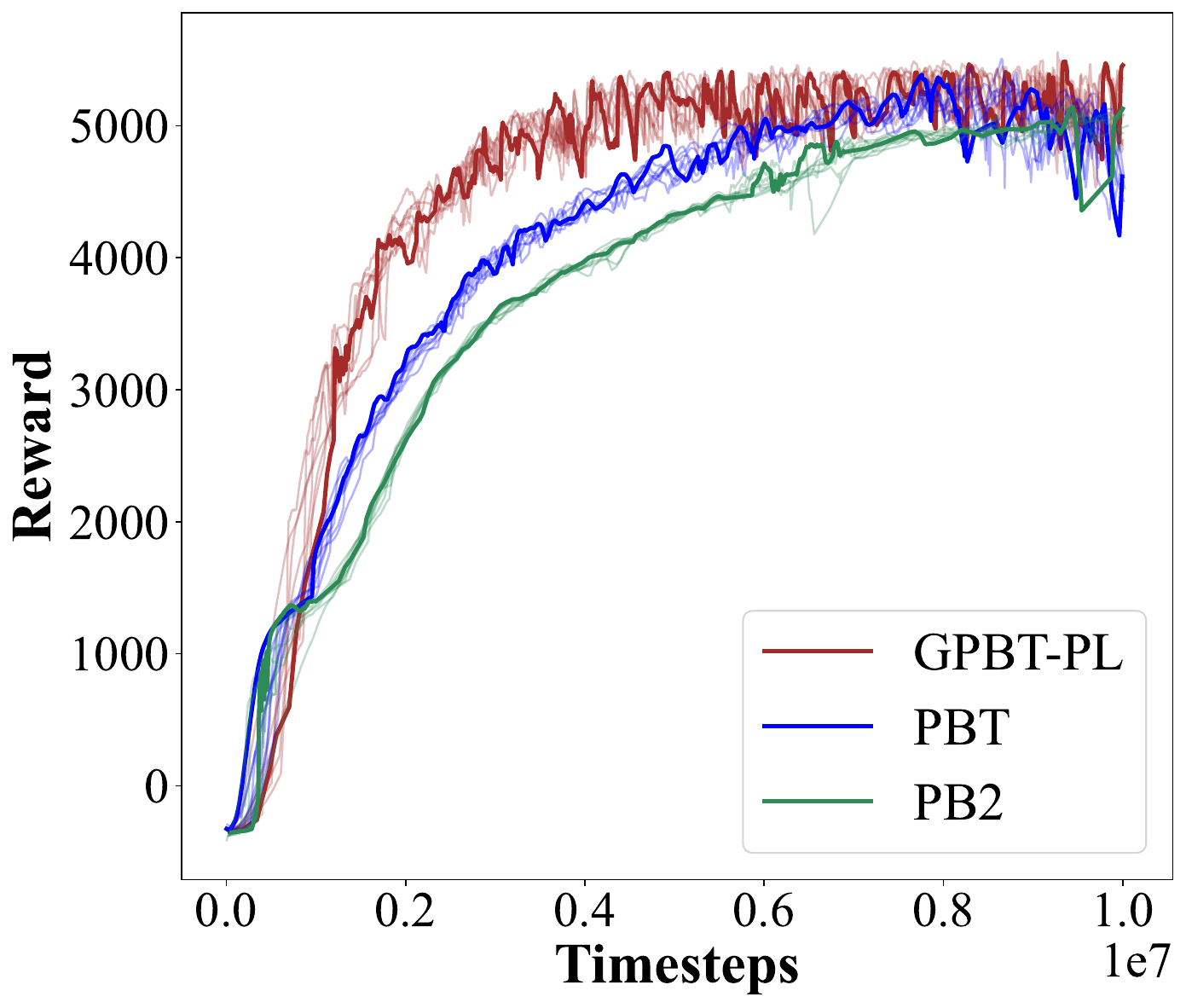}

}\hfill{}

\hfill{}\subfloat[\label{fig:ant_4_5000_time}Ant (4)]{\includegraphics[scale=0.18]{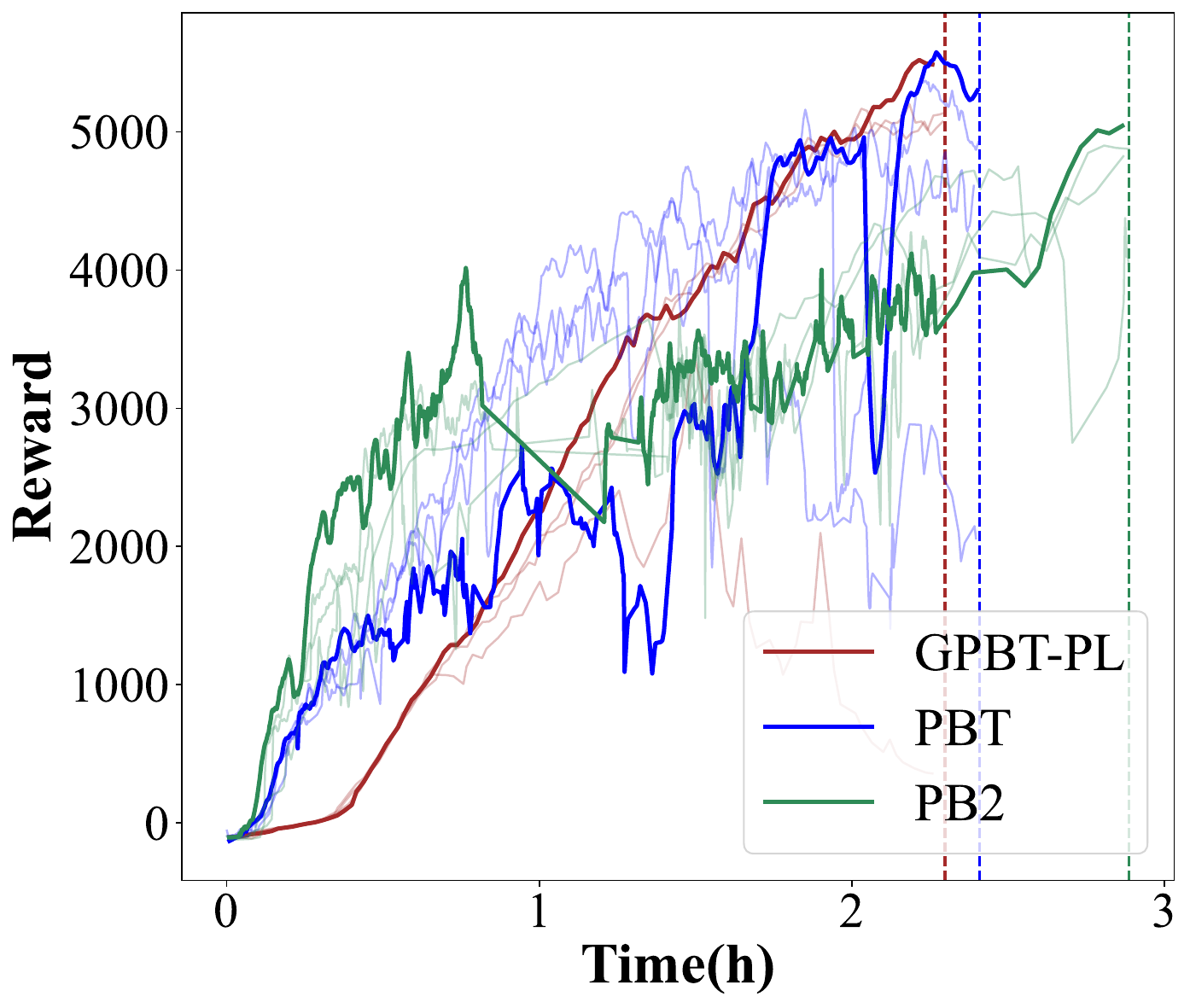}

}\hfill{}\subfloat[\label{fig:ant_8_5000_time}Ant (8)]{\includegraphics[scale=0.18]{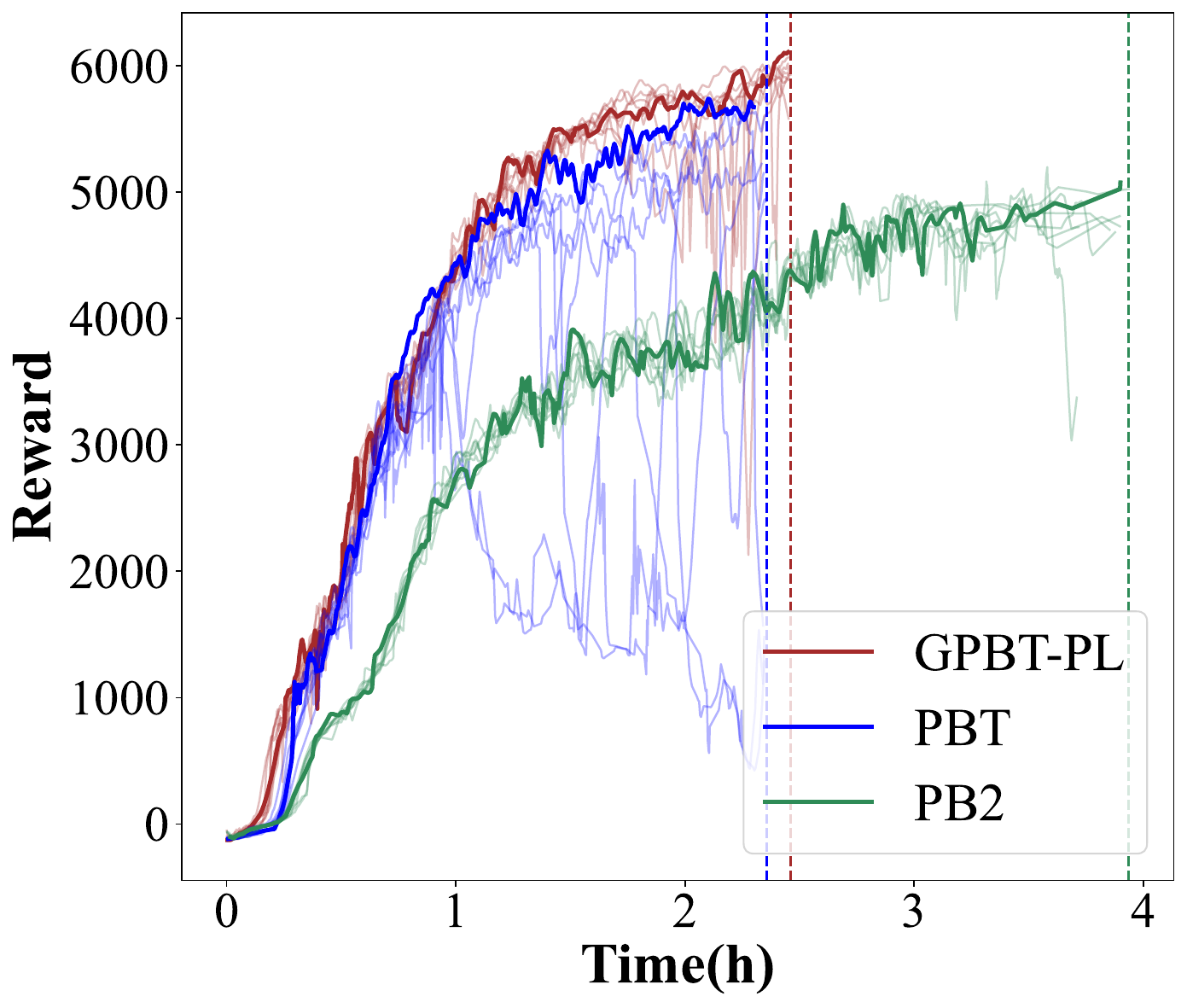}

}\hfill{}\subfloat[\label{fig:halfcheetah_4_50000_time}HalfCheetah (4)]{\includegraphics[scale=0.18]{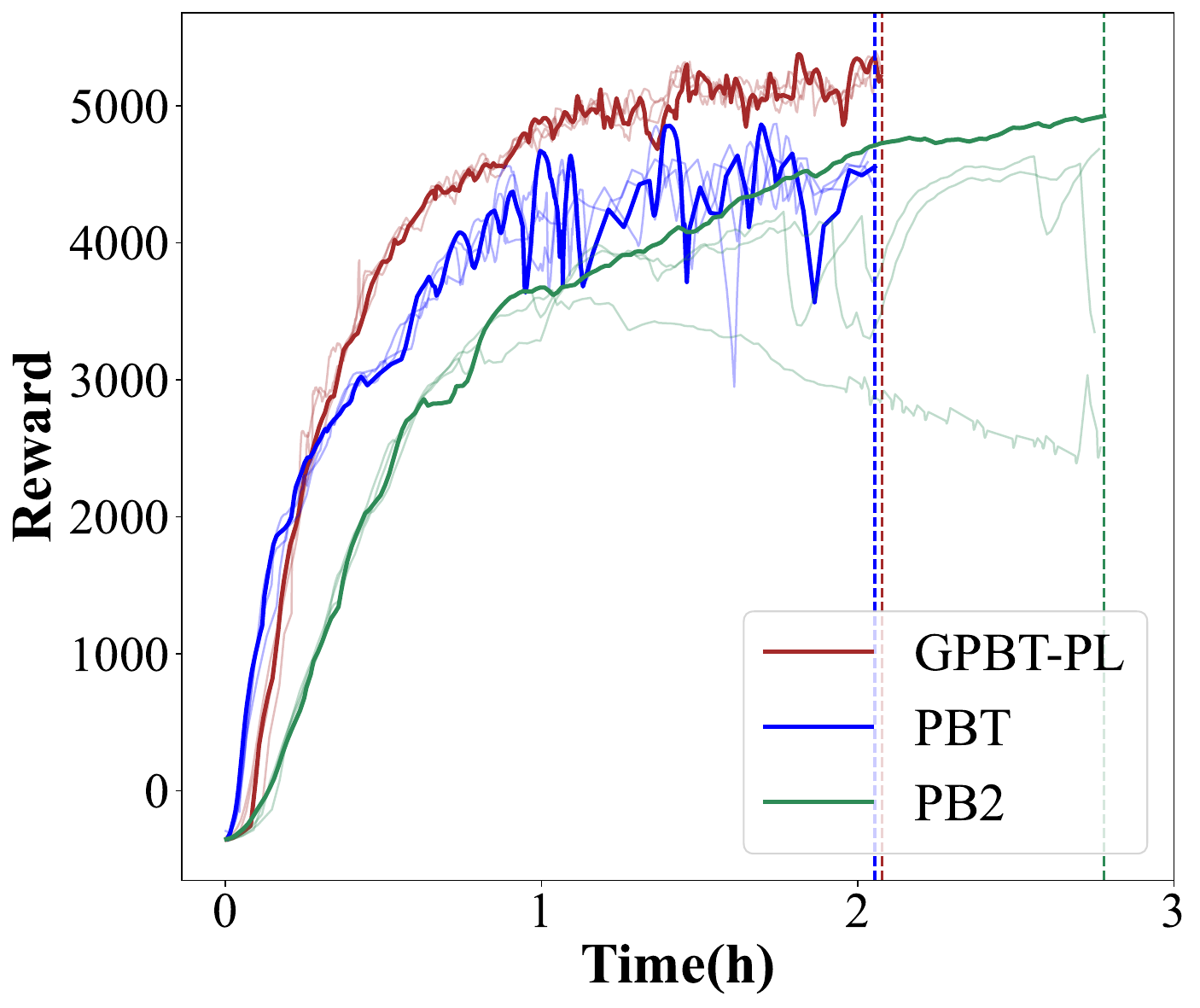}

}\hfill{}\subfloat[\label{fig:halfcheetah_8_50000_time}HalfCheetah (8)]{\includegraphics[scale=0.18]{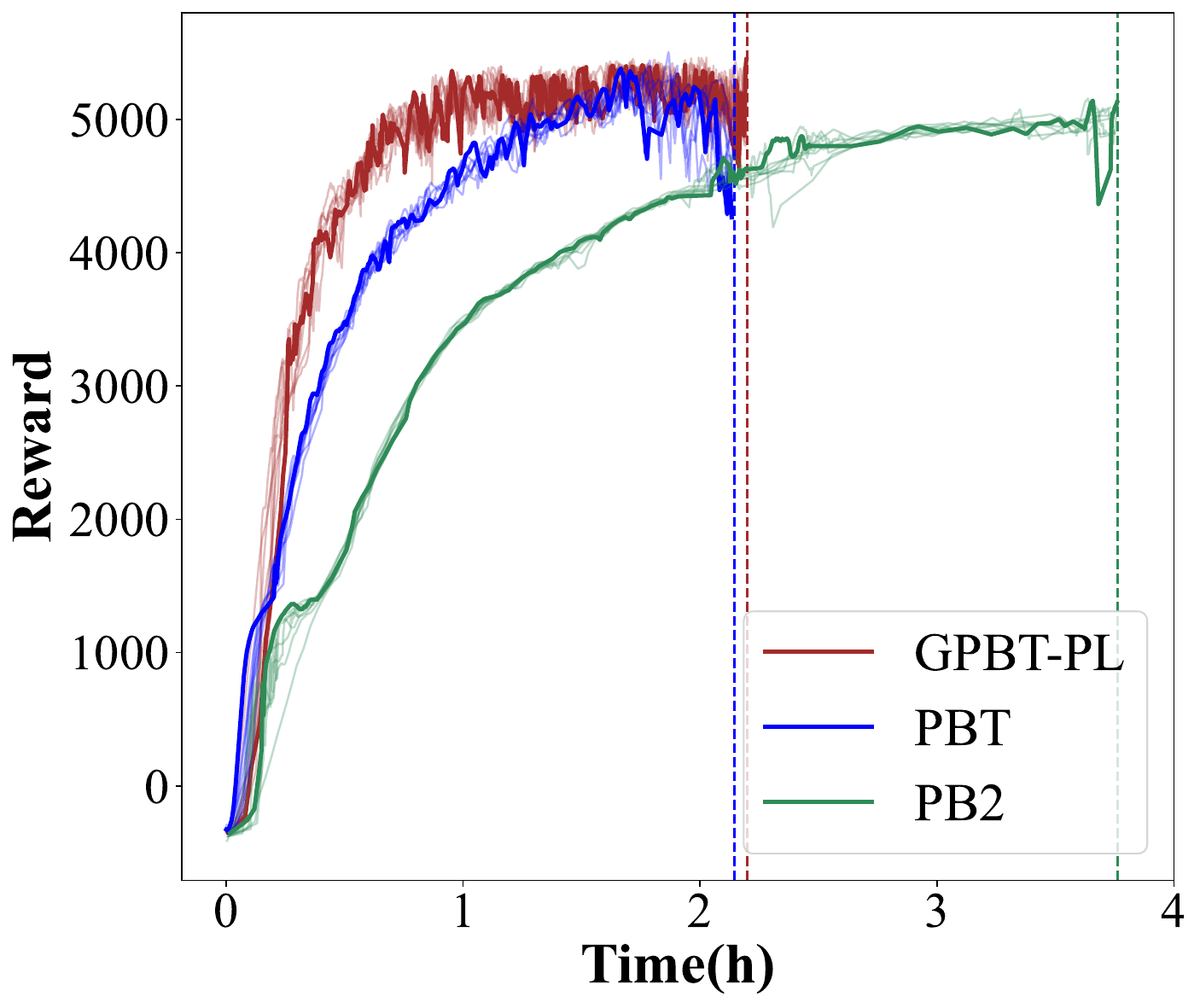}

}\hfill{}

\caption{\label{fig:PPO_50000_evolution}Training curves for Ant and HalfCheetah using populations of 4 and 8 agents with GPBT-PL, PBT, and PB2. (a)-(d) take timesteps as the x-axis and (c)-(h) take time (in hours) as the x-axis. Thick lines are the best-performing members of the population of each HPO method, with faint lines representing each member. Brackets specify the population size, and the perturbation interval is set to \(5 \times 10^{4}\).
}
\end{figure*}

\subsection{On-Policy Reinforcement Learning \label{subsec:On-Policy-RL}}

In experiments for HPO in on-policy RL, we aimed to optimize four hyperparameters: batch size, GAE \(\lambda\), PPO clip \(\epsilon\), and learning rate \(\eta\) for the PPO algorithm. 
The chosen continuous control tasks were BipedalWalker, Ant, HalfCheetah, InvertedDoublePendulum, Swimmer, and Walker2D. We employed population sizes of 4 and 8 with a perturbation interval of \(5 \times 10^{4}\). 
Experiments concluded when all algorithms achieved stable convergence, ensuring a fair comparison, contrasting with the approach in the PB2 paper where a fixed number of timesteps defined the stopping criterion. 
\tablename~\ref{tab:PPO-50000} tabulates the best mean rewards.
\figurename~\ref{fig:PPO_50000} depicts their mean and standard deviation across all seeds, with population sizes annotated within brackets, and \figurename~\ref{fig:PPO_50000_evolution} further shows the population evolution process along the training timesteps and time (in hours) for the top-performing seed.

Predominantly, GPBT-PL showcased superior performance over PBT. In 75\% of the scenarios, GPBT-PL surpassed other algorithms in realizing the performance ceiling. For the remainder, GPBT-PL's performance closely mirrored the top-performing algorithms.

For smaller populations (\(n=4\)), GPBT-PL registered marked enhancements of 24\% and 32\% over PBT for the Swimmer and Walker2D tasks, respectively. 
The efficacy of GPBT-PL was notably higher with smaller populations compared to larger ones (\(n=8\)).
 Conversely, PBT's performance lagged behind RS for BipedalWalker and Walker2D but improved upon increasing the population size. This trend underscores PBT's dependency on extensive computational resources, a sentiment echoed in the original PBT publication \cite{Jaderberg2017}. PBT's aggressive strategy, which results in the early dismissal of promising candidates, may underpin its performance limitations. 
 Such premature decisions become especially detrimental with smaller populations.

All four HPO strategies exhibited commendable performance on the InvertedDoublePendulum task, potentially attributable to the task's inherent simplicity, as noted in the PB2 study \cite{Parker-Holder2020a}. 
While PB2's performance was subpar to RS for Walker2D (4) and Swimmer (8), it outshone the other three methods on BipedalWalker (4) and InvertedDoublePendulum. 
This observation suggests PB2's aptitude for less complex tasks. 
However, PB2's reliance on Bayesian optimization for hyperparameter generation is computationally intensive, particularly with increasing population sizes, as shown in (e)-(h) of \figurename~\ref{fig:PPO_50000_evolution}. Yet, PBT and GPBT-PL exhibited comparable time efficiencies, with neither approach showing an increase in time consumption as the population size expanded. Additionally, PB2 grapples with the exploration-exploitation dilemma during hyperparameter generation, amplifying the intricacies of the tuning challenge.

In summary, GPBT-PL consistently delivered promising outcomes across both small and large populations, recording impressive rewards on challenging tasks like Ant and Walker2D.
Based on the visualization of population evolution, the performance of GPBT-PL exhibits a gradual increase in the initial stages, followed by a rapid ascent in the middle and later stages. This is attributed to GPBT-PL's ability to preserve late bloomers, thereby maintaining superior global search capability.

\begin{table}[tbh]
\caption{\label{tab:IMPALA}Best mean rewards across 7 seeds. The standout algorithms are highlighted in bold. The final column details the performance difference percentage between GPBT and PBT.}
\hfill{}\subfloat{\hfill{}%
\begin{tabular}{ccccccc}
\toprule 
Benchmarks & $n$ & RS & PB2 & PBT & GPBT-PL & vs. PBT\tabularnewline
\midrule 
Breakout & 4 & 131 & 130 & 141 & \textbf{185} & 31\%\tabularnewline
SpaceInvaders & 4 & 611 & 485 & 634 & \textbf{725} & 14\%\tabularnewline
\bottomrule
\end{tabular}\hfill{}}\hfill{}
\end{table}

\begin{figure}[tbh]
\hfill{}\subfloat[\label{fig:breakout}Breakout]{\includegraphics[scale=0.18]{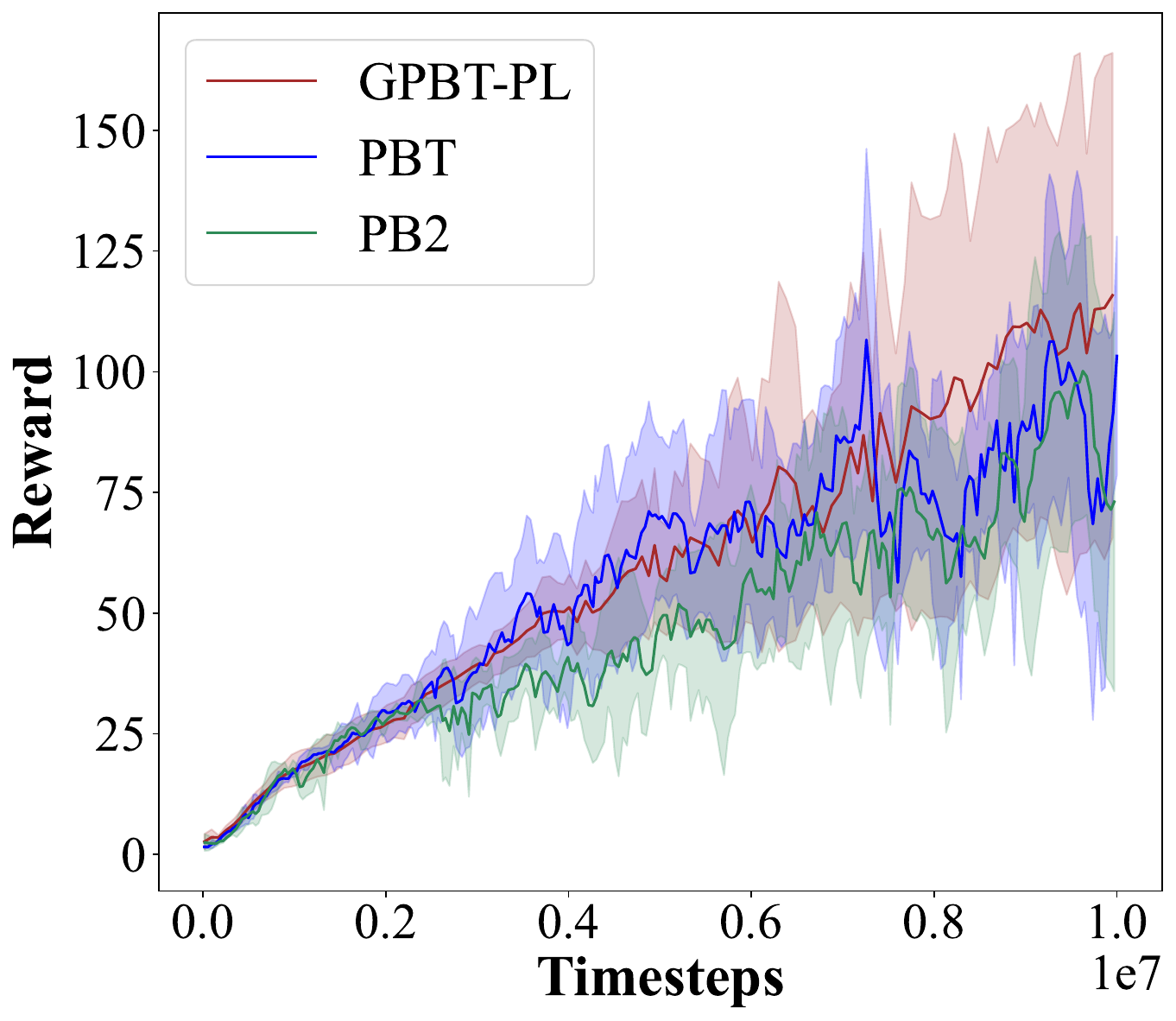}

}\hfill{}\subfloat[\label{fig:spaceinvaders}Space Invaders]{\includegraphics[scale=0.18]{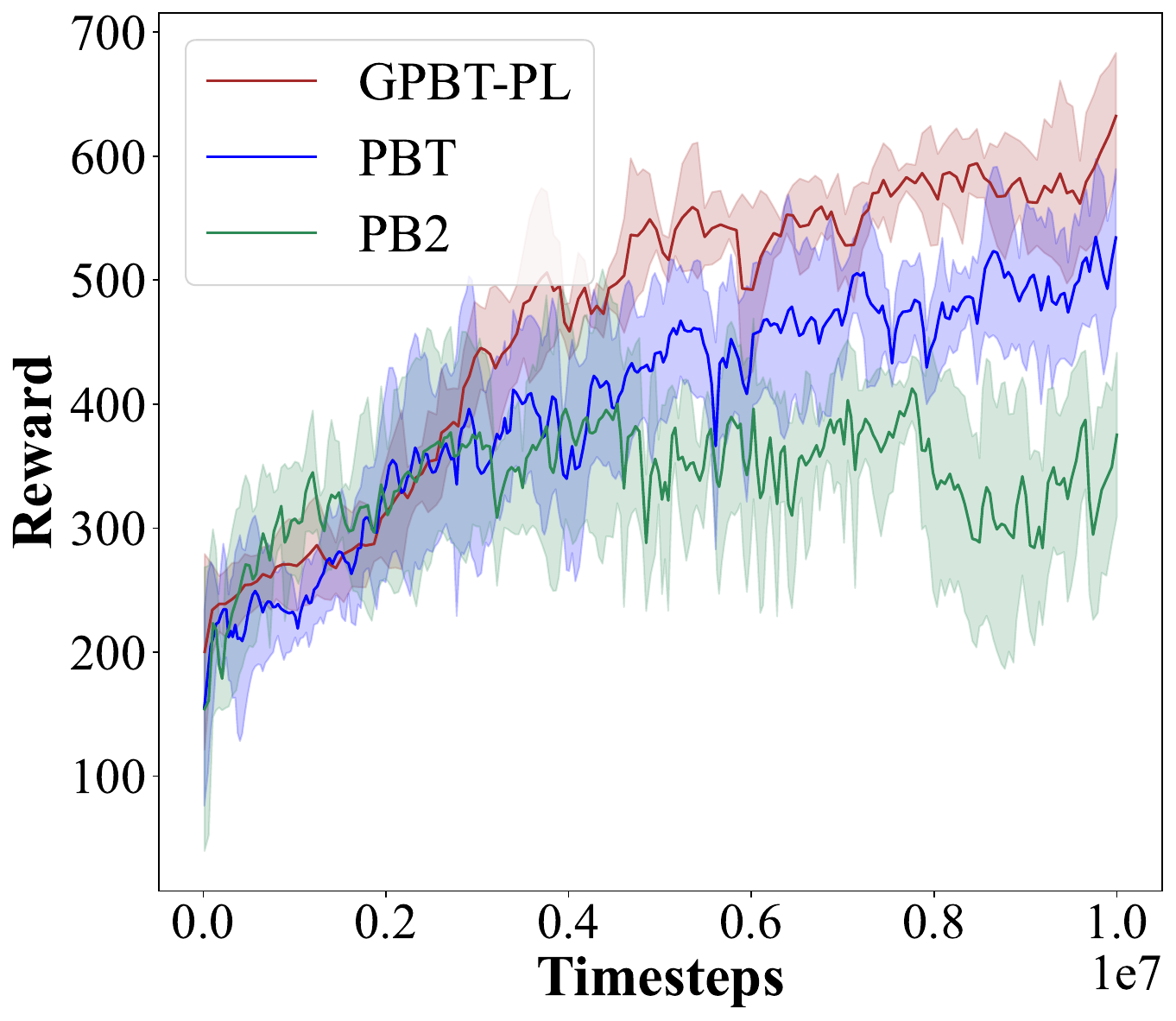}

}\hfill{}
\caption{\label{fig:IMPALA}Training curves for 4-agent populations using GPBT-PL, PBT, and PB2 on two OpenAI Gym benchmarks. Thick lines denote average best rewards over 7 seeds, and shaded regions indicate standard deviation. The learning rate is set between $[10^{-5},10^{-3}]$.}
\end{figure}

\subsection{\label{subsec:Off-Policy-RL}Off-Policy Reinforcement Learning}

In experiments for HPO in off-policy RL, we optimized the hyperparameters for the IMPALA algorithm on two games from the Arcade Learning Environment: Breakout and SpaceInvaders \cite{Bellemare2013}. 
We used the same three hyperparameters as the original IMPALA paper (epsilon, learning rate \(\eta\), and entropy coefficient) and conducted our experiments with a population size of 4. 
Training was performed over 10 million timesteps, equivalent to 40 million frames, with a perturbation interval of \(5 \times 10^{4}\) timesteps. Results are presented in \tablename~\ref{tab:IMPALA} and visualized in \figurename~\ref{fig:IMPALA}.

GPBT-PL consistently outperformed PBT, with significant improvements of 31\% in Breakout and 14\% in SpaceInvaders. 
This lead is more evident when compared against PB2 and RS. 
It is noteworthy that agents with more workers tend to perform better during training \cite{Espeholt2018}. 
Impressively, GPBT-PL, with each agent trained using only 5 workers, matched the performance of a hand-tuned IMPALA with 32 workers in SpaceInvaders, as observed in RLlib\footnote{\href{https://github.com/ray-project/rl-experiments}{RLlib IMPALA 32-workers experiments}}. 
In Breakout, while not reaching this benchmark, GPBT-PL still matched the performance of A3C with 16 workers, as referenced from Fig. 3 in \cite{Mnih2016}. 
This underscores GPBT-PL's ability to achieve high-level performance in RL, even with constrained computational resources.

\begin{table}[tbh]
\caption{\label{tab:PPO-10000}Best mean rewards across 7 seeds. The best-performing
algorithms are bolded. The last column presents the percentage of
performance difference between GPBT and PBT, where differences less
than 1\% are represented with $\approx$.}

\hfill{}\subfloat{\hfill{}%
\begin{tabular}{ccccccc}
\toprule 
Benchmarks & $n$ & RS & PB2 & PBT & GPBT-PL & vs. PBT\tabularnewline
\midrule 
BipedalWalker & 4 & 292 & 295 & 300 & \textbf{308} & +2\%\tabularnewline
Ant & 4 & 4283 & 3010 & 4051 & \textbf{5722} & +41\%\tabularnewline
HalfCheetah & 4 & 4834 & 5163 & 5161 & \textbf{6264} & +21\%\tabularnewline
InvertedDP & 4 & 8531 & 9343 & \textbf{9358} & 9355 & $\approx$\tabularnewline
Swimmer & 4 & 133 & 140 & 157 & \textbf{158} & $\approx$\tabularnewline
Walker2D & 4 & \textbf{2267} & 1829 & 2074 & 2200 & +6\%\tabularnewline
\midrule 
BipedalWalker & 8 & 284 & 294 & \textbf{303} & 297 & -2\%\tabularnewline
Ant & 8 & 4508 & 4876 & 5207 & \textbf{5724} & +10\%\tabularnewline
HalfCheetah & 8 & 4842 & 4570 & 4705 & \textbf{5853} & +24\%\tabularnewline
InvertedDP & 8 & 9185 & 9344 & \textbf{9357} & 9353 & $\approx$\tabularnewline
Swimmer & 8 & \textbf{155} & 129 & 132 & 135 & +2\%\tabularnewline
Walker2D & 8 & 2776 & 1339 & 3017 & \textbf{3054} & +1\%\tabularnewline
\bottomrule
\end{tabular}\hfill{}}\hfill{}
\end{table}

\begin{figure*}[tbh]
\hfill{}\subfloat[\label{fig:bipedal_4_10000}BipedalWalker (4)]{\includegraphics[scale=0.18]{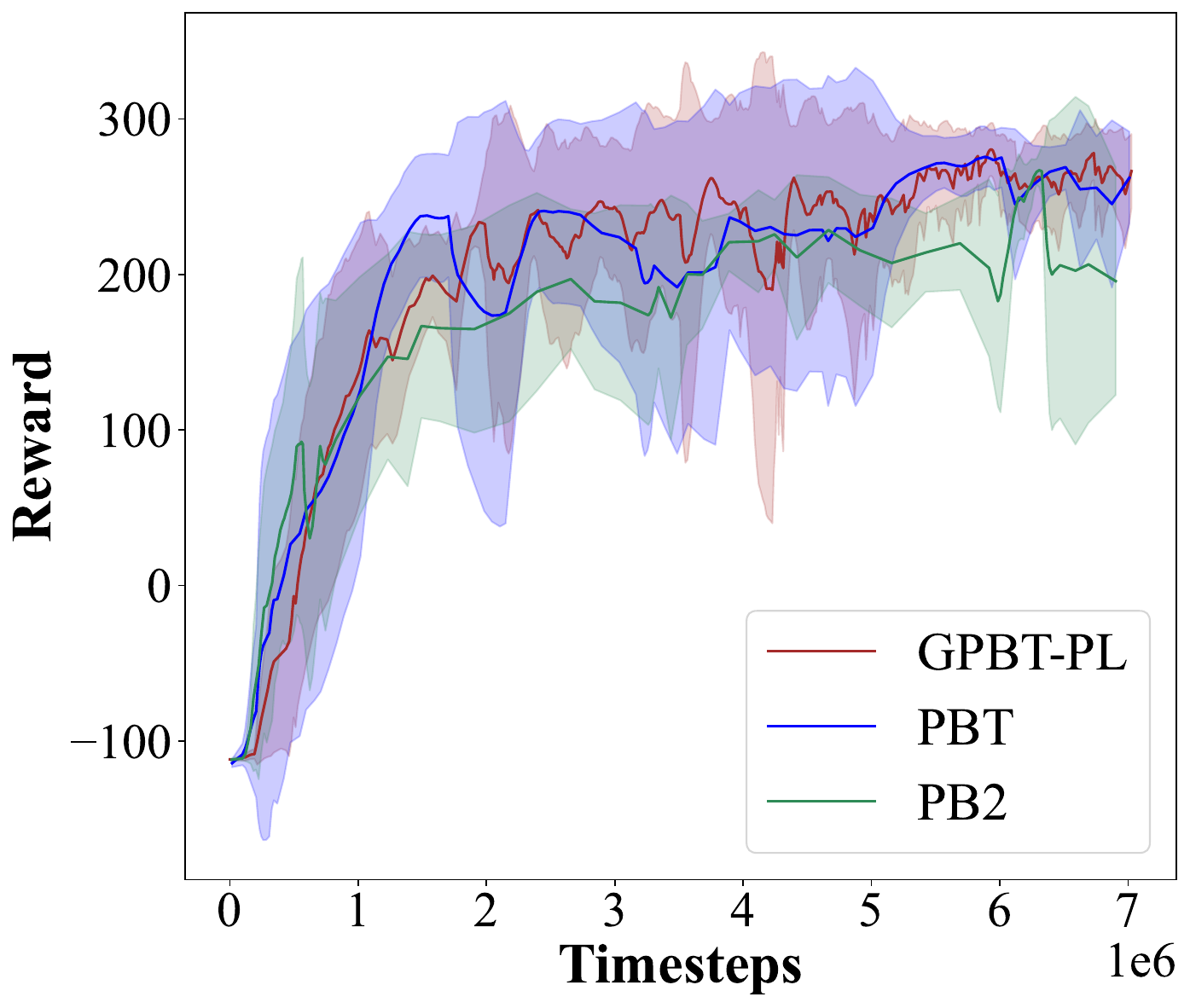}

}\hfill{}\subfloat[\label{fig:ant_4_10000}Ant (4)]{\includegraphics[scale=0.18]{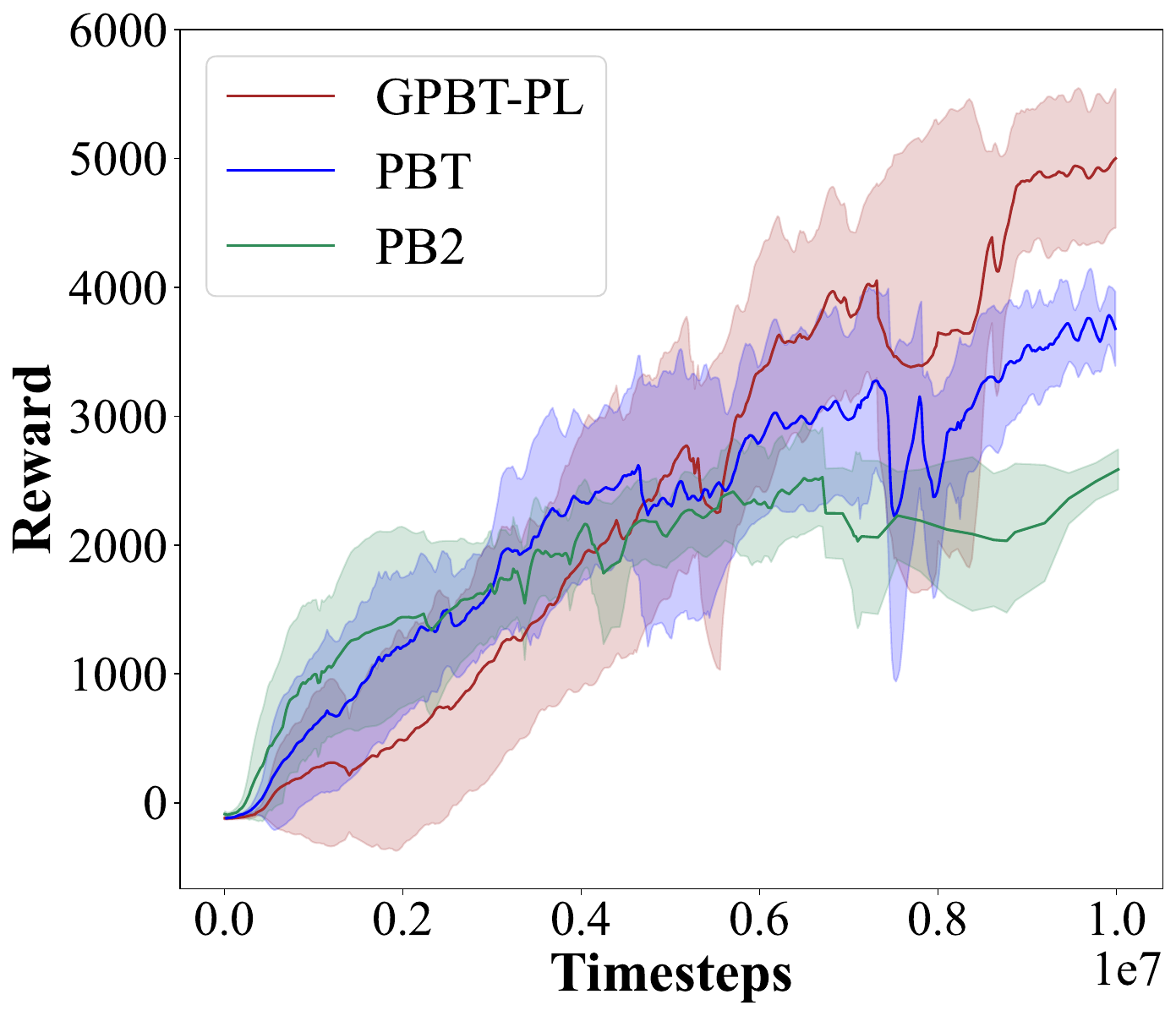}

}\hfill{}\subfloat[\label{fig:halfcheetah_4_10000}HalfCheetah (4)]{\includegraphics[scale=0.18]{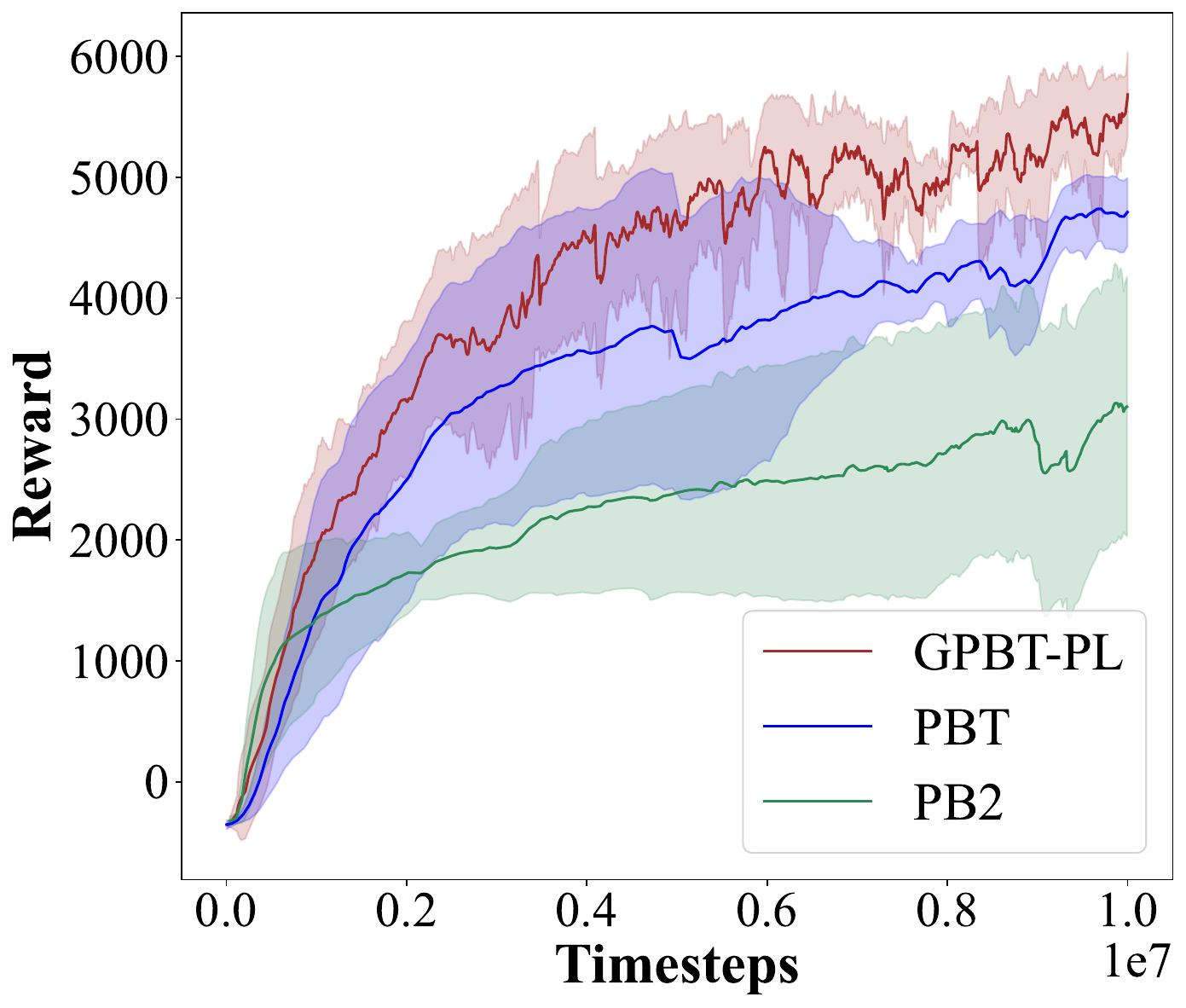}

}\hfill{}\subfloat[\label{fig:invertedDP_4_10000}InvertedDoublePendulum (4)]{\includegraphics[scale=0.18]{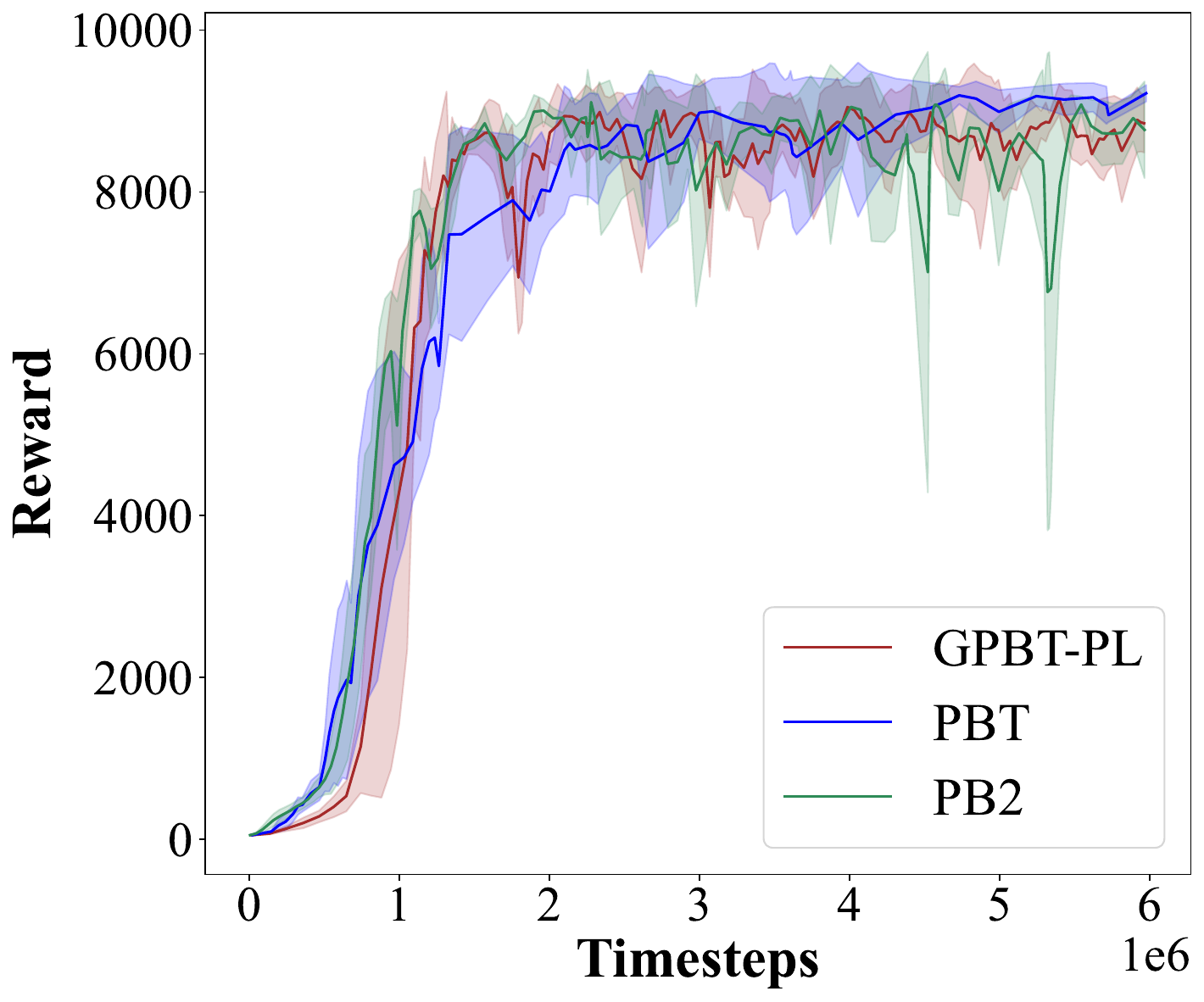}

}\hfill{}

\hfill{}\subfloat[\label{fig:swimmer_4_10000}Swimmer (4)]{\includegraphics[scale=0.18]{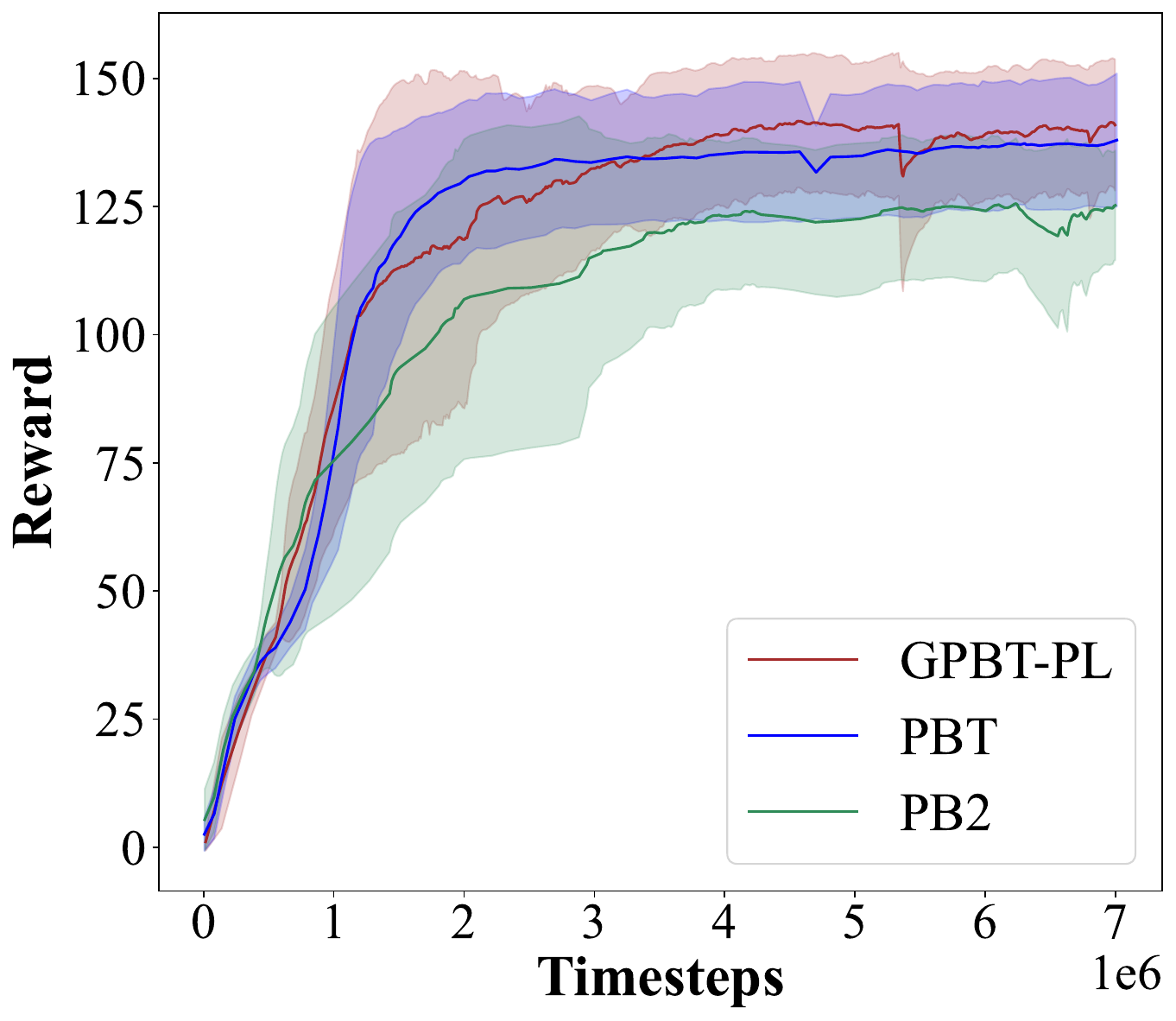}

}\hfill{}\subfloat[\label{fig:walker2d_4_10000}Walker2D (4)]{\includegraphics[scale=0.18]{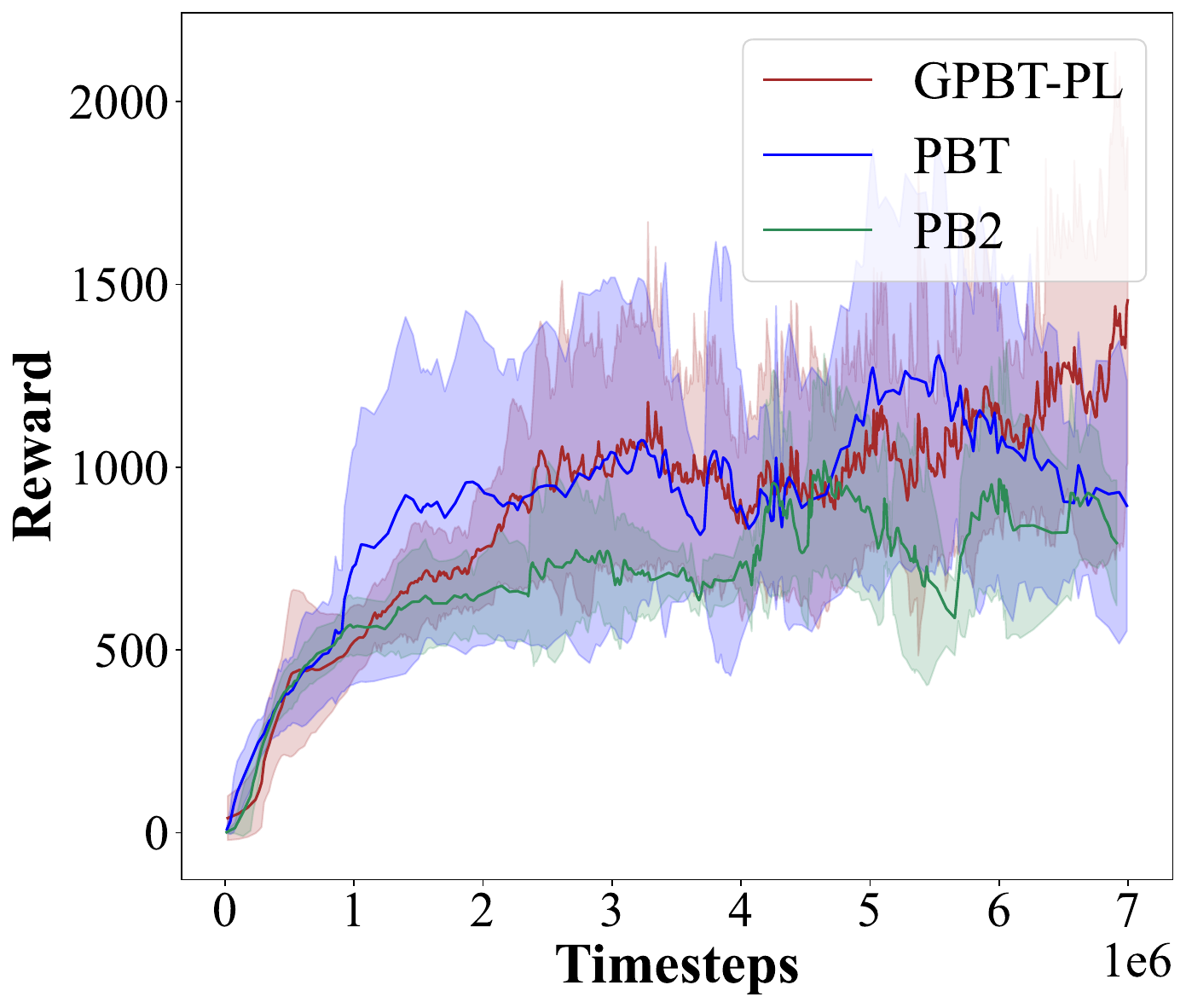}

}\hfill{}\subfloat[\label{fig:bipedal_8_10000}BipedalWalker (8)]{\includegraphics[scale=0.18]{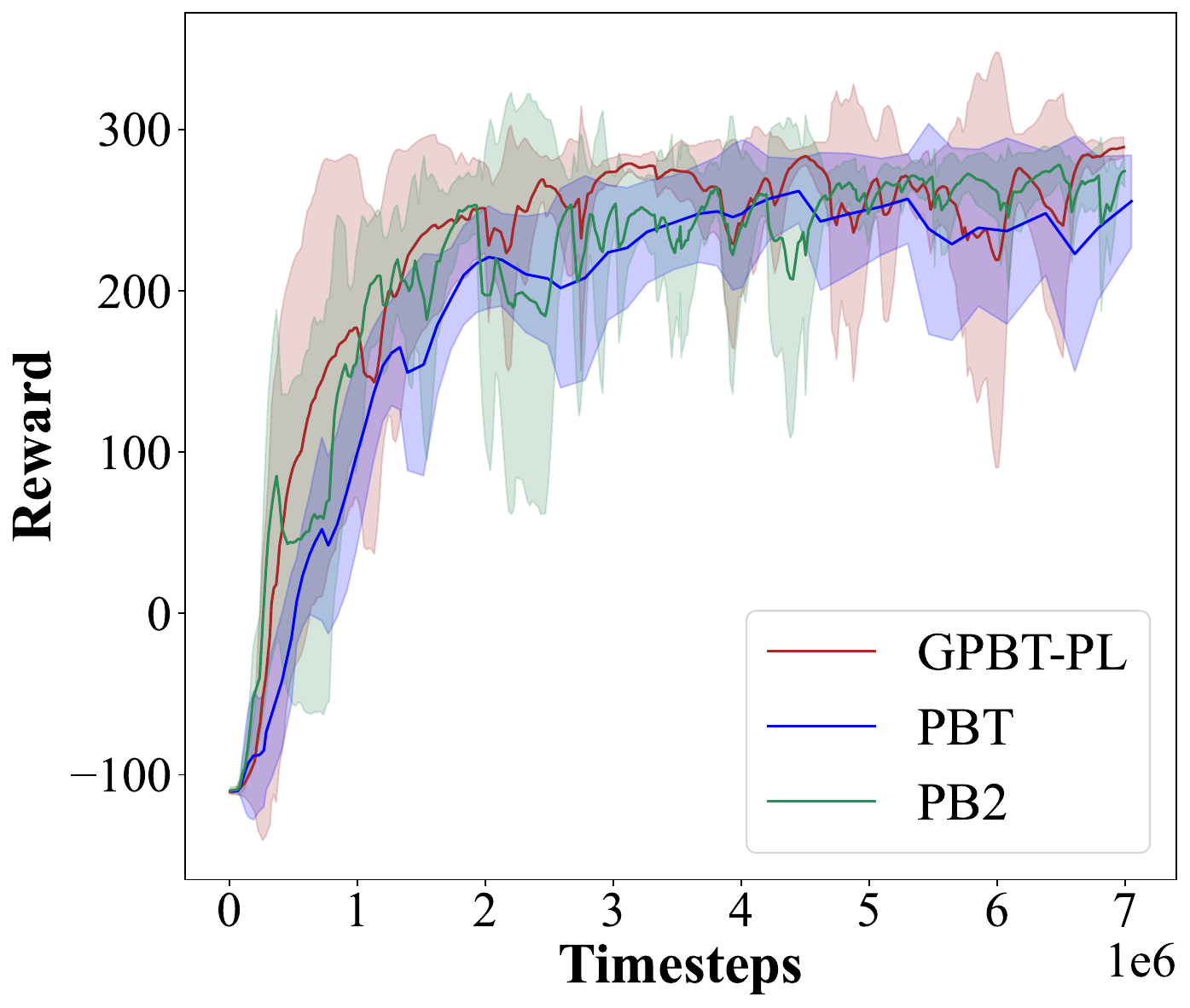}

}\hfill{}\subfloat[\label{fig:ant_8_10000}Ant (8)]{\includegraphics[scale=0.18]{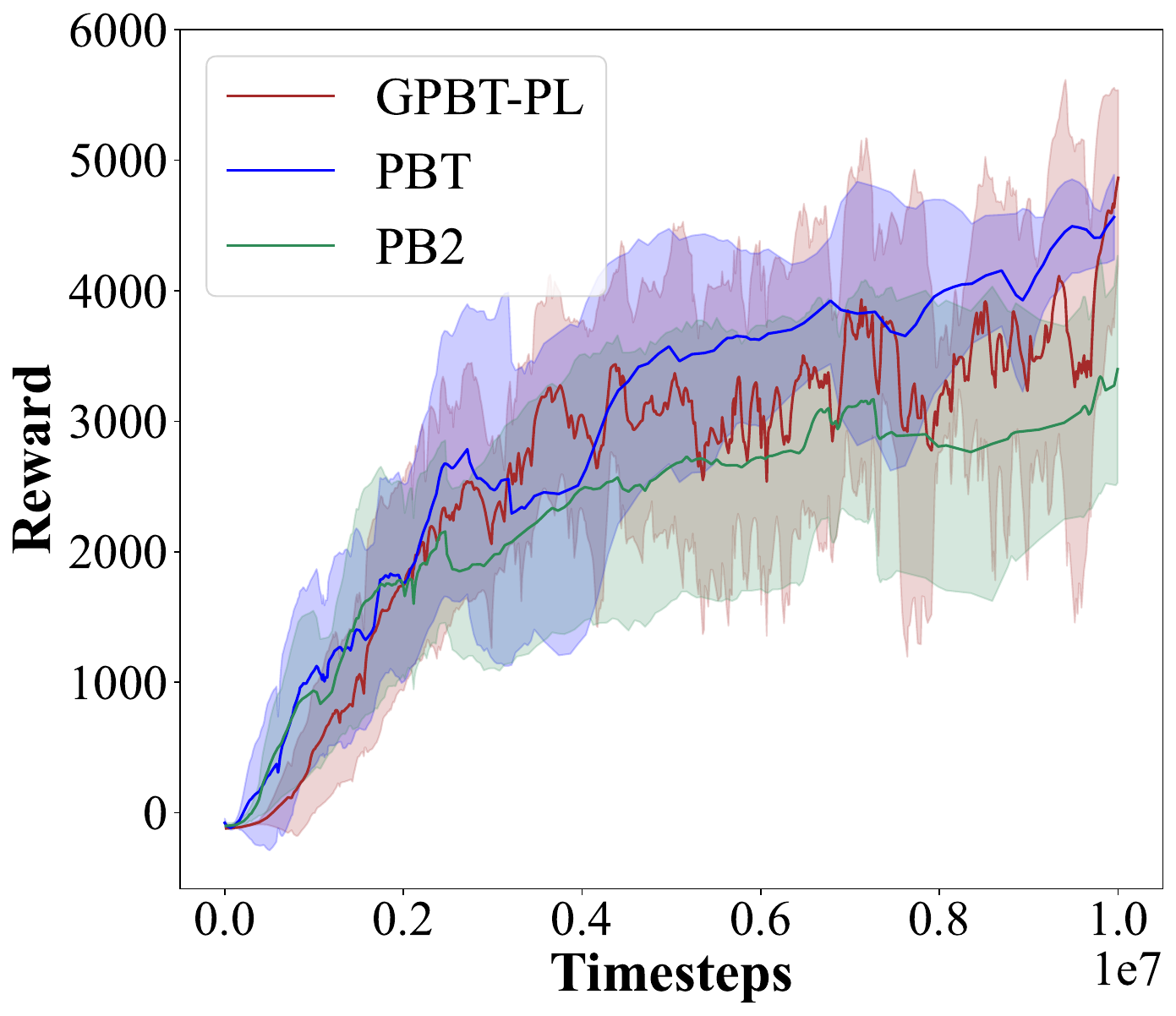}

}\hfill{}

\hfill{}\subfloat[\label{fig:halfcheetah_8_10000}HalfCheetah (8)]{\includegraphics[scale=0.18]{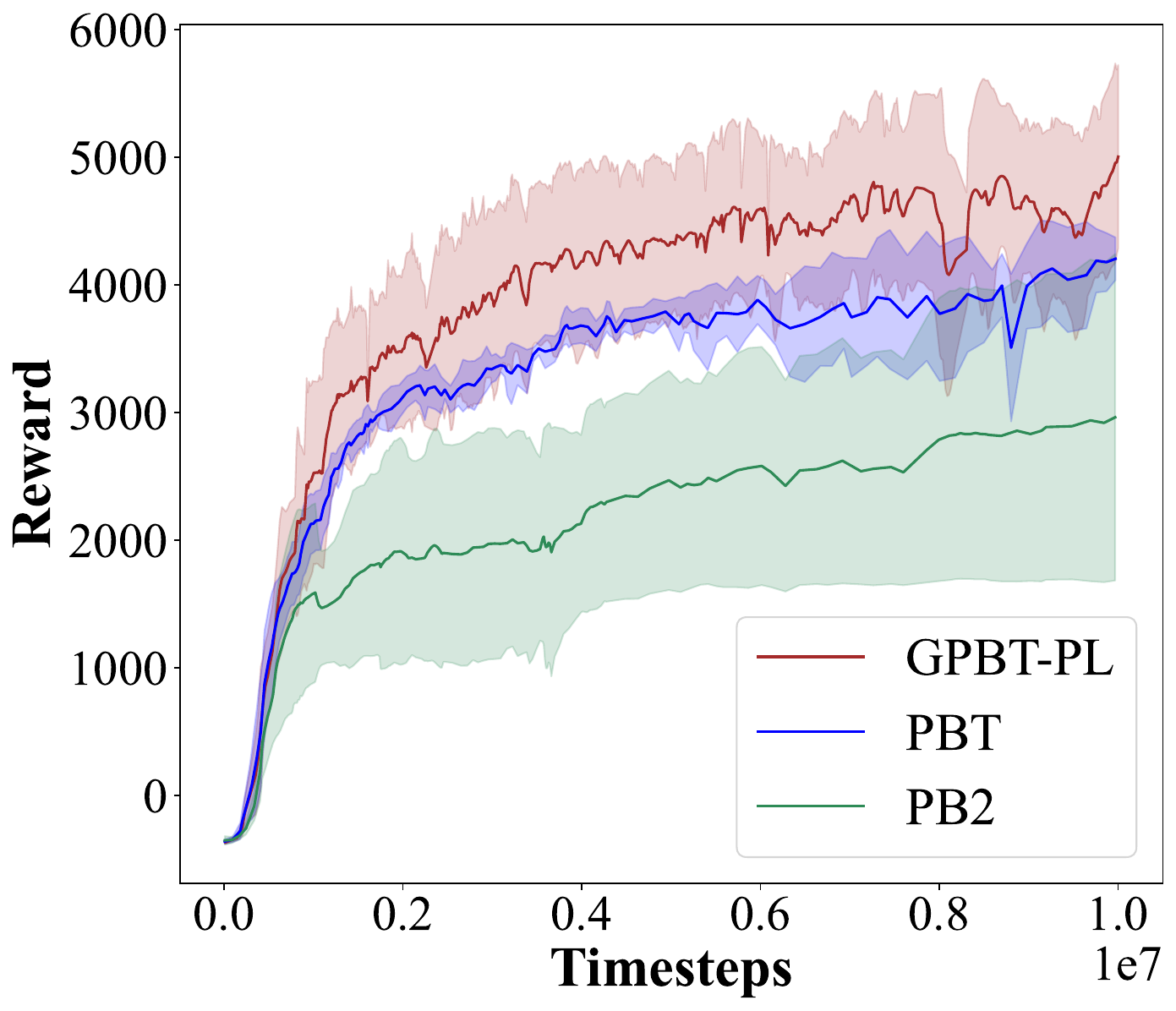}

}\hfill{}\subfloat[\label{fig:invertedDP_8_10000}InvertedDoublePendulum (8)]{\includegraphics[scale=0.18]{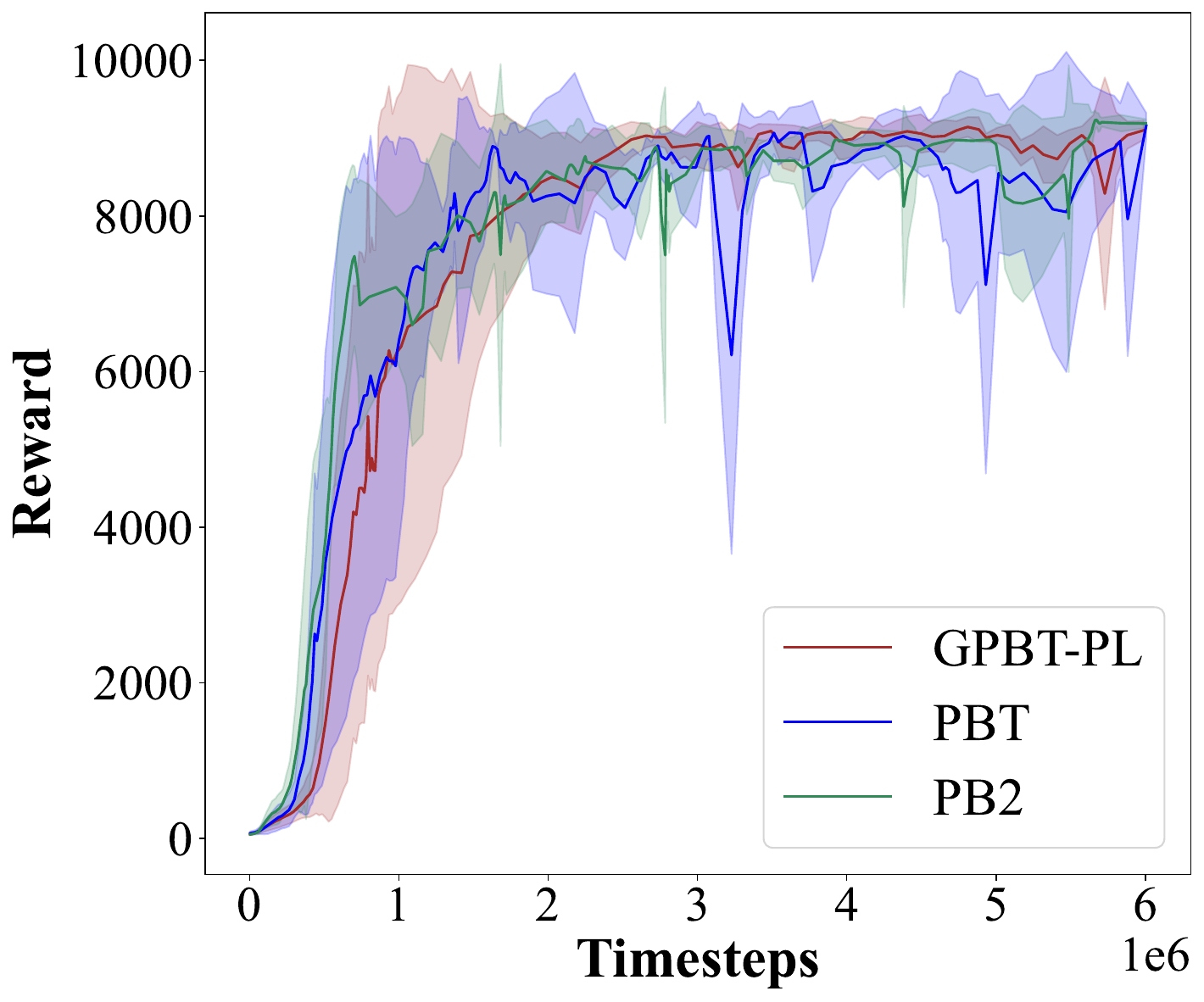}

}\hfill{}\subfloat[\label{fig:swimmer_8_10000}Swimmer (8)]{\includegraphics[scale=0.18]{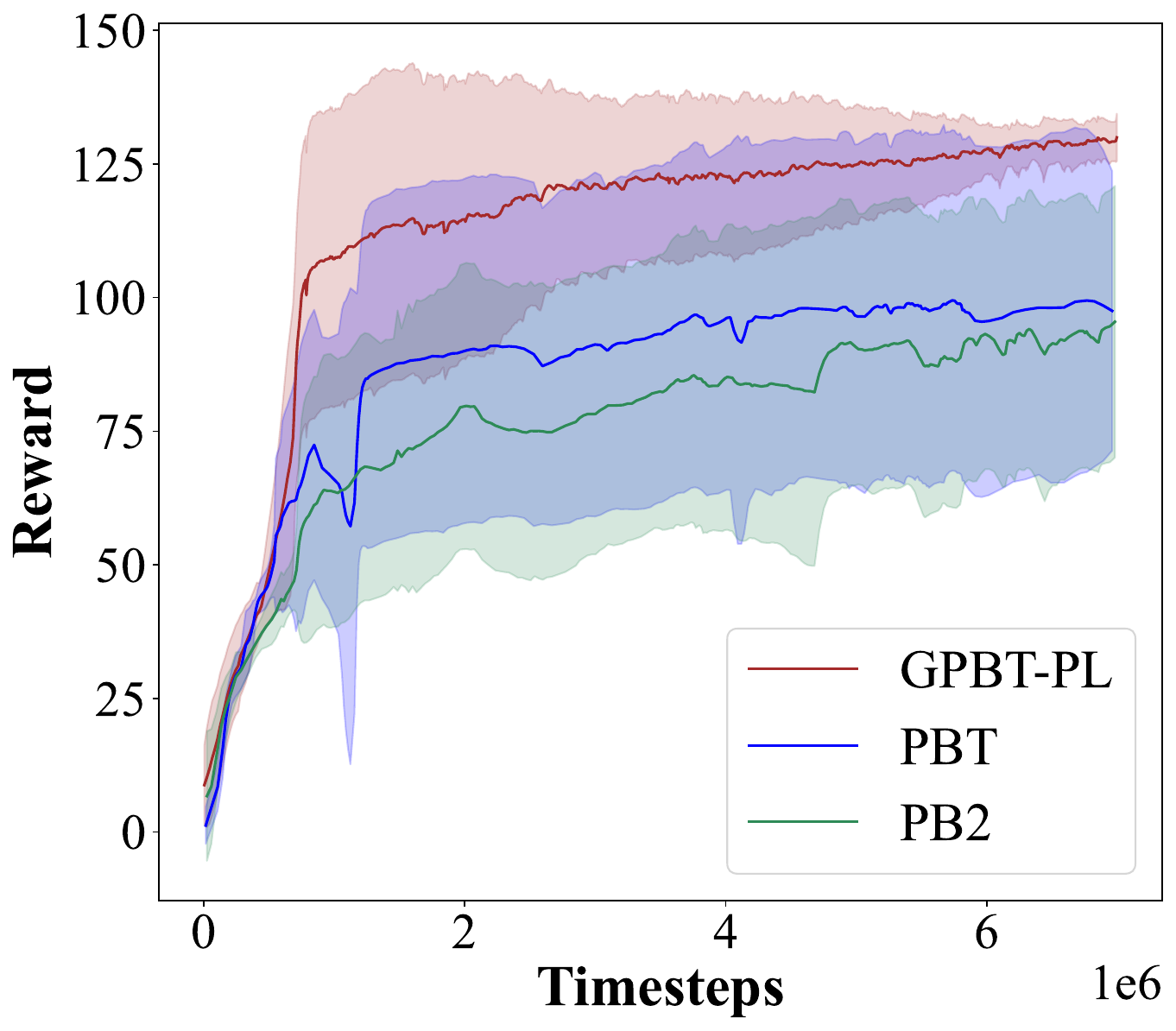}

}\hfill{}\subfloat[\label{fig:walker2d_8_10000}Walker2D (8)]{\includegraphics[scale=0.18]{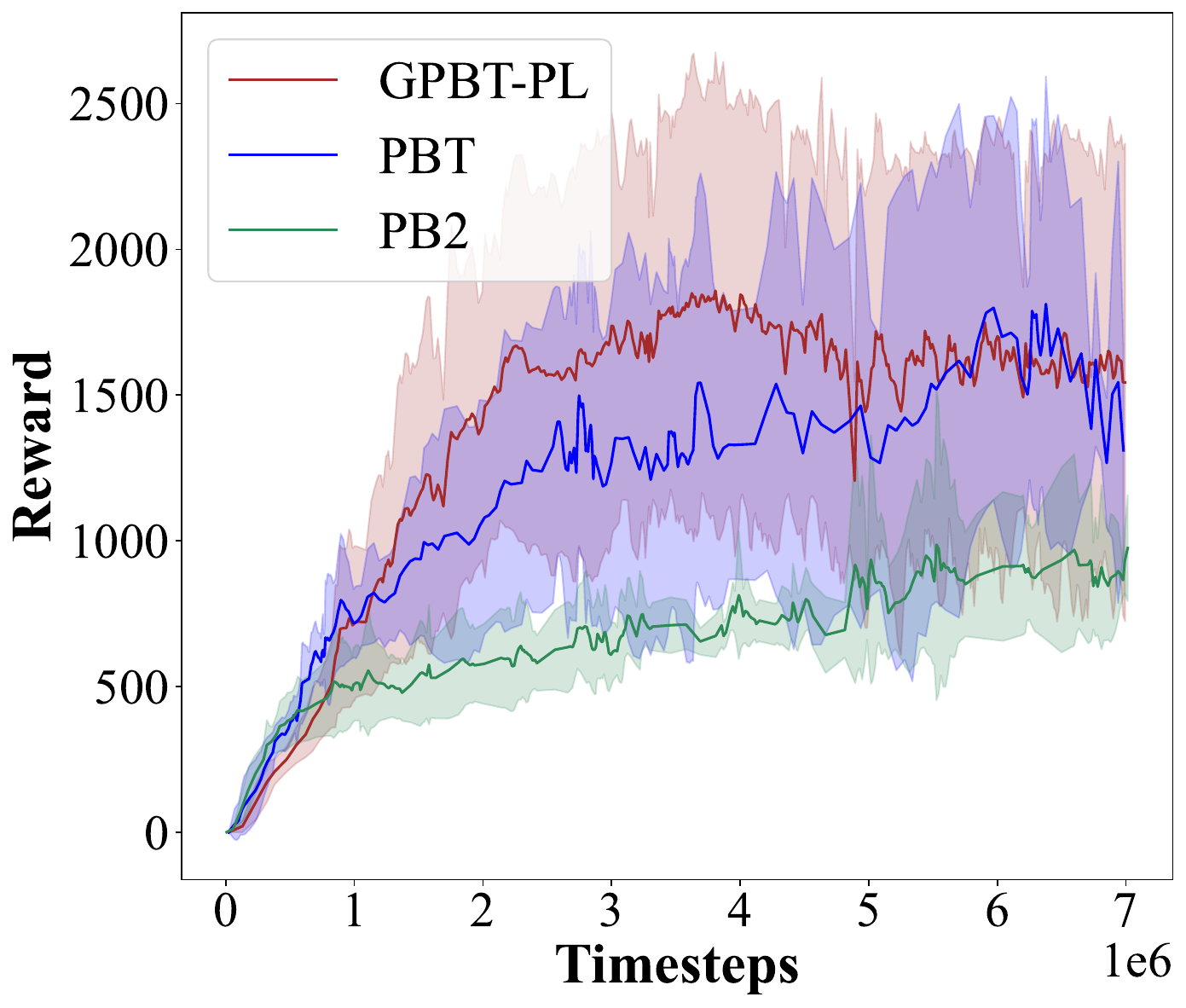}

}\hfill{}

\caption{\label{fig:PPO_10000}Training curves for six OpenAI Gym benchmarks using populations of 4 and 8 agents with GPBT-PL, PBT, and PB2. Thick lines represent the average of the best mean rewards over 7 seeds, with shaded regions denoting the standard deviation. Brackets specify the population size, and the perturbation interval is set to \(1 \times 10^{4}\).}

\end{figure*}

\subsection{\label{subsec:Robustness-to-Perturbation-Interval}Robustness to Perturbation Interval}

The perturbation interval plays a pivotal role in PBT-class HPO algorithms. 
With a fixed training budget, a larger interval allows agents more training before each hyperparameter change, ensuring accurate performance evaluations. 
This, however, may result in fewer hyperparameter adjustments and reduced search space exploration. 
Conversely, a shorter interval can lead to frequent yet potentially unstable hyperparameter updates due to less precise agent evaluations. Given the task-specific nature of an optimal interval, it is crucial that PBT algorithms remain robust against its variations.

To assess this robustness, we reduced the perturbation interval from \(5 \times 10^{4}\) to \(1 \times 10^{4}\) timesteps, testing on our earlier tasks with populations of 4 and 8. 
The outcomes are detailed in \tablename~\ref{tab:PPO-10000} and \figurename~\ref{fig:PPO_10000}, using RS rewards from \tablename~\ref{tab:PPO-50000} as a baseline. 

GPBT-PL consistently outperformed PBT and PB2. When examining \figurename~\ref{fig:PPO_50000} versus \figurename~\ref{fig:PPO_10000}, a shared trend emerged: training curves became erratic and reward variances widened with a shorter interval. 
Though simpler tasks like BipedalWalker and InvertedDoublePendulum remained relatively stable, more complex tasks saw notable performance shifts, especially in PBT and PB2. 

Interestingly, GPBT-PL showcased heightened performance upper bounds in several instances, hinting at its resilience to interval changes.
This could be due to its adaptive nature, which corrects early performance misjudgments as evaluations refine over time. 
In contrast, PBT's propensity to discard potential solutions prematurely might stunt its long-term performance, and PB2, reliant on a Bayesian optimization model built on suboptimal solutions, might fail to produce superior hyperparameters.

\begin{figure}[tbh]
\hfill{}\subfloat[\label{fig:16-ant}Ant]{\includegraphics[scale=0.18]{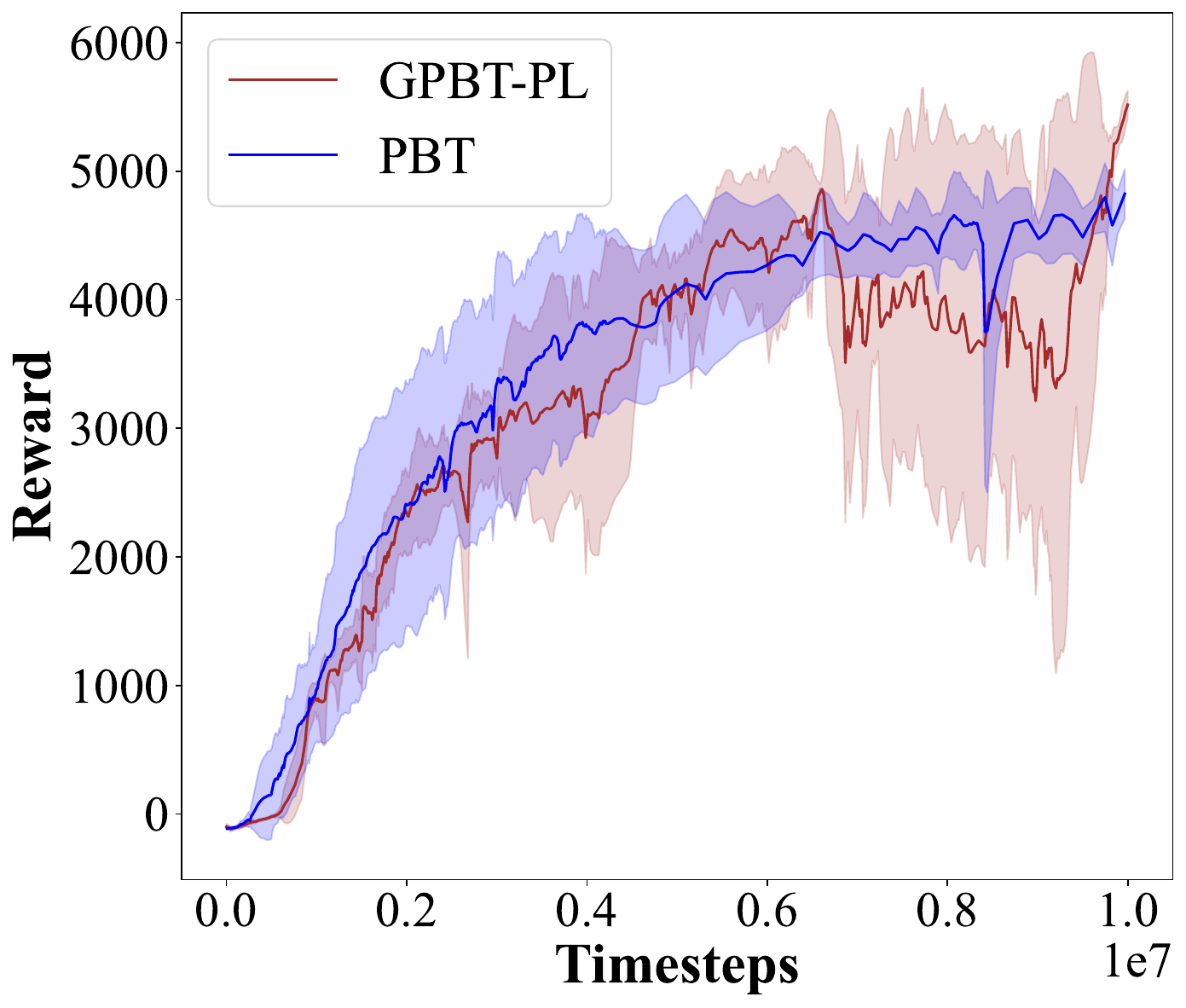}

}\hfill{}\subfloat[\label{fig:16-halfcheetah}HalfCheetah]{\includegraphics[scale=0.18]{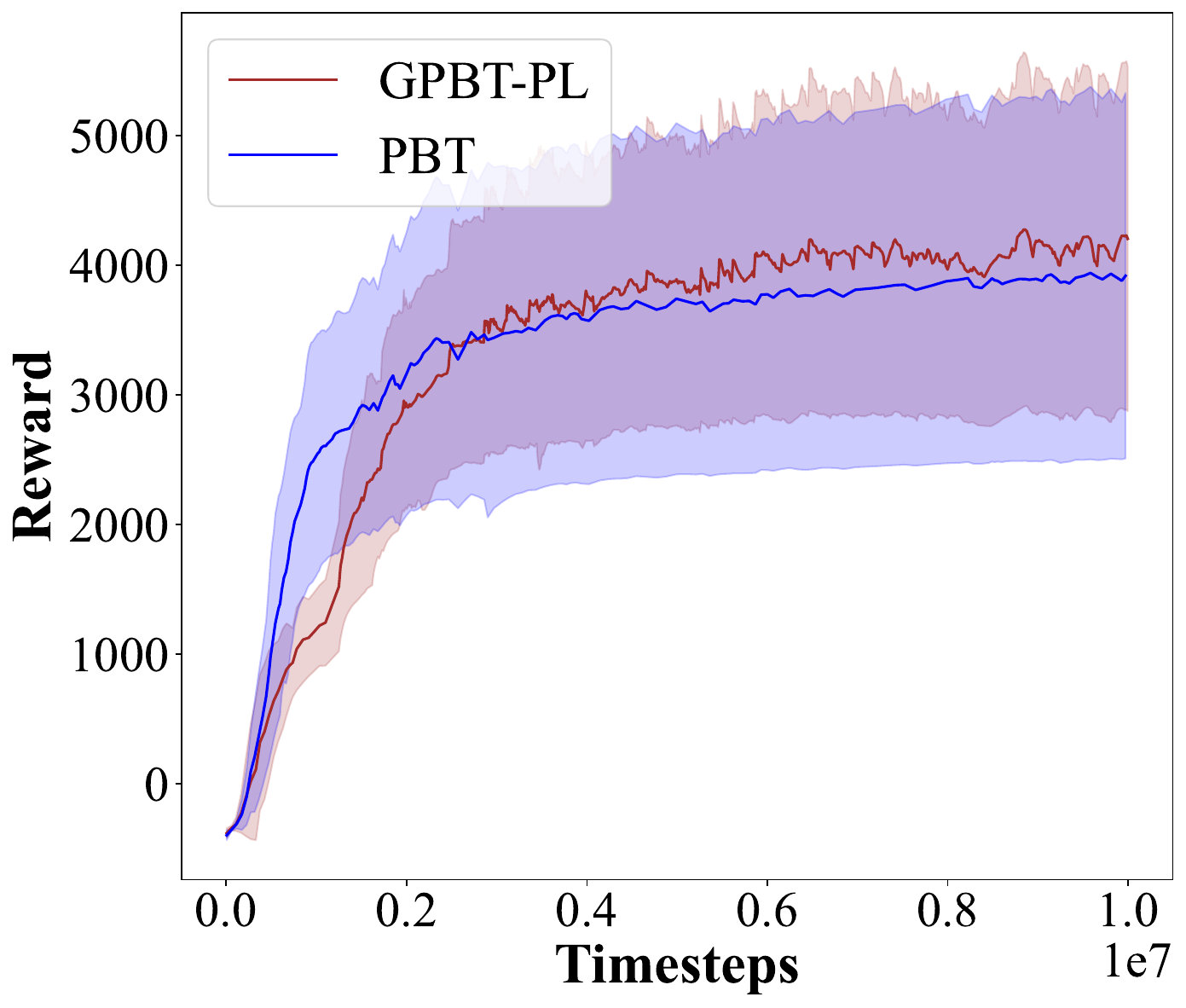}

}\hfill{}
\caption{\label{fig:PPO-16}Training curves for 16-agent populations using GPBT-PL and PBT. Thick lines represent average best rewards over 7 seeds, with shaded regions indicating standard deviation.}
\end{figure}

\subsection{\label{subsec:Scalability-to-Larger-Populations}Scalability to Larger Populations}

With the availability of more computational resources, understanding the scalability of algorithms becomes vital. 
To this end, we conducted tests on Ant and HalfCheetah using a population size of 16 and a perturbation interval of \(5 \times 10^{4}\) timesteps. 

Upon comparing \figurename~\ref{fig:PPO_50000} with \figurename~\ref{fig:PPO-16}, it becomes evident that for Ant, merely increasing the population from 8 to 16 does not amplify PBT's upper-performance limits. 
In contrast, GPBT-PL showcases its ability to surpass local optima in later stages, accessing regions of higher rewards. 
For both GPBT-PL and PBT, larger populations appear to enhance efficiency in pinpointing favorable solutions.

In the HalfCheetah test, even though the average of the best mean rewards remains relatively stable for both algorithms, there is a marked uplift in the performance upper bounds. 
In essence, while GPBT-PL and PBT exhibit variable scalability across tasks, both consistently benefit from increased efficiency with more expansive populations.

\begin{figure*}[tbh]
\hfill{}\subfloat[\label{fig:many_hyperpa_ant_4_50000}Ant (4)]{\includegraphics[scale=0.18]{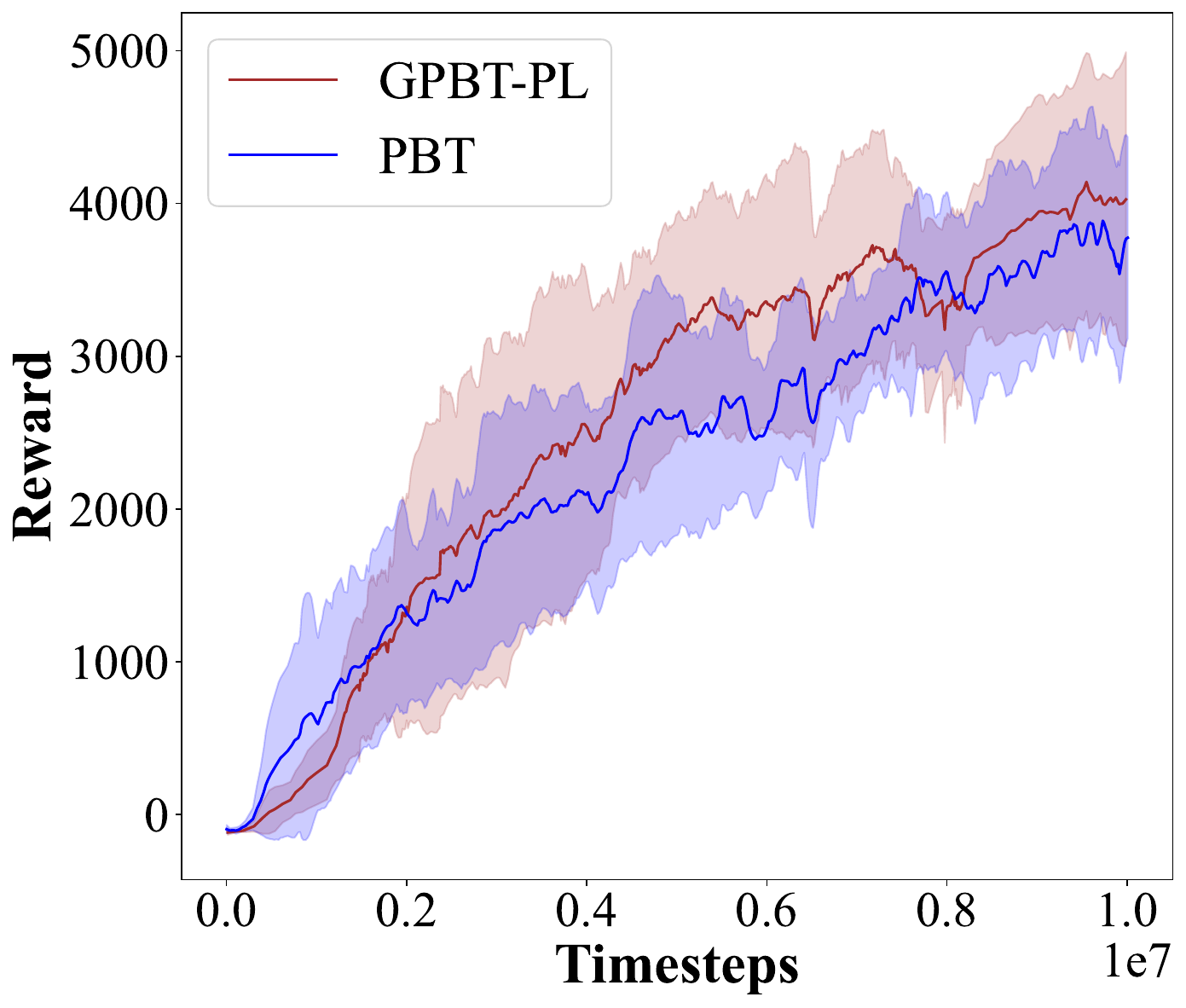}

}\hfill{}\subfloat[\label{fig:many_hyperpa_ant_8_50000}Ant (8)]{\includegraphics[scale=0.18]{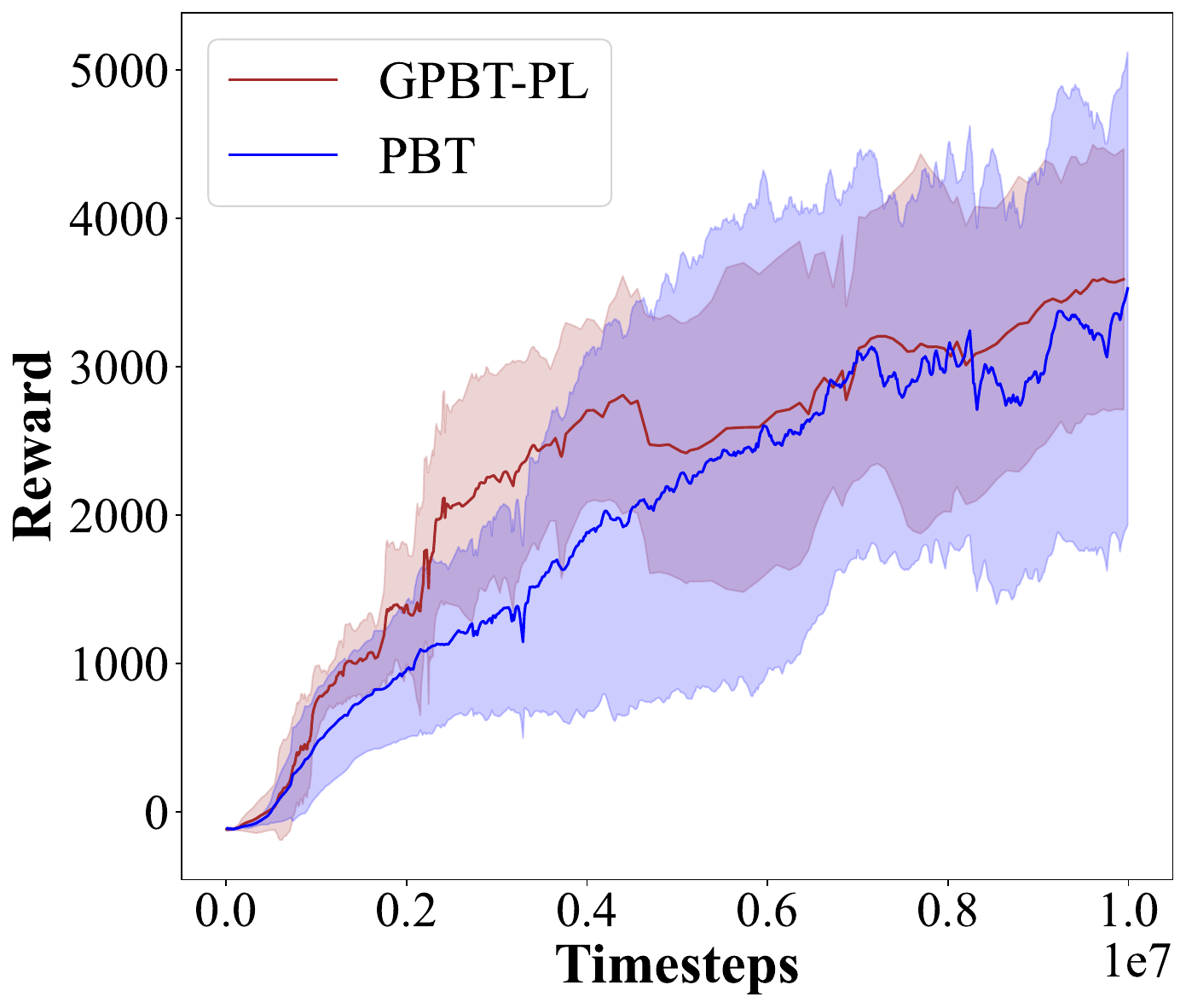}

}\hfill{}\subfloat[\label{fig:many_hyperpa_ant_16_50000}Ant (16)]{\includegraphics[scale=0.18]{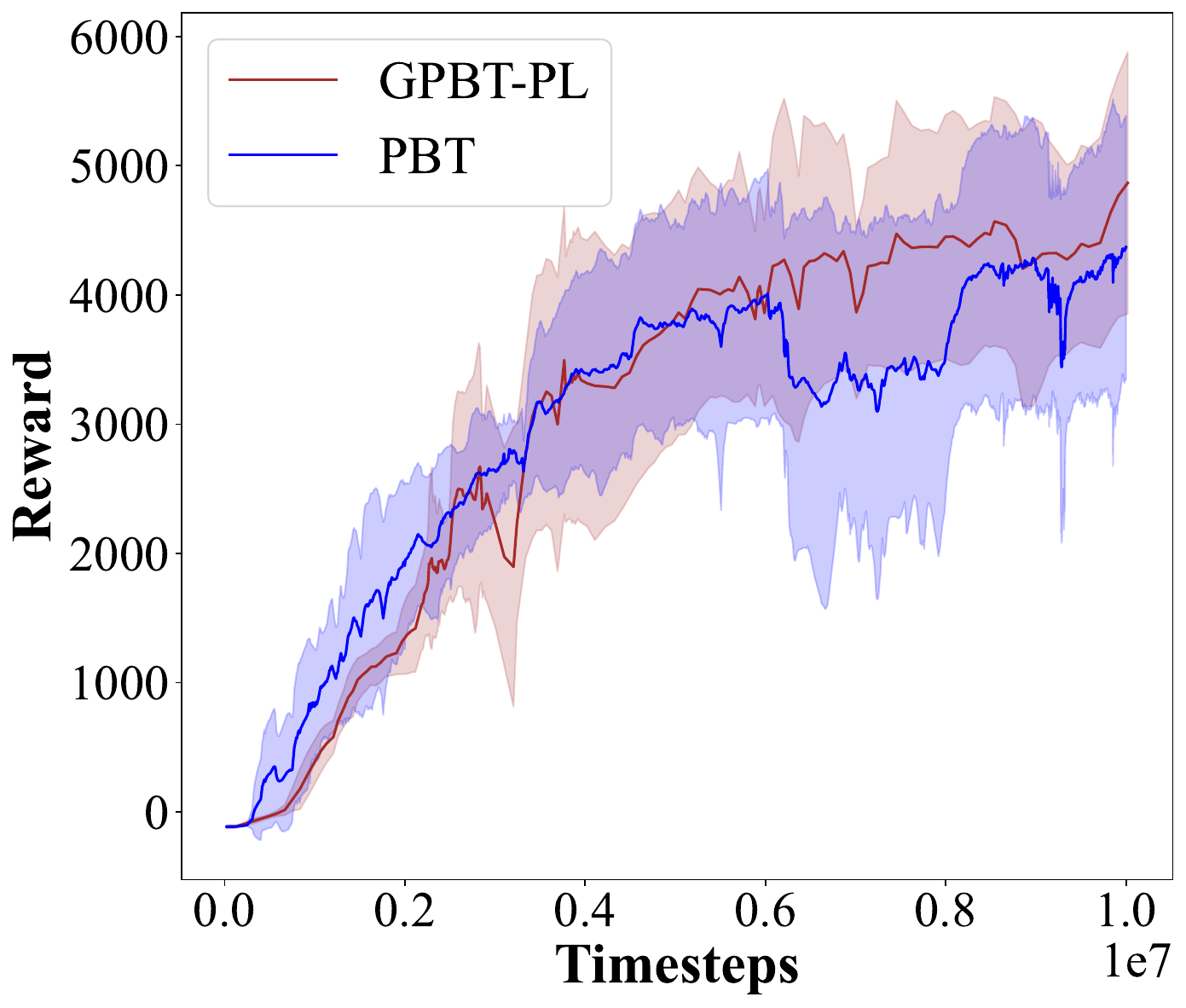}

}\hfill{}

\hfill{}\subfloat[\label{fig:many_hyperpa_half_4_50000}HalfCheetah (4)]{\includegraphics[scale=0.18]{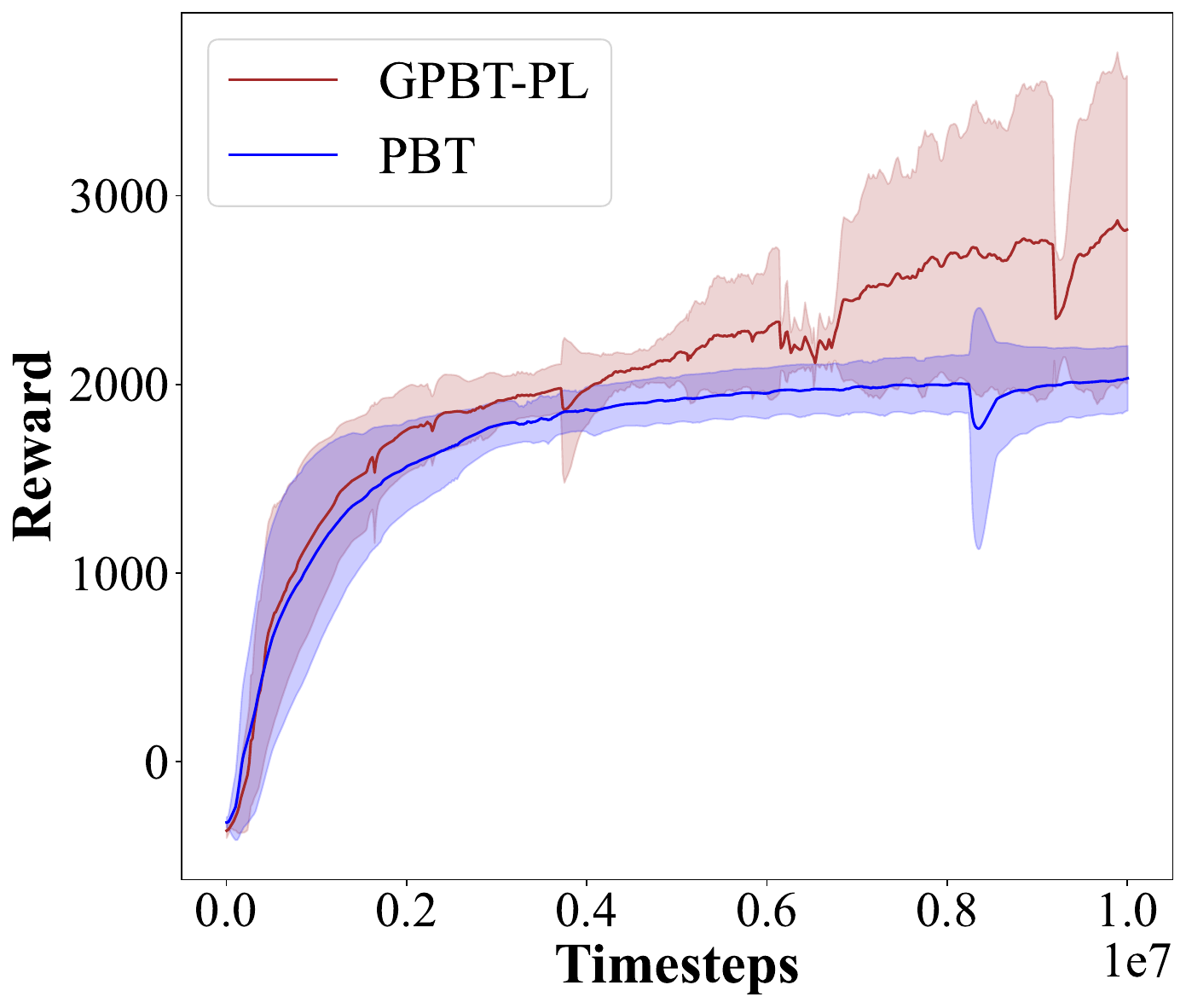}

}\hfill{}\subfloat[\label{fig:many_hyperpa_half_8_50000}HalfCheetah (8)]{\includegraphics[scale=0.18]{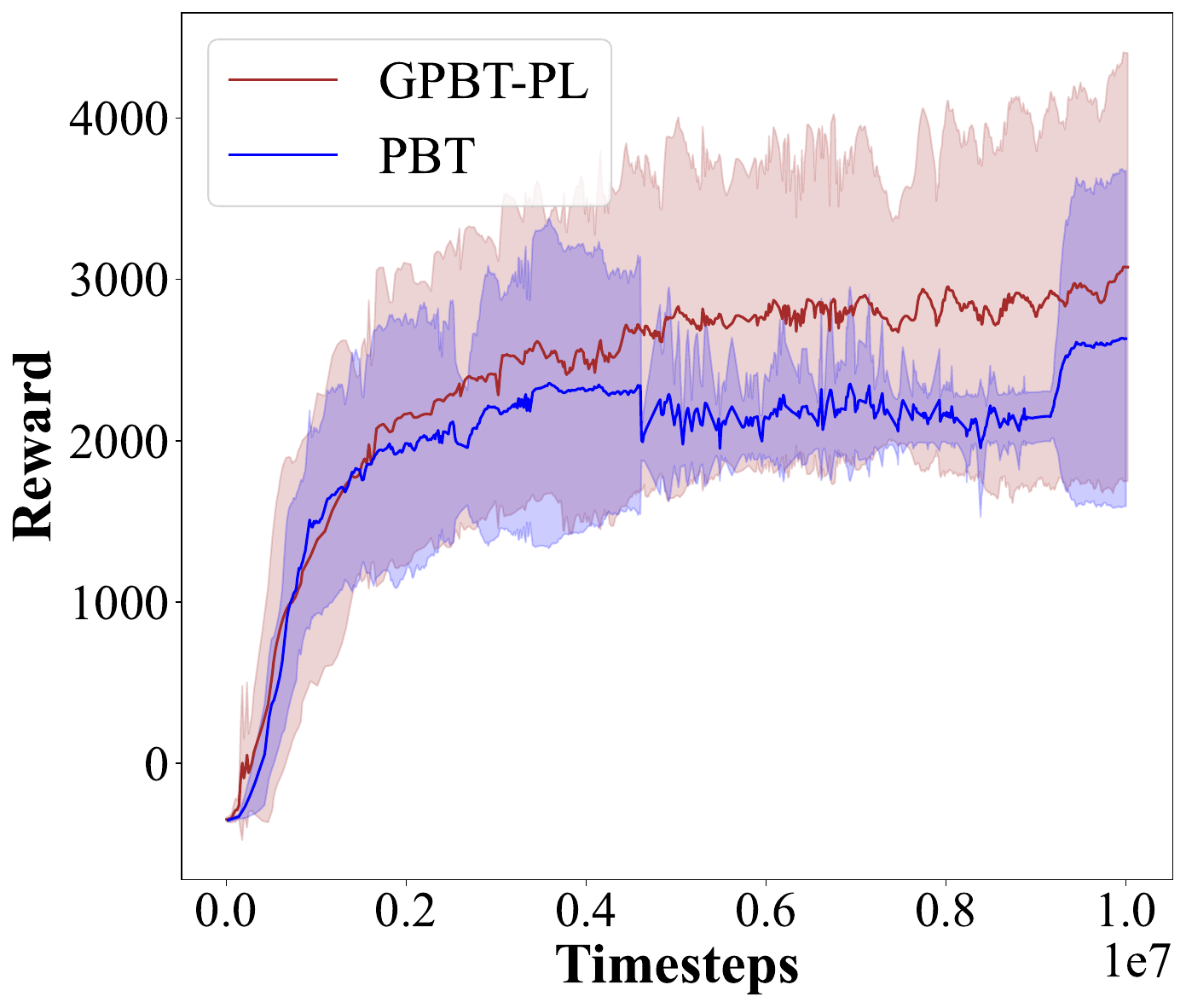}

}\hfill{}\subfloat[\label{fig:many_hyperpa_half_16_50000}HalfCheetah (16)]{\includegraphics[scale=0.18]{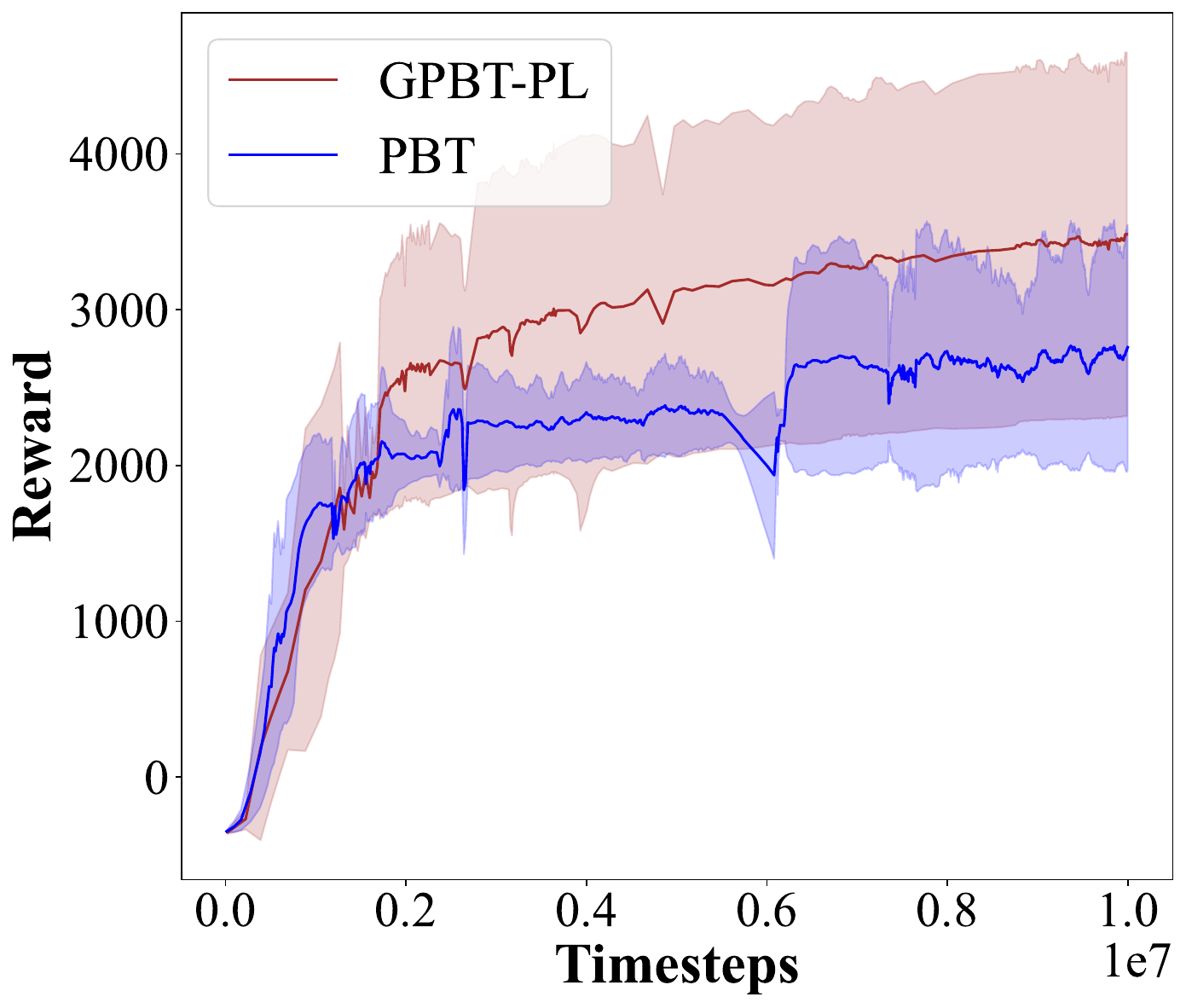}

}\hfill{}

\caption{\label{fig:many_hyperparameters}Training curves for Ant and HalfCheetah using populations of 4, 8, and 16 agents with GPBT and PBT. Thick lines represent the average of the best mean rewards over 7 seeds, with shaded regions denoting the standard deviation. Brackets specify the population size, and the perturbation interval is set to \(5 \times 10^{4}\). The number of optimized hyperparameters is increased to 7.}

\end{figure*}

\subsection{\label{subsec:Many-Hyperparameters}Scalability to Many Hyperparameters}
As the number of hyperparameters increases, the complexity of HPO problems escalates, typically resulting in a decrease in the effectiveness of general HPO methods. Although existing PBT-like HPO algorithms have not demonstrated their behavior when the number of hyperparameters exceeds four, it remains crucial to explore their efficacy under such circumstances. Consequently, we conducted experiments on Ant and HalfCheetah using population sizes of 4, 8, and 16, with a perturbation interval of \(5 \times 10^{4}\). Notably, the number of optimized hyperparameters was increased from 4 to 7, encompassing batch size, GAE $\lambda$, PPO clip $\epsilon$, discount $\gamma$, SGD minibatch size, and SGD iterations. The training curves are illustrated in \figurename~\ref{fig:many_hyperparameters}.

GPBT-PL consistently outperforms PBT on Ant and HalfCheetah across different population sizes. However, upon comparing \figurename~\ref{fig:many_hyperparameters} with \figurename~\ref{fig:PPO_50000} and \figurename~\ref{fig:PPO-16}, it is evident that both GPBT-PL and PBT exhibit inferior performance with an increased number of hyperparameters. Although augmenting the population size proves beneficial, it also substantially escalates the demand for computational resources, rendering HPO algorithms less generalizable. Consequently, when selecting hyperparameters for simultaneous optimization, it is imperative to consult relevant literature and conduct preliminary experiments to determine the types and search ranges of hyperparameters. For instance, pertinent hyperparameters should be optimized together, and those that vary according to specific problems must be included in the optimization process.

\begin{table}[tbh]
\caption{\label{tab:IMPALA-1}Best mean rewards for 7 seeds. Top-performing algorithms are highlighted. The final column shows the performance difference percentage between GPBT and PBT.}

\hfill{}\subfloat{\hfill{}%
\begin{tabular}{ccccccc}
\toprule 
Benchmarks & $n$ & RS & PB2 & PBT & GPBT-PL & vs. PBT\tabularnewline
\midrule 
Breakout & 4 & 131 & 110 & 83 & \textbf{151} & 82\%\tabularnewline
SpaceInvaders & 4 & 611 & 339 & 551 & \textbf{685} & 24\%\tabularnewline
\bottomrule
\end{tabular}\hfill{}}\hfill{}
\end{table}

\begin{figure}[tbh]
\hfill{}\subfloat[\label{fig:breakout-1}Breakout]{\includegraphics[scale=0.18]{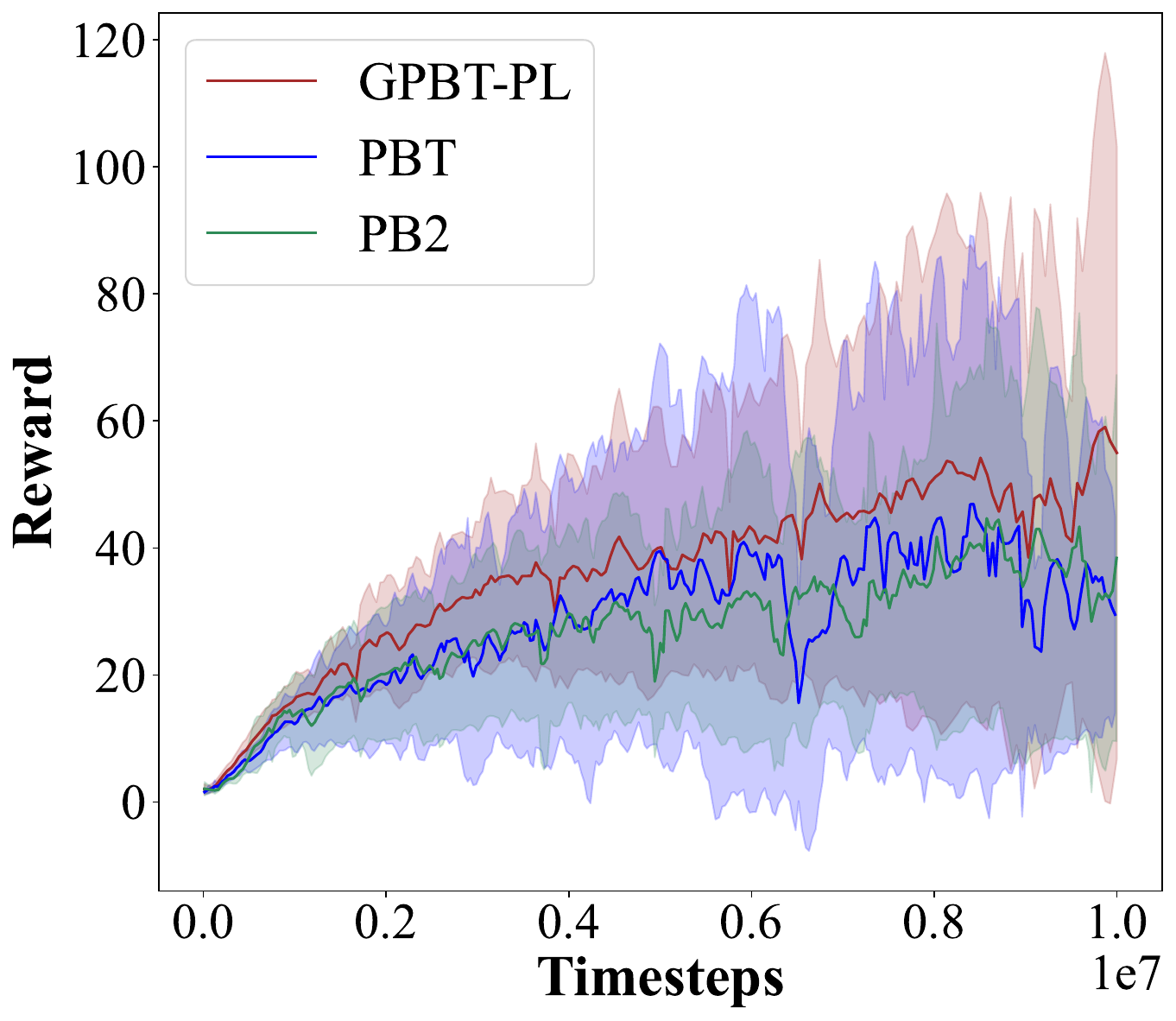}

}\hfill{}\subfloat[\label{fig:spaceinvaders-1}Space Invaders]{\includegraphics[scale=0.18]{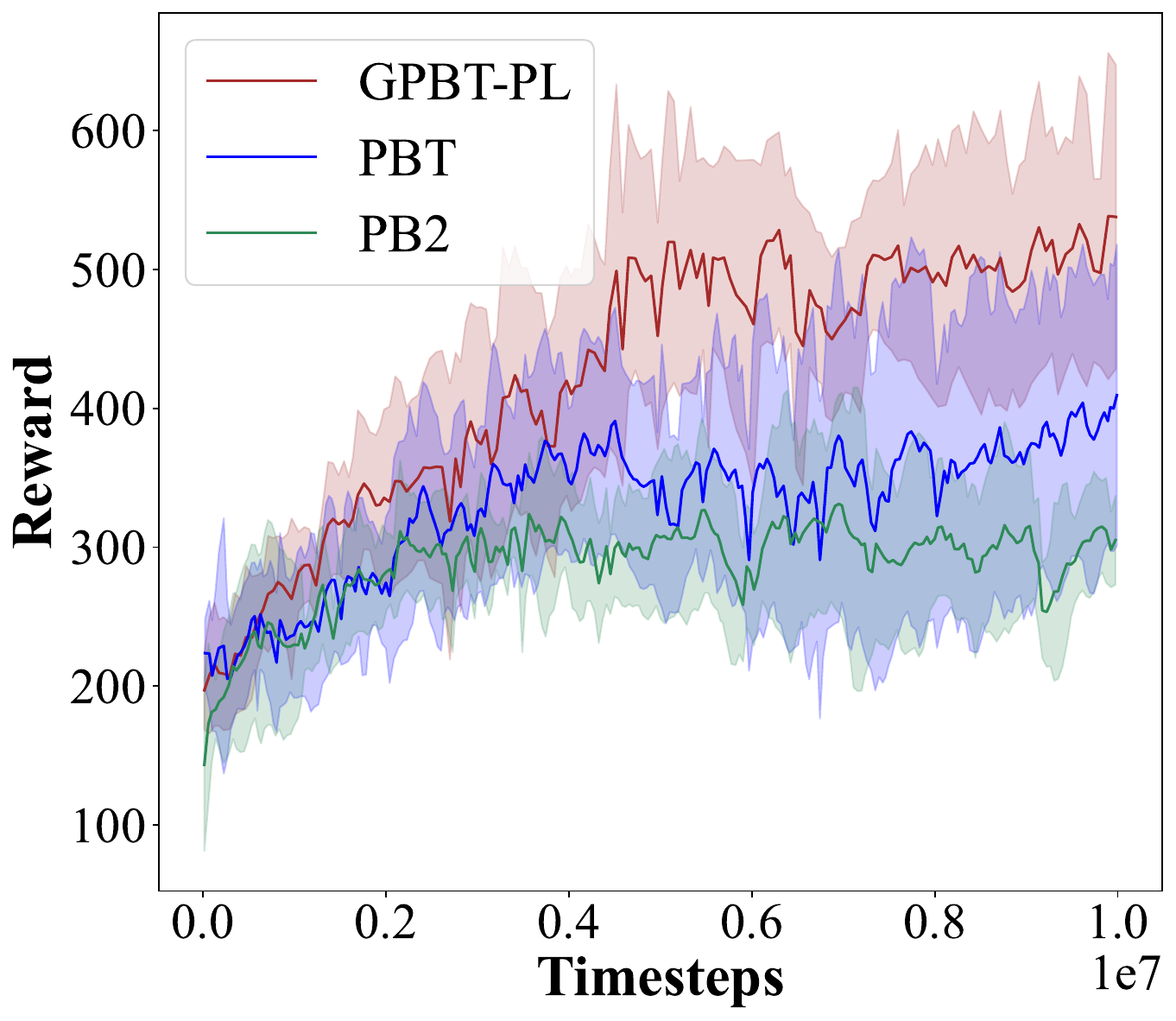}

}\hfill{}

\caption{\label{fig:IMPALA-1}Training curves for 4-agent populations using GPBT-PL, PBT, and PB2 on two OpenAI Gym games. The bold lines represent the average of the best rewards across 7 seeds, with shaded regions indicating standard deviation. The learning rate spans $[10^{-5},10^{-2}]$.}
\end{figure}

\subsection{\label{subsec:Robustness-to-Hyperparameter-Ranges}Robustness to Hyperparameter Ranges}
PBT-like HPO algorithms often struggle when the hyperparameter range is either not optimally defined or unknown. 
This issue is exacerbated by their reliance on population size for effective hyperparameter space exploration.
GPBT-PL addresses this challenge by learning potential update directions within the given range, thus negating the need for random sampling or incremental adjustments to pinpoint optimal regions.

To assess this robustness, we performed experiments on Breakout and SpaceInvaders, extending the learning rate range to {[}10$^{-5}$, 10$^{-2}${]}. 
This broader range can lead to inefficient agent initialization, especially when an agent uses a learning rate of 10$^{-2}$, causing policy learning to become unstable and divergent. 
Results in \tablename~\ref{tab:IMPALA-1} and \figurename~\ref{fig:IMPALA-1} show that while both GPBT-PL and PBT's performance declined compared to results in \tablename~\ref{tab:IMPALA} and \figurename~\ref{fig:IMPALA}, GPBT-PL consistently outperformed PBT. 
Specifically, GPBT-PL showcased an impressive 82\% improvement in Breakout.

\section{\label{sec:Conclusions}Conclusion}
In this work, we have endeavored to advance the state-of-the-art in hyperparameter optimization by refining the principles of Population-Based Training (PBT).
We began by introducing the Generalized Population-Based Training (GPBT), an extension of PBT that offers increased versatility. 
By allowing users to tailor hyperparameter learning methods during the perturbation phase, GPBT paves the way for a more adaptable optimization framework. 
Its asynchronous parallel structure is designed to cater to diverse optimization challenges efficiently.

Recognizing the inherent limitations of PBT, particularly its propensity for excessive greed and reliance on random heuristics, we proposed the Pairwise Learning (PL) method. 
By drawing on insights from the top-performing agents in the population, PL provides nuanced guidance for underperforming agents. This strategy not only facilitates quicker convergence but also ensures a comprehensive exploration of the hyperparameter space, thereby mitigating the risk of local optima.
The culmination of these efforts is the GPBT-PL framework, an amalgamation of the strengths of both GPBT and PL. 

Through rigorous experimentation in the RL domain, we demonstrated the superiority of GPBT-PL over traditional PBT and its Bayesian-optimized counterpart. 
Even in resource-constrained scenarios, GPBT-PL delivered consistently superior results.
Nevertheless, the performance superiority of GPBT-PL is not readily apparent in simpler problems; rather, its strengths are more pronounced in handling complex tasks. As the number of hyperparameters increases along with their expansive ranges, effectively navigating the search space becomes increasingly challenging, a common hurdle faced by conventional HPO methods. Notably, when an underperforming agent learns from a superior one, its hyperparameter updates may deviate in the wrong direction, potentially exacerbating its performance, particularly in scenarios with a large number of hyperparameters and broad ranges. Hence, there is a pressing need to devise a strategy capable of discerning the accurate update direction for hyperparameters.
Additionally, selecting hyperparameters from an extensive array of options presents a combinatorial optimization challenge, introducing novel hurdles for EC methods. Future investigations should delve into EC-based HPO methods capable of effectively navigating high-dimensional and combinatorial search spaces.

\bibliographystyle{IEEEtran}
\bibliography{main}

\begin{thebibliography}{10}
\providecommand{\url}[1]{#1}
\csname url@samestyle\endcsname
\providecommand{\newblock}{\relax}
\providecommand{\bibinfo}[2]{#2}
\providecommand{\BIBentrySTDinterwordspacing}{\spaceskip=0pt\relax}
\providecommand{\BIBentryALTinterwordstretchfactor}{4}
\providecommand{\BIBentryALTinterwordspacing}{\spaceskip=\fontdimen2\font plus
\BIBentryALTinterwordstretchfactor\fontdimen3\font minus \fontdimen4\font\relax}
\providecommand{\BIBforeignlanguage}[2]{{%
\expandafter\ifx\csname l@#1\endcsname\relax
\typeout{** WARNING: IEEEtran.bst: No hyphenation pattern has been}%
\typeout{** loaded for the language `#1'. Using the pattern for}%
\typeout{** the default language instead.}%
\else
\language=\csname l@#1\endcsname
\fi
#2}}
\providecommand{\BIBdecl}{\relax}
\BIBdecl

\bibitem{Silver2016}
D.~Silver, A.~Huang, C.~J. Maddison, A.~Guez, L.~Sifre, G.~Van Den~Driessche, J.~Schrittwieser, I.~Antonoglou, V.~Panneershelvam, M.~Lanctot \emph{et~al.}, ``Mastering the game of go with deep neural networks and tree search,'' \emph{Nature}, vol. 529, no. 7587, p. 484, 2016.

\bibitem{Mnih2015}
V.~Mnih, K.~Kavukcuoglu, D.~Silver, A.~A. Rusu, J.~Veness, M.~G. Bellemare, A.~Graves, M.~Riedmiller, A.~K. Fidjeland, G.~Ostrovski \emph{et~al.}, ``Human-level control through deep reinforcement learning,'' \emph{Nature}, vol. 518, no. 7540, pp. 529--533, 2015.

\bibitem{Lillicrap2015}
T.~P. Lillicrap, J.~J. Hunt, A.~Pritzel, N.~Heess, T.~Erez, Y.~Tassa, D.~Silver, and D.~Wierstra, ``Continuous control with deep reinforcement learning,'' in \emph{International Conference on Learning Representations}, 2016.

\bibitem{Chen2018a}
Y.~Chen, A.~Huang, Z.~Wang, I.~Antonoglou, J.~Schrittwieser, D.~Silver, and N.~de~Freitas, ``Bayesian optimization in alphago,'' \emph{arXiv preprint arXiv:1812.06855}, 2018.

\bibitem{Elsken2018}
T.~Elsken, J.~H. Metzen, and F.~Hutter, ``Neural architecture search: a survey,'' \emph{Journal of Machine Learning Research}, vol.~20, no.~1, pp. 1997--2017, 2019.

\bibitem{Parker-Holder2022}
J.~Parker-Holder, R.~Rajan, X.~Song, A.~Biedenkapp, Y.~Miao, T.~Eimer, B.~Zhang, V.~Nguyen, R.~Calandra, A.~Faust \emph{et~al.}, ``Automated reinforcement learning (autorl): A survey and open problems,'' \emph{arXiv preprint arXiv:2201.03916}, 2022.

\bibitem{Feurer2019}
M.~Feurer and F.~Hutter, ``Hyperparameter optimization,'' in \emph{AutoML: Methods, Sytems, Challenges}, F.~Hutter, L.~Kotthoff, and J.~Vanschoren, Eds.\hskip 1em plus 0.5em minus 0.4em\relax Springer, 2019, ch.~1, pp. 3--33.

\bibitem{Wu2019}
J.~Wu, X.-Y. Chen, H.~Zhang, L.-D. Xiong, H.~Lei, and S.-H. Deng, ``Hyperparameter optimization for machine learning models based on bayesian optimization,'' \emph{Journal of Electronic Science and Technology}, vol.~17, no.~1, pp. 26--40, 2019.

\bibitem{Bergstra2012}
J.~Bergstra and Y.~Bengio, ``Random search for hyper-parameter optimization,'' \emph{Journal of Machine Learning Research}, vol.~13, no.~2, 2012.

\bibitem{Jaderberg2017}
M.~Jaderberg, V.~Dalibard, S.~Osindero, W.~M. Czarnecki, J.~Donahue, A.~Razavi, O.~Vinyals, T.~Green, I.~Dunning, K.~Simonyan \emph{et~al.}, ``Population based training of neural networks,'' \emph{arXiv preprint arXiv:1711.09846}, 2017.

\bibitem{Jaderberg2019}
M.~Jaderberg, W.~M. Czarnecki, I.~Dunning, L.~Marris, G.~Lever, A.~G. Castaneda, C.~Beattie, N.~C. Rabinowitz, A.~S. Morcos, A.~Ruderman \emph{et~al.}, ``Human-level performance in 3{D} multiplayer games with population-based reinforcement learning,'' \emph{Science}, vol. 364, no. 6443, pp. 859--865, 2019.

\bibitem{Liu2019a}
S.~Liu, G.~Lever, J.~Merel, S.~Tunyasuvunakool, N.~Heess, and T.~Graepel, ``Emergent coordination through competition,'' in \emph{International Conference on Learning Representations}, 2019.

\bibitem{Liu2020e}
Y.~Liu, Y.~Gao, and W.~Yin, ``An improved analysis of stochastic gradient descent with momentum,'' \emph{Advances in Neural Information Processing Systems}, vol.~33, pp. 18\,261--18\,271, 2020.

\bibitem{Snoek2012}
J.~Snoek, H.~Larochelle, and R.~P. Adams, ``Practical bayesian optimization of machine learning algorithms,'' \emph{Advances in Neural Information Processing Systems}, vol.~25, 2012.

\bibitem{Li2017d}
L.~Li, K.~Jamieson, G.~DeSalvo, A.~Rostamizadeh, and A.~Talwalkar, ``Hyperband: A novel bandit-based approach to hyperparameter optimization,'' \emph{The Journal of Machine Learning Research}, vol.~18, no.~1, pp. 6765--6816, 2017.

\bibitem{Li2018b}
L.~Li, K.~Jamieson, A.~Rostamizadeh, E.~Gonina, M.~Hardt, B.~Recht, and A.~Talwalkar, ``Massively parallel hyperparameter tuning,'' \emph{arXiv preprint arXiv:1810.05934}, vol.~5, 2018.

\bibitem{Eriksson2003}
A.~Eriksson, G.~Capi, and K.~Doya, ``Evolution of meta-parameters in reinforcement learning algorithm,'' in \emph{Proceedings 2003 IEEE/RSJ International Conference on Intelligent Robots and Systems}, vol.~1.\hskip 1em plus 0.5em minus 0.4em\relax IEEE, 2003, pp. 412--417.

\bibitem{Elfwing2018}
S.~Elfwing, E.~Uchibe, and K.~Doya, ``Online meta-learning by parallel algorithm competition,'' in \emph{Proceedings of the Genetic and Evolutionary Computation Conference}, 2018, pp. 426--433.

\bibitem{Paul2019}
S.~Paul, V.~Kurin, and S.~Whiteson, ``Fast efficient hyperparameter tuning for policy gradient methods,'' \emph{Advances in Neural Information Processing Systems}, vol.~32, 2019.

\bibitem{Parker-Holder2020a}
J.~Parker-Holder, V.~Nguyen, and S.~J. Roberts, ``Provably efficient online hyperparameter optimization with population-based bandits,'' \emph{Advances in Neural Information Processing Systems}, vol.~33, pp. 17\,200--17\,211, 2020.

\bibitem{Bai2023}
\BIBentryALTinterwordspacing
H.~Bai, R.~Cheng, and Y.~Jin, ``Evolutionary reinforcement learning: A survey,'' \emph{Intelligent Computing}, vol.~2, p. 0025, 2023. [Online]. Available: \url{https://spj.science.org/doi/abs/10.34133/icomputing.0025}
\BIBentrySTDinterwordspacing

\bibitem{Falkner2018}
S.~Falkner, A.~Klein, and F.~Hutter, ``Bohb: Robust and efficient hyperparameter optimization at scale,'' in \emph{International Conference on Machine Learning}.\hskip 1em plus 0.5em minus 0.4em\relax PMLR, 2018, pp. 1437--1446.

\bibitem{Aszemi2019}
N.~M. Aszemi and P.~Dominic, ``Hyperparameter optimization in convolutional neural network using genetic algorithms,'' \emph{International Journal of Advanced Computer Science and Applications}, vol.~10, no.~6, 2019.

\bibitem{Espeholt2018}
L.~Espeholt, H.~Soyer, R.~Munos, K.~Simonyan, V.~Mnih, T.~Ward, Y.~Doron, V.~Firoiu, T.~Harley, I.~Dunning \emph{et~al.}, ``{IMPALA:} scalable distributed deep-rl with importance weighted actor-learner architectures,'' in \emph{International Conference on Machine Learning}, 2018.

\bibitem{Wu2020}
T.~R. Wu, T.~H. Wei, and I.~C. Wu, ``Accelerating and improving alphazero using population based training,'' in \emph{Proceedings of the AAAI Conference on Artificial Intelligence}, 2020.

\bibitem{Schmitt2018}
S.~Schmitt, J.~J. Hudson, A.~Zidek, S.~Osindero, C.~Doersch, W.~M. Czarnecki, J.~Z. Leibo, H.~Kuttler, A.~Zisserman, K.~Simonyan \emph{et~al.}, ``Kickstarting deep reinforcement learning,'' \emph{arXiv preprint arXiv:1803.03835}, 2018.

\bibitem{Jung2020}
W.~Jung, G.~Park, and Y.~Sung, ``Population-guided parallel policy search for reinforcement learning,'' in \emph{International Conference on Learning Representations}, 2020.

\bibitem{Liu2021b}
Q.~Liu, Y.~Wang, and X.~Liu, ``Pns: Population-guided novelty search for reinforcement learning in hard exploration environments,'' in \emph{2021 IEEE/RSJ International Conference on Intelligent Robots and Systems}, 2021.

\bibitem{Khadka2018}
S.~Khadka and K.~Tumer, ``Evolution-guided policy gradient in reinforcement learning,'' in \emph{International Conference on Neural Information Processing Systems}, 2018.

\bibitem{Khadka2019}
S.~Khadka, S.~Majumdar, T.~Nassar, Z.~Dwiel, E.~Tumer, S.~Miret, Y.~Liu, and K.~Tumer, ``Collaborative evolutionary reinforcement learning,'' \emph{International Conference on Machine Learning}, 2019.

\bibitem{Majumdar2020}
S.~Majumdar, S.~Khadka, S.~Miret, S.~Mcaleer, and K.~Tumer, ``Evolutionary reinforcement learning for sample-efficient multiagent coordination,'' in \emph{International Conference on Machine Learning}, 2020.

\bibitem{Conti2018}
E.~Conti, V.~Madhavan, F.~Petroski~Such, J.~Lehman, K.~Stanley, and J.~Clune, ``Improving exploration in evolution strategies for deep reinforcement learning via a population of novelty-seeking agents,'' \emph{Advances in Neural Information Processing Systems}, vol.~31, 2018.

\bibitem{Lehman2011}
J.~Lehman and K.~O. Stanley, ``Evolving a diversity of virtual creatures through novelty search and local competition,'' in \emph{Proceedings of the 13th Annual Conference on Genetic and Evolutionary Computation}, 2011.

\bibitem{Golovin2017}
D.~Golovin, B.~Solnik, S.~Moitra, G.~Kochanski, J.~Karro, and D.~Sculley, ``Google vizier: A service for black-box optimization,'' in \emph{Proceedings of the 23rd ACM SIGKDD international conference on knowledge discovery and data mining}, 2017, pp. 1487--1495.

\bibitem{Dalibard2021}
V.~Dalibard and M.~Jaderberg, ``Faster improvement rate population based training,'' \emph{arXiv preprint arXiv:2109.13800}, 2021.

\bibitem{Wan2022}
X.~Wan, C.~Lu, J.~Parker-Holder, P.~J. Ball, V.~Nguyen, B.~Ru, and M.~Osborne, ``Bayesian generational population-based training,'' in \emph{International Conference on Learning Representations Workshop on Agent Learning in Open-Endedness}, 2022.

\bibitem{Vavak1996}
F.~Vavak and T.~C. Fogarty, ``Comparison of steady state and generational genetic algorithms for use in nonstationary environments,'' in \emph{Proceedings of IEEE International Conference on Evolutionary Computation}.\hskip 1em plus 0.5em minus 0.4em\relax IEEE, 1996, pp. 192--195.

\bibitem{Jiang2017}
S.~Jiang and S.~Yang, ``A steady-state and generational evolutionary algorithm for dynamic multiobjective optimization,'' \emph{IEEE Transactions on Evolutionary Computation}, vol.~21, no.~1, pp. 65 -- 82, February 2017.

\bibitem{Dyer2012}
J.~D. Dyer, R.~J. Hartfield, G.~V. Dozier, and J.~E. Burkhalter, ``Aerospace design optimization using a steady state real-coded genetic algorithm,'' \emph{Applied Mathematics and Computation}, vol. 218, no.~9, pp. 4710--4730, 2012.

\bibitem{Lozano2008}
M.~Lozano, F.~Herrera, and J.~R. Cano, ``Replacement strategies to preserve useful diversity in steady-state genetic algorithms,'' \emph{Information sciences}, vol. 178, no.~23, pp. 4421--4433, 2008.

\bibitem{Cormen2001}
T.~H. Cormen, C.~E. Leiserson, R.~L. Rivest, and C.~Stein, ``The knuth-morris-pratt algorithm,'' in \emph{Introduction to algorithms second edition}.\hskip 1em plus 0.5em minus 0.4em\relax MIT press, 2001, pp. 1002--1013.

\bibitem{Livni2014}
R.~Livni, S.~Shalev-Shwartz, and O.~Shamir, ``On the computational efficiency of training neural networks,'' \emph{Advances in neural information processing systems}, vol.~27, 2014.

\bibitem{Cheng2014}
R.~Cheng and Y.~Jin, ``A competitive swarm optimizer for large scale optimization,'' \emph{IEEE Transactions on Cybernetics}, vol.~45, no.~2, pp. 191--204, 2014.

\bibitem{Henderson2018}
P.~Henderson, R.~Islam, P.~Bachman, J.~Pineau, D.~Precup, and D.~Meger, ``Deep reinforcement learning that matters,'' in \emph{Proceedings of the AAAI conference on artificial intelligence}, vol.~32, no.~1, 2018.

\bibitem{Brockman2016}
G.~Brockman, V.~Cheung, L.~Pettersson, J.~Schneider, J.~Schulman, J.~Tang, and W.~Zaremba, ``Open{AI} {G}ym,'' \emph{arXiv preprint arXiv:1606.01540}, 2016.

\bibitem{Schulman2017}
J.~Schulman, F.~Wolski, P.~Dhariwal, A.~Radford, and O.~Klimov, ``Proximal policy optimization algorithms,'' \emph{arXiv preprint arXiv:1707.06347}, 2017.

\bibitem{Moritz2018}
P.~Moritz, R.~Nishihara, S.~Wang, A.~Tumanov, R.~Liaw, E.~Liang, M.~Elibol, Z.~Yang, W.~Paul, M.~I. Jordan, and I.~Stoica, ``Ray: A distributed framework for emerging {AI} applications,'' in \emph{13th {USENIX} Symposium on Operating Systems Design and Implementation ({OSDI} 18)}.\hskip 1em plus 0.5em minus 0.4em\relax Carlsbad, CA: {USENIX} Association, Oct. 2018, pp. 561--577.

\bibitem{Liang2018}
E.~Liang, R.~Liaw, R.~Nishihara, P.~Moritz, R.~Fox, K.~Goldberg, J.~Gonzalez, M.~Jordan, and I.~Stoica, ``{RL}lib: Abstractions for distributed reinforcement learning,'' in \emph{International Conference on Machine Learning (ICML)}, 2018.

\bibitem{Bellemare2013}
M.~G. Bellemare, Y.~Naddaf, J.~Veness, and M.~Bowling, ``The arcade learning environment: An evaluation platform for general agents,'' \emph{Journal of Artificial Intelligence Research}, vol.~47, pp. 253--279, 2013.

\bibitem{Mnih2016}
V.~Mnih, A.~P. Badia, M.~Mirza, A.~Graves, T.~Lillicrap, T.~Harley, D.~Silver, and K.~Kavukcuoglu, ``Asynchronous methods for deep reinforcement learning,'' in \emph{International Conference on Machine Learning}, 2016.

\end{thebibliography}




\end{document}